\documentclass{article}

\usepackage[preprint]{neurips_2025}

\usepackage{natbib}
\usepackage[T1]{fontenc}
\usepackage{amsmath,amssymb,amsfonts}
\usepackage{amsthm}
\usepackage{algorithmic}
\usepackage{graphicx}
\usepackage{subcaption}
\usepackage{textcomp}
\usepackage{xcolor}
\usepackage{booktabs}
\usepackage{makecell}

\usepackage{titlesec}
\usepackage[ruled,linesnumbered]{algorithm2e}

\usepackage{hyperref}

\usepackage{amsmath}
\usepackage{amssymb}
\usepackage{mathtools}
\usepackage{amsthm}

\usepackage[capitalize,noabbrev]{cleveref}

\theoremstyle{plain}
\newtheorem{theorem}{Theorem}[section]

\theoremstyle{definition}

\newtheorem{assumption}[theorem]{Assumption}
\theoremstyle{remark}

\usepackage[textsize=tiny]{todonotes}

\title{GradStop: Exploring Training Dynamics in Unsupervised Outlier Detection through Gradient}

\author{
  Yuang Zhang \\
  East China Normal University \\
  Shanghai, China \\
  \texttt{zgcj20@gmail.cn} \\
  \And
  Liping Wang \\
  East China Normal University \\
  Shanghai, China \\
  \texttt{\small lipingwang@sei.ecnu.edu.cn} \\
  \And
  Yihong Huang \\
  Bilibili Inc.\\
  Shanghai, China \\
  \texttt{hyh957947142@gmail.com}\\
  \And
  Yuanxing Zheng \\
  East China Normal University \\
  Shanghai, China \\
  \texttt{2957573678@qq.com}\\
  \And
  Fan Zhang \\
  Guangzhou University \\
  Guangzhou, China \\
  \texttt{fanzhang.cs@gmail.com}\\
  \And
  Xuemin Lin \\
  Shanghai Jiao Tong University\\
  Shanghai, China\\
  xuemin.lin@gmail.com\\
}

\begin{document}

\maketitle

\begin{abstract}
Unsupervised Outlier Detection (UOD) is a crucial task that identifies majority-deviated instances. Due to the absence of labels, deep UOD methods often struggle with misalignment between optimization goals and OD performance goals, causing performance degradation. To address this, we propose an early stopping algorithm, GradStop, which provides an effective, robust, and label-free method based on the connection between UOD assumptions and training dynamics. It monitors and stops training when the model performs optimally in detecting outliers. Specifically, we devise a label-free sampling method to retrieve two sets, each more closely reflecting the distribution characteristics of inliers and outliers respectively, and propose a gradient-based approach to estimate real-time model performance during training. Experimental results on four classic deep UOD algorithms across 47 real-world datasets and theoretical proofs demonstrate that GradStop effectively mitigates performance degradation. AutoEncoder (AE) with GradStop significantly outperforms vanilla AE and also surpasses ensemble AEs and other SOTA UOD methods.
\end{abstract}

\section{Introduction}
\subsection{Unsupervised Outlier Detection}

Outlier Detection (OD) is a fundamental task in data mining and machine learning, focused on identifying instances that significantly deviate from the majority \cite{od-survey}, i.e., in our context, inliers. Outliers, usually a minority in the dataset, are alternatively referred to as anomalies, deviants, novelties, or exceptions \cite{od-survey}. OD has received continuous research interests \cite{ADbench,Ts-benchmark,god-benchmark} due to its wide applications in various fields, such as finance \cite{financial-example}, security \cite{security-example}, and so on. Depending on the availability of label information, OD methodologies can be classified into three categories: Supervised OD, Semi-Supervised OD, and Unsupervised OD \cite{inlier-priority}. Recently, with the development of deep learning, deep OD algorithms have been proposed \cite{deep-od-survey-2021,deep-od-survey-2019,deep-od-survey-3}, demonstrating their superior capabilities in managing complex and high-dimensional data compared to traditional methods.

Unsupervised OD (UOD) targets to detect outliers in a contaminated dataset having both inliers and outliers without the availability of any label \cite{inlier-priority} by assigning higher anomaly scores to outliers. It is important to differentiate between two fundamental paradigms within UOD. The first paradigm, including algorithms such as DeepSVDD \cite{deep-svdd}, NeuTraL AD \cite{NTL}, ICL \cite{ICL}, and AnoGAN \cite{AnoGAN}, is training models exclusively on clean datasets free of outliers. However, this approach requires collecting large volumes of uncontaminated data before performing OD, thus being less practical.

In contrast, the second category of UOD algorithms including methods like RandNet \cite{randnet}, ROBOD \cite{robod}, RDP \cite{RDP}, RDA \cite{RDA}, IsolationForest \cite{IsolationForest}, and GAAL \cite{gan-ensemble}, is designed to function directly on contaminated datasets that contain outliers. Thus, these models are able to identify outliers within the training set itself and unseen datasets provided that the sample characteristic of the unseen dataset aligns with that of the original training set. Our research focuses on the latter paradigm, where UOD models are trained on datasets that include outliers, presenting a wider use along with more challenges. For both training and validation, no label is available. 

\subsection{Challenge}



\begin{figure}[t]
  \centering
  \begin{minipage}{0.48\textwidth}
    \centering
    \includegraphics[width=0.48\textwidth]{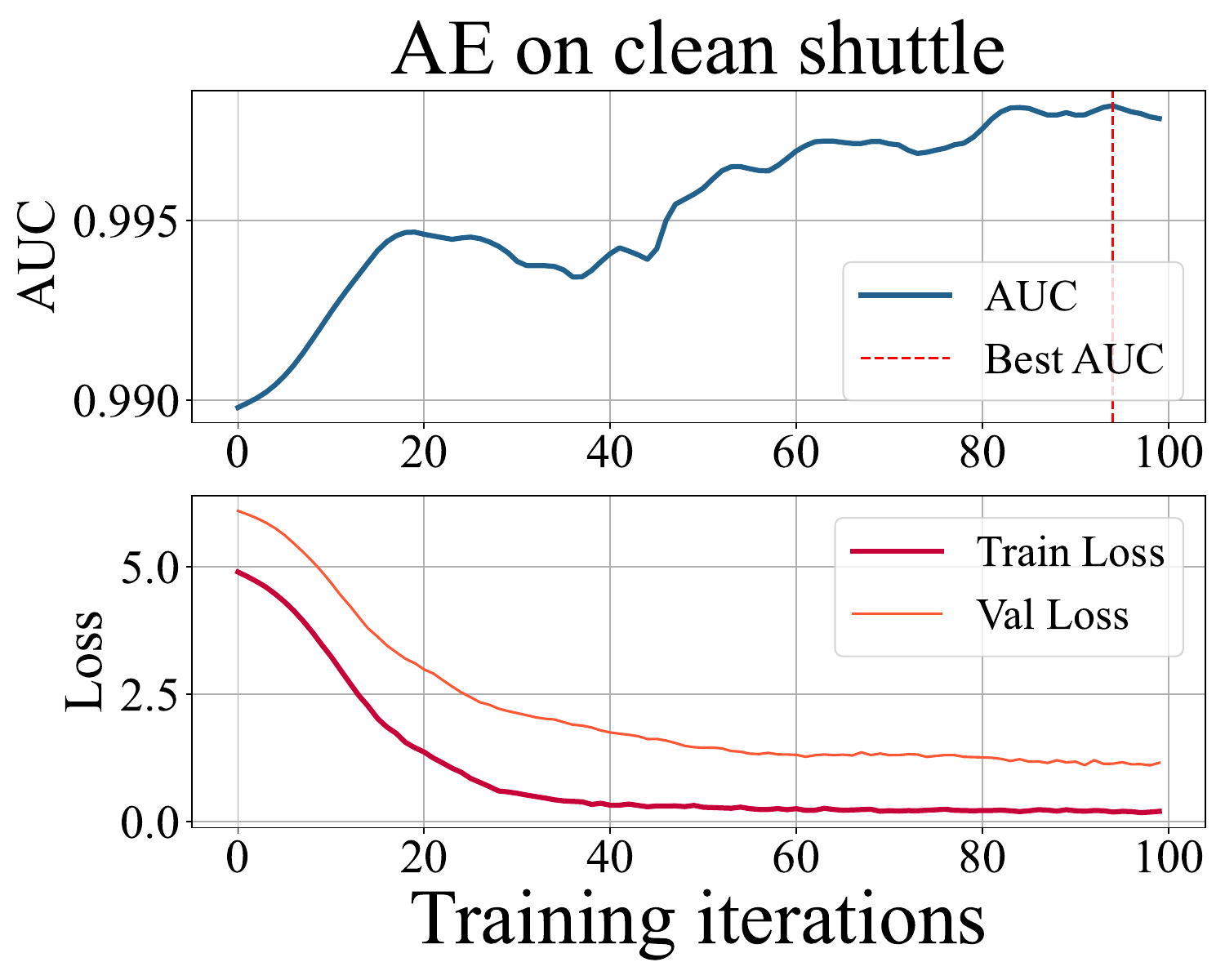}
    \includegraphics[width=0.48\textwidth]{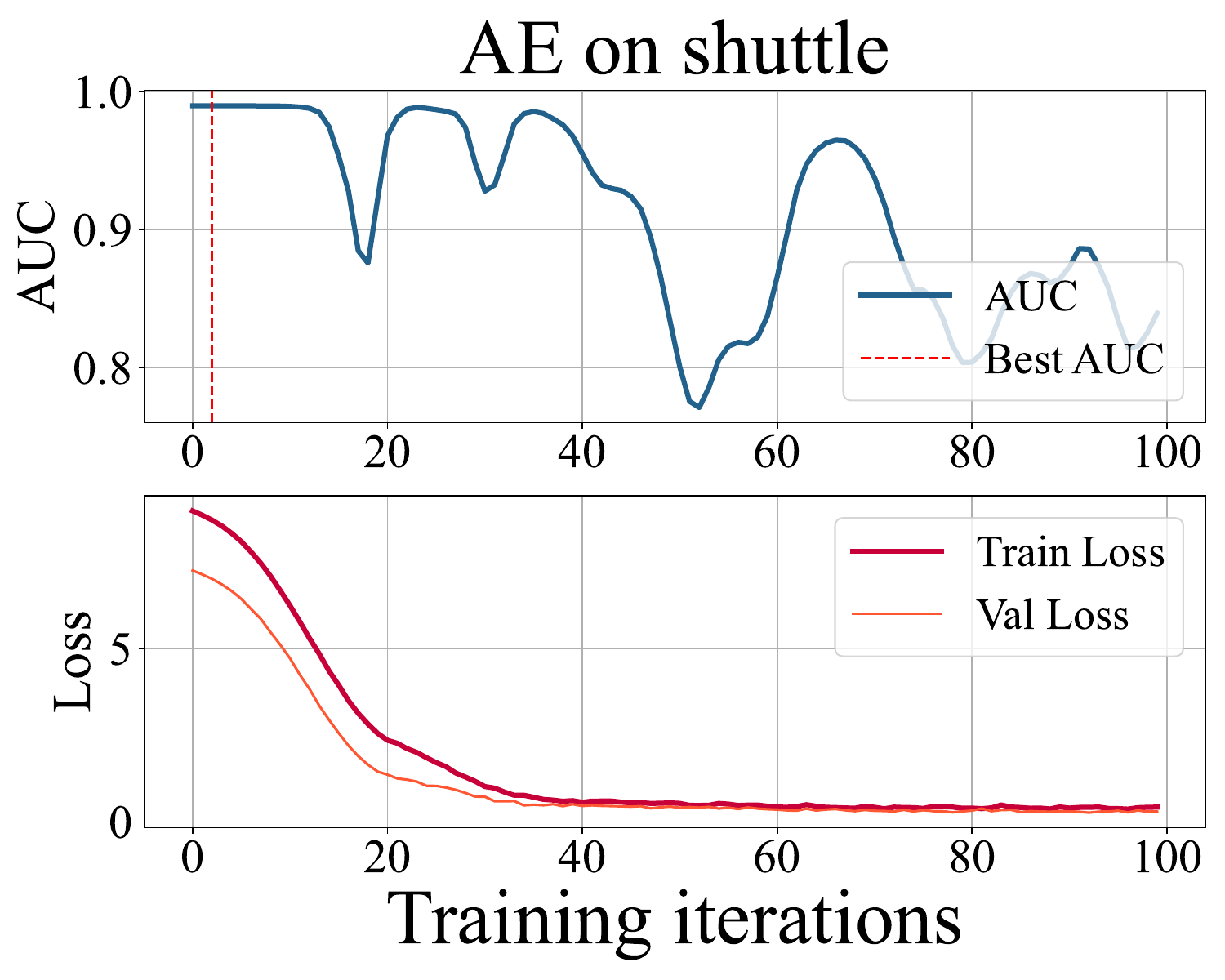}
    \caption{Performance degradation: UOD training process of AutoEncoder on \textbf{clean} dataset \textit{shuttle} and \textbf{original} polluted \textit{shuttle}.}
    \label{Fig:loss-auc-data}
  \end{minipage}\hfill
  \begin{minipage}{0.48\textwidth}
    \centering
    \includegraphics[width=0.48\textwidth]{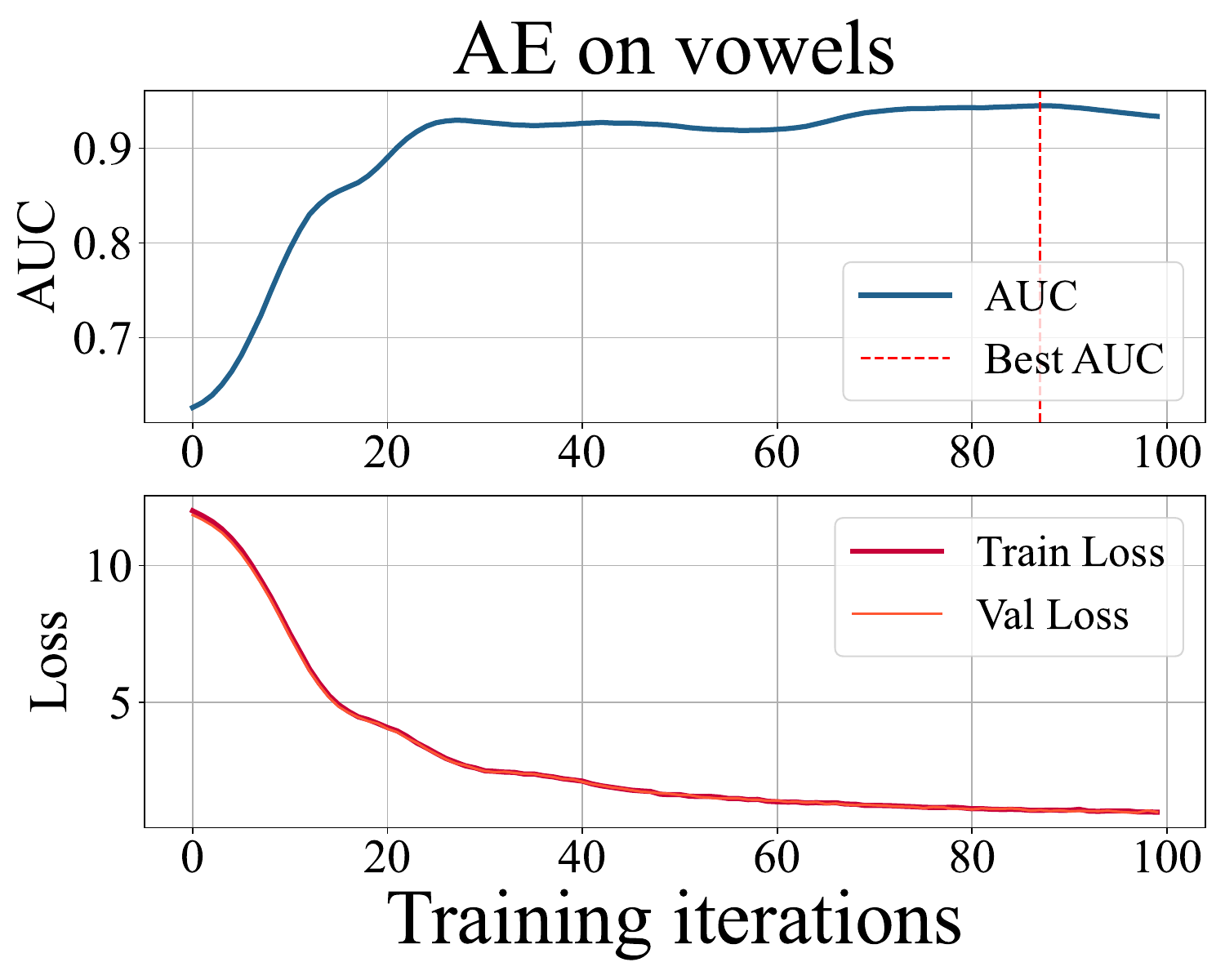}
    \includegraphics[width=0.48\textwidth]{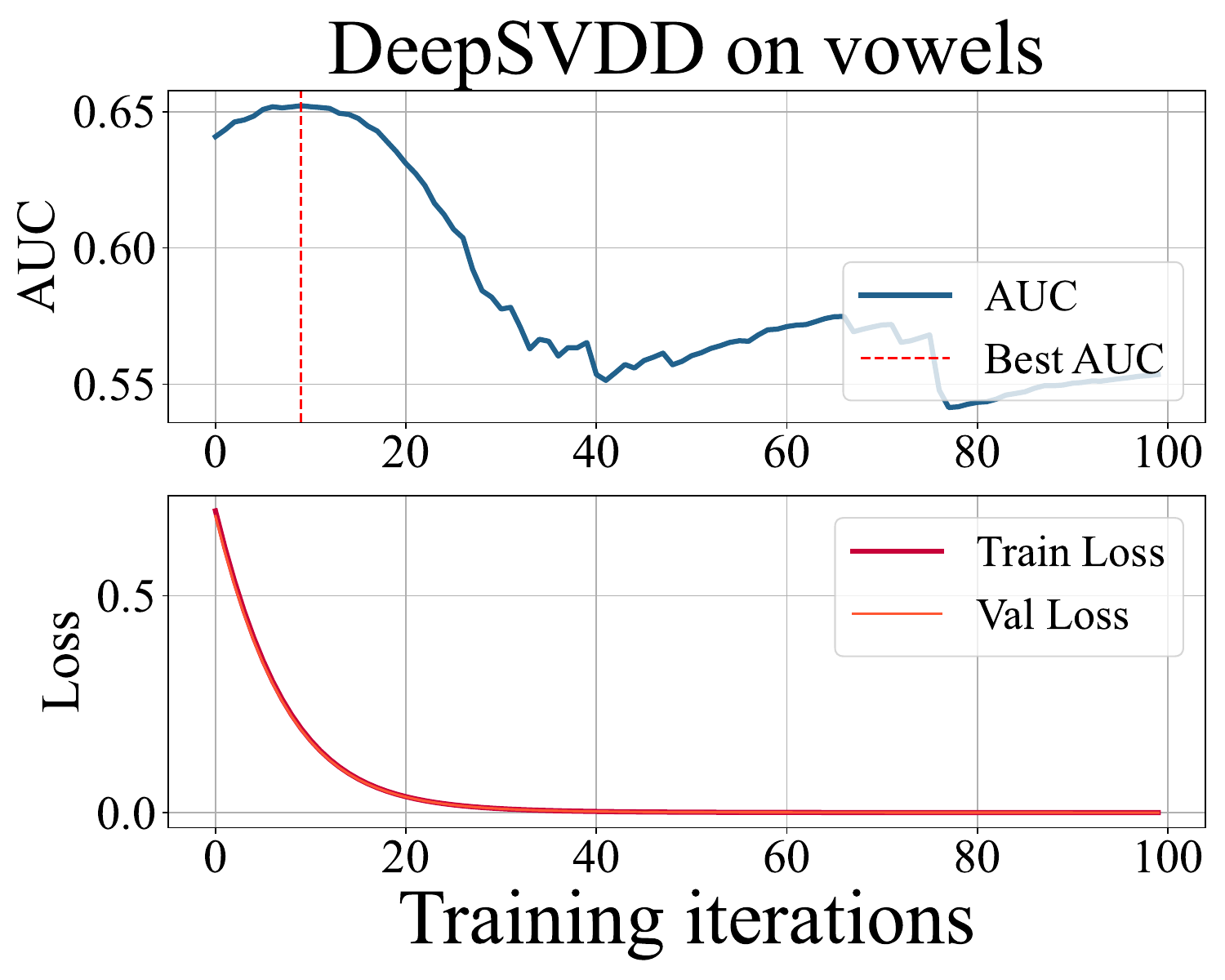}
    \caption{Various performance trends: UOD training process of AutoEncoder and DeepSVDD on dataset \textit{vowels}.}
    \label{Fig:loss-auc-alg}
  \end{minipage}
\end{figure}

Training on contaminated datasets leads to performance degradation. 
Specifically for UOD training scenarios, it can be viewed as model misfitting or overfitting to outliers. In essence, it is caused by the misalignment between the model's direct optimization goal and the final performance goal of the OD task. 

Due to the unsupervised training setup, UOD can only leverage the distributional differences between outliers and normal instances instead of label information to detect outliers. Consequently, UOD models are designed based on outlier assumptions targeting these distributional discrepancies \cite{ADbench}. This leads to an inherent feature of UOD: the optimization objective of the model is not aligned with the final performance goal of anomaly detection. Therefore, in UOD, performance metrics (such as AUC, AP, etc.) do not necessarily improve strictly with the investment of training resources (e.g., number of training epochs, model size, etc.).

A typical example of AE is shown in Fig. \ref{Fig:loss-auc-data}. Outliers from the original dataset \textit{shuttle} are excluded to gain \textit{\textbf{clean} shuttle}. On \textit{shuttle}, although the loss value is consistently optimized, AE's performance tends to fluctuate or even decline as the number of epochs increases due to the existence of outliers, which is also observed by \citeauthor{EntropyStop} By contrast, training on \textit{clean shuttle} (which requires labels) is not degrading AUC.
This phenomenon occurs because, as training progresses, the model begins to reconstruct outliers more accurately, thereby violating the outlier assumption of AE which posits that outliers have larger losses than inliers. Although the model fits better to the entire dataset minimizing the losses, OD performance still deteriorates.
Previous studies have identified and termed it ``inlier priority'' \cite{inlier-priority} or ``inlier-memorization'' \cite{inlier-memorization} which signifies that the model fits inliers faster and outliers slower. This phenomenon is a manifestation of the outlier assumption during the model training process and also a fundamental mechanism ensuring UOD can effectively detect anomalies. Because inliers are easier to fit, there exists a disparity in the degree to which the model fits the inlier distribution versus the outlier distribution at specific stages of training, leading the model to assign higher anomaly scores to outliers. As training continues, this disparity diminishes, resulting in performance degradation.

To achieve the full potential of deep models in OD tasks on polluted datasets, there are currently two main branches of research. 
One approach mainly relies on overfitting-based ensemble learning \cite{gan-ensemble, RDP, randnet, robod}. However, overtraining numerous models brings about large time and computation costs. 
Another feasible approach is to use early stopping to prevent degradation. Previous study \cite{EntropyStop} utilizes a negative probability correlation between AUC and an entropy metric of sample losses to stop training on AE. However, the entropy metric neglects the connection between training dynamics and OD goal, limiting its performance and robustness. Early stopping in UOD still involves the following challenges:


\begin{itemize}
    \item  \textbf{Unsupervised training setup on polluted datasets.} Given the fully unsupervised training configuration, a validation set to monitor the model's real-time performance is unavailable. Instead, we need a label-free yet effective evaluation metric that can infer the model's OD performance in real-time.
    \item \textbf{Generalization ability on algorithm-dataset pairs.} The satisfaction degree of outlier assumption highly depends on the specific combination of algorithm and dataset, which presents a challenge in finding a generalized metric across different algorithm-dataset pairs.
    \item \textbf{Various performance trends.} The diversity of algorithm-dataset pairs and the associated outlier assumptions can lead the performance during training to exhibit various trends. For example, when training AE and DeepSVDD on dataset \textit{vowels} as shown in Fig. \ref{Fig:loss-auc-alg}, the performance of DeepSVDD starts declining since its beginning. Conversely, AE exhibits the opposite behavior with continuous ascending performance. The proposed algorithm must be robust enough to stop training DeepSVDD immediately and to continue training AE on \textit{vowels}.
\end{itemize}

\subsection{Our Solution}\label{solution}
To meet the challenges, we aim to provide an effective, robust, and label-free method that can utilize various training dynamics of deep UOD to prevent degradation, enhancing the performance on the OD task rather than blindly fitting the contaminated dataset.

From the perspective of data distribution, we thoroughly analyze the connection between training dynamics and outlier assumptions both theoretically and empirically. This also provides a comprehensive explanation for the inlier priority phenomenon previously observed by other studies \cite{inlier-priority, inlier-memorization}. Inlier priority is a typical phenomenon in UOD training, indicating that training is beneficial for OD tasks as the disparity between inliers and outliers increases.

Based on the theoretical and empirical analysis, we propose an early stopping algorithm, \textit{GradStop}, to stop training when the training dynamics indicate that inlier priority no longer holds. First, inspired by outlier distribution characteristics and their effect on UOD training dynamics, a gradient-based and label-free sampling method is devised to retrieve two sets of data points. Each of them approximates the distributions of inliers and outliers respectively and manifests their distributional characteristics.
Then two novel metrics, the inner cohesion degree and inter-divergence degree, are applied to the two sets to estimate the satisfaction degree of inlier priority during training. The metrics closely reflect inlier priority, thus focusing on the fundamental deep UOD working mechanism regardless of diverse algorithm-dataset pairs. Finally, an automated early stopping algorithm determines whether to stop training. Targeting unsupervised setup, the whole process is designed to be \textbf{label-free}. 
Our key contributions are:

\begin{itemize}
    \item We elaborately analyze the inlier priority phenomenon both theoretically and empirically, explaining the converging mechanism of deep UOD models to inliers and outliers.
    \item We propose a label-free sampling method to retrieve data point sets with collective-level label information. Two novel metrics are designed to explore UOD training dynamics and identify toxic training. An early stopping algorithm, GradStop, is then proposed to prevent degradation.
    \item Experiments on four classic deep UOD models demonstrate that GradStop can effectively mitigate performance degradation problems. Specifically, GradStop can significantly improve AE performance and outperform state-of-the-art methods. 
\end{itemize}

\section{Related Work}
\subsection{Unsupervised Outlier Detection}
Unsupervised Outlier Detection (UOD) is a vibrant and rapidly evolving research area focused on identifying outliers in datasets without labeled data \cite{od-survey}. Traditional methods include Isolation Forest \cite{IsolationForest}, ECOD \cite{ecod}, KNN \cite{knn}, LOF \cite{LOF}, etc. Recently, deep methods with neural networks have shown advantages in handling large-scale, high-dimensional, and complex data \cite{deep-od-survey-2019, deep-od-survey-2021, deep-od-survey-3}, and have much more potential in generalization on various datasets than traditional methods.

Many deep UOD models \cite{ICL, NTL, deep-svdd, AnoGAN} are trained exclusively on clean datasets to learn the distribution of inliers, thus excluding unseen outliers when testing. However, real-world datasets are usually large and may inadvertently contain outliers that the model should detect \cite{LOE}. To address this, studies \cite{dagmm, RDA, RDP} have focused on algorithms that train directly on contaminated datasets, where the goal misalignment exists.

Efforts have been made to mitigate the model degradation caused by the misalignment. Model ensemble approaches \cite{robod, randnet, gan-ensemble} are proposed with better performance and robustness against hyperparameters (HPs). Additionally, \cite{LOE, outlier_refine_2015, outlier_refine_2021} adapt models initially trained on clean datasets to perform effectively on contaminated datasets through outlier refinement processes.

\subsection{Early Stopping}
Early stopping is an effective and widely employed technique in machine learning, designed to halt training when further iterations no longer benefit the final performance. One well-known application of early stopping is to mitigate overfitting with cross-validation \cite{ES_1st}. More recently, a deeper understanding of training dynamics has revealed the practical utility of early stopping in scenarios involving noisy labels \cite{ES_noise_0, ES_noise_1, ES_noise_2, ES_noise_3}. 
These studies indicate that overfitting to noisy samples in the later training stages can degrade model performance, which can be alleviated remarkably with early stopping with validation labels. However, label-free early stopping methods remain scarce. \citeauthor{EntropyStop} is the first to explore the potential of early stopping in UOD with loss entropy as the stopping metric to solve model degradation problems. However, loss entropy neglects the connection between training dynamics and OD goals, limiting its performance and generalization ability. Our work further delves into early stopping UOD, proposing a more intriguing stopping scheme.

\section{Preliminary}
\noindent \textbf{Problem Formulation} (Unsupervised OD).
\textit{Considering a data space $\mathcal{X}$,
an unlabeled dataset $\mathcal{D} = \{\textbf{x}_j\}_{j=1}^n$ consists of an inlier set $\mathcal{D}_{in}$ and an outlier set $\mathcal{D}_{out}$, which originate from two different underlying distributions $\mathcal{X}_{in}$ and $\mathcal{X}_{out}$, respectively \cite{uod-definition}. The goal is to learn an outlier score function $f(\cdot)$ to calculate the outlier score value $v_j = f(\textbf{x}_j)$ for each data point $\textbf{x}_j \in \mathcal{D}$. Without loss of generality, a higher $f(\textbf{x}_j)$ indicates more likelihood of $\textbf{x}_j$ to be an outlier.
}

\noindent \textbf{Unsupervised Training Formulation for OD.} Given a UOD model $M$, at each iteration, a batch of instances $B = \{x_0,x_1,...,x_{n}\}$ is sampled from the data space $\mathcal{X}$. The loss $\mathcal{L}$ for model $M$ is calculated over $B$ as follows:


\begin{equation}
    \scalebox{0.9}{
    $\displaystyle
    \mathcal{L}(M; B) = \frac{1}{|B|} \sum_{x \in B} \mathcal{J}_M(x) =  \\
    \frac{1}{|B|} \sum_{x \in B} f_M(x) = \frac{1}{|B|} \sum_i v_i$
    }
\end{equation}

where $\mathcal{J}_M(\cdot)$ denotes the unsupervised loss function of $M$ while  $\mathcal{L}$ denotes the loss based on which the model $M$ updates its parameters by minimizing $\mathcal{L}$. To facilitate understanding, we assume here the unsupervised loss function $\mathcal{J}_M(\cdot)$ and outlier score function $f_M(\cdot)$ are identical. 

\noindent \textbf{Objective: }
The objective is to train the model $M$ such that it achieves the best detection performance on $\mathcal{X}$. Specifically, we aim to maximize the probability that an inlier from $\mathcal{X}_{in}$ has a lower outlier score than an outlier from $\mathcal{X}_{out}$, i.e., 
\begin{equation}
    \scalebox{0.85}{
    $\displaystyle
        P(v^{-} < v^{+}) = P(f_M(x_{in})< f_M(x_{out})| x_{in} \sim \mathcal{X}_{in}, x_{out} \sim \mathcal{X}_{out})$ \notag
    }
\end{equation}
as large as possible, where $f_M(\cdot)$ is the outlier score function learned by model $M$. Let $\mathcal{O}_{in}$ and $\mathcal{O}_{out}$ represent the distributions of $f_M(x)$, where $x$ is drawn from $\mathcal{X}_{in}$ and $\mathcal{X}_{out}$, respectively. Therefore, $v^{-} \sim \mathcal{O}_{in}$ and $v^{+} \sim \mathcal{O}_{out}$ denotes the corresponding random variable of outlier score. AUC \cite{auc} is a widely-used metric to evaluate the OD performance, which can be formulated as:
\begin{align}
 &AUC(M,\mathcal{D}) = \frac{1}{|\mathcal{D}_{in}| |\mathcal{D}_{out}|} \sum_{\textbf{x}_i \in \mathcal{D}_{in}} \sum_{\textbf{x}_j \in \mathcal{D}_{out}}\mathbb{I}(f_M(\textbf{x}_i) < f_M(\textbf{x}_j)) 
 \label{eq: auc}
 \end{align}
 where $\mathbb{I}$ is an indicator function. 
In practice, AUC is discretely computed on a dataset, and the expression $P(v^- < v^+)$ is the continuous form of AUC.

\section{Methodology}
\begin{figure*}[tb]
    \centering
    \begin{subfigure}{0.78\textwidth}
        \centering
        \includegraphics[width=\textwidth]{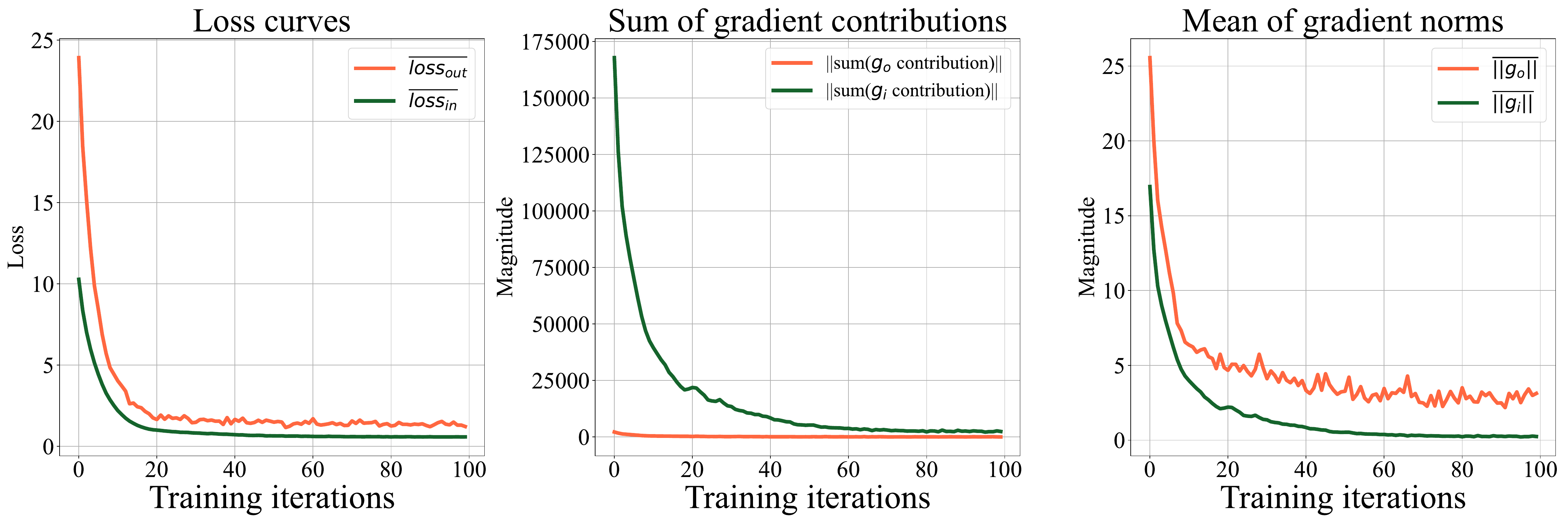}
        \caption{The curves from left to right represent: the average in/outlier loss values, the magnitude of the sum of in/outlier gradient contributions in gradient updates, and the average in/outlier gradient magnitudes. }
        \label{fig:training-dynamics-sub1}
    \end{subfigure}

    \begin{subfigure}{0.27\textwidth}
        \centering
        \includegraphics[width=\textwidth]{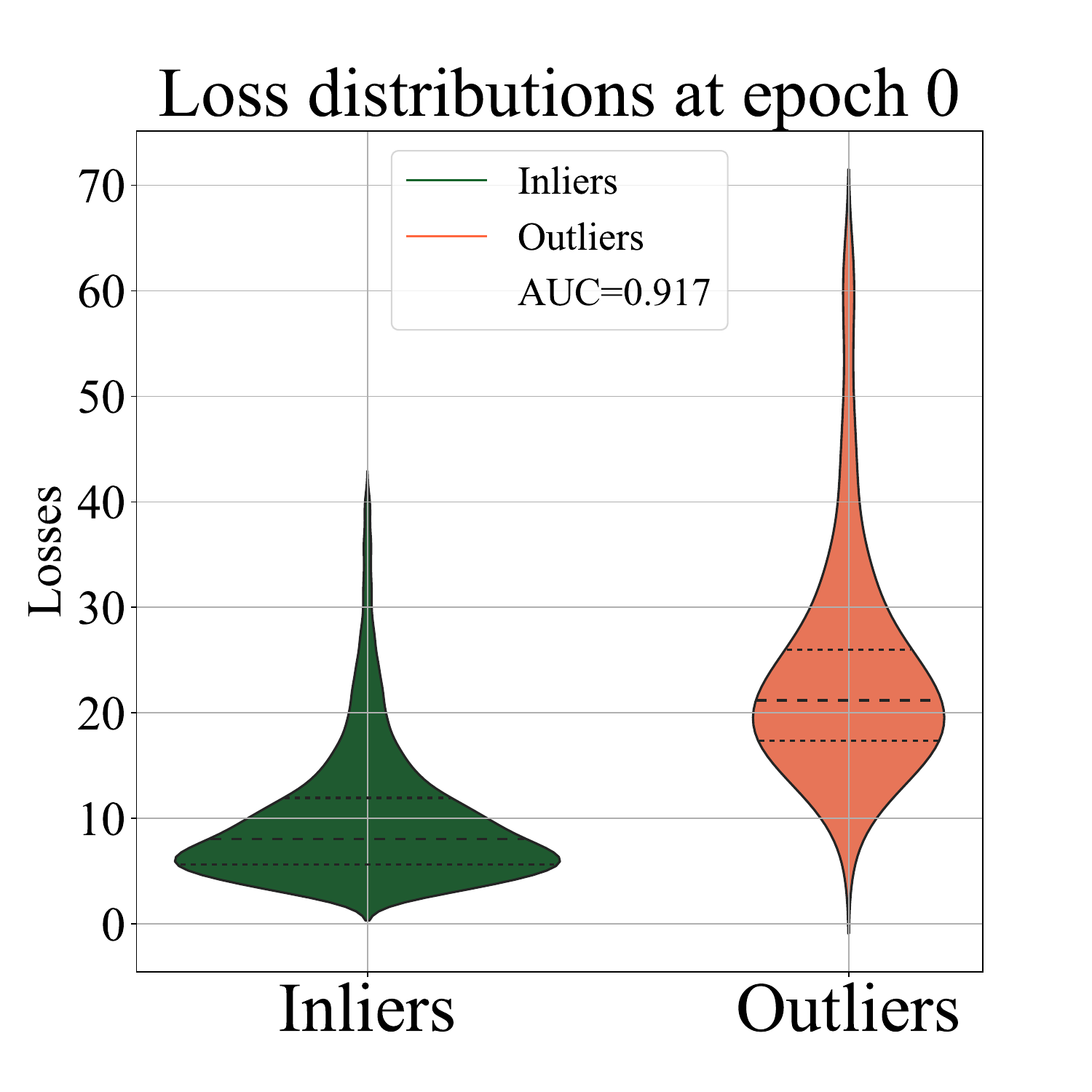}
        \caption{$t=0$, $AUC=0.917$}
        \label{fig:loss-distribution-0}
    \end{subfigure}
    \begin{subfigure}{0.27\textwidth}
        \centering
        \includegraphics[width=\textwidth]{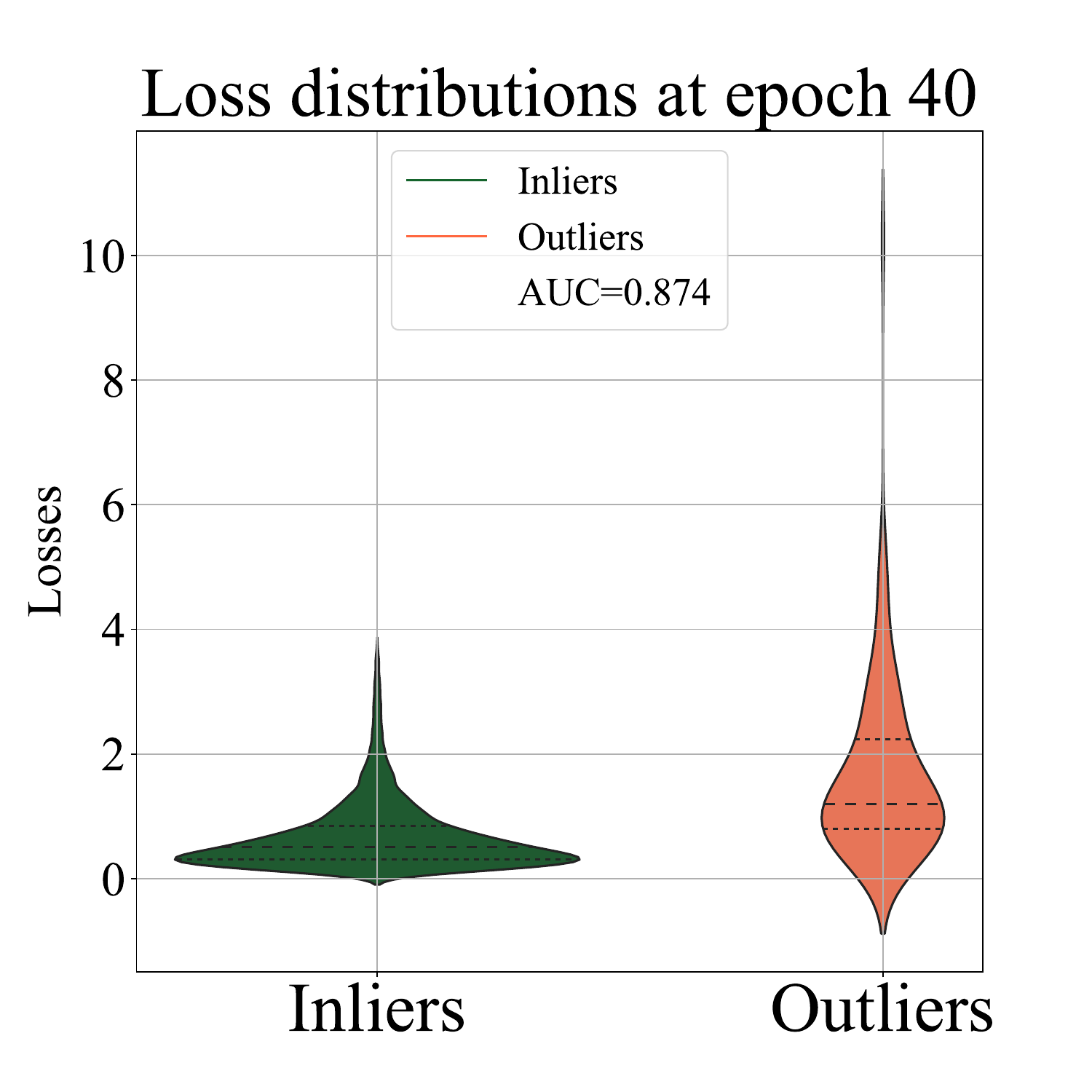}
        \caption{$t=40$, $AUC=0.874$}
        \label{fig:loss-distribution-40}
    \end{subfigure}
    \begin{subfigure}{0.27\textwidth}
        \centering
        \includegraphics[width=\textwidth]{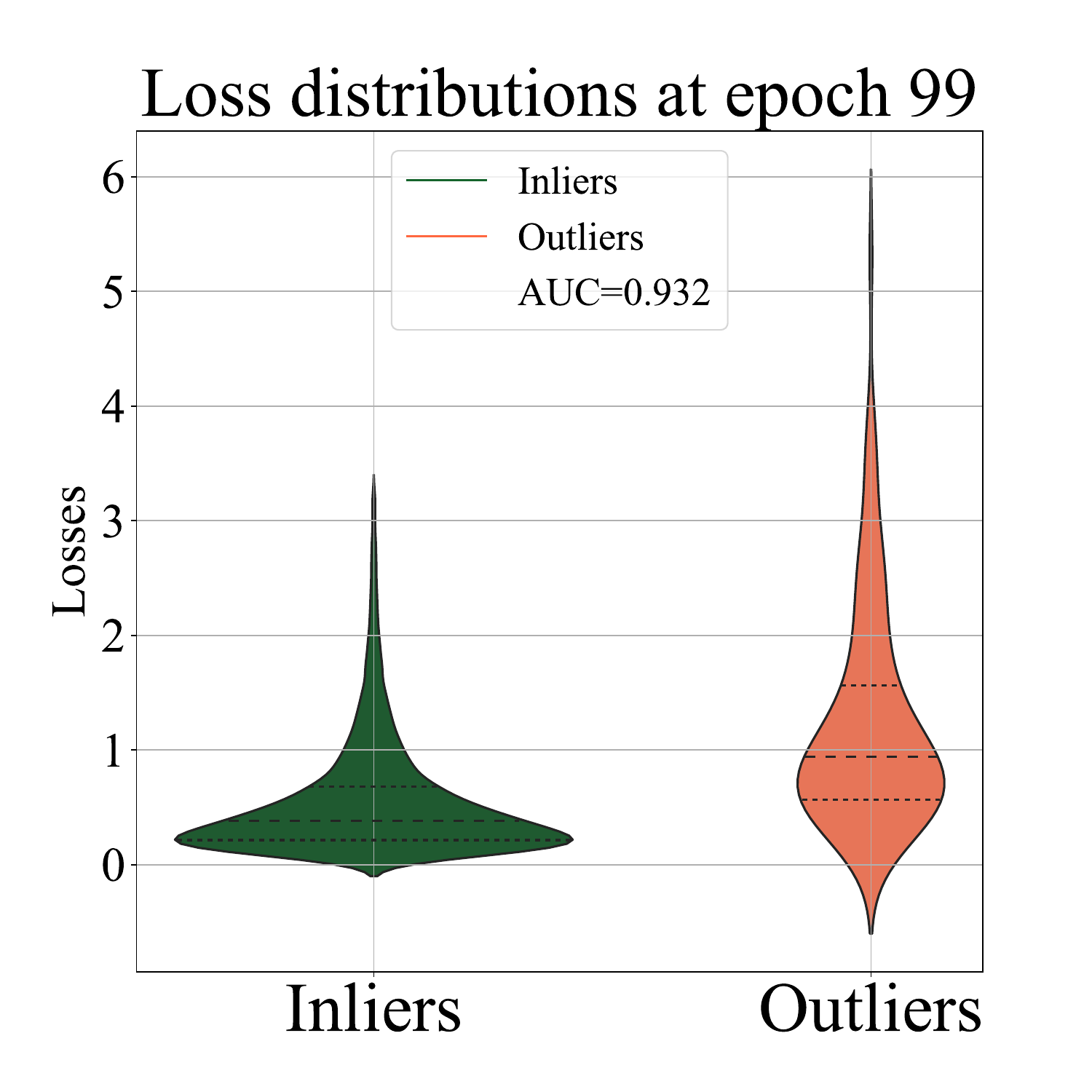}
        \caption{$t=99$, $AUC=0.932$}
        \label{fig:loss-distribution-99}
    \end{subfigure}

    \caption{Training dynamics of AE training on dataset \textit{cover}, in which outlier proportion is $0.96\%$. Dark green denotes the inliers, and orange denotes the outliers.}
    \label{fig:training-dynamics}
\vskip -0.1in
\end{figure*}

In this section, we elucidate the methodology of the proposed early stopping method, \textbf{\textit{GradStop}}. First, we explain the connection between training dynamics and the final outlier detection goal from both intuitive and theoretical perspectives, showing its fundamental mechanism. Subsequently, we introduce a gradient-based sample method that can retrieve two small sets of training samples—one is more likely to contain inliers and the other outliers—without any label. Then, a gradient-based metric is calculated upon the two sets to evaluate their inner cohesion and inter-divergence, reflecting the model's OD performance. Finally, we introduce GradStop by utilizing an automated early stopping algorithm with our novel sample method and metrics.

\subsection{Connection between Training Dynamics and Outlier Detection Goal}\label{metho-connection}
As mentioned, there is a misalignment between the optimization goal and the OD goal in deep UOD. The optimization of the OD goal is grounded in outlier assumptions, which can be reflected through training dynamics during training. Therefore, by analyzing training dynamics, one can infer OD performance's variation trend in the current training stage.

A widely used and core outlier assumption in UOD is inlier priority, for AE which posits that outliers are harder for the model to reconstruct during training. An illustrative example shown in Fig. \ref{fig:training-dynamics} describes training dynamics of AE training on dataset \textit{cover}. The upper right and lower plots witness that the mean of the outlier loss distribution is consistently larger than that of the inlier, aligning with the outlier assumption. This assumption holds in most OD conditions due to the differences in normal and abnormal data distribution, including:
\begin{itemize}
    \item \textbf{Gap in quantity and cohesion degree.} Inliers are typically more numerous and exhibit better cohesion, leading to more consistent gradient directions.
    \item \textbf{Inherent difficulty in reconstructing outliers.} Empirically, outliers are inherently harder to reconstruct. They often yield larger losses and gradient values than inliers from the beginning of training.
    \item \textbf{Irregular distribution of outliers.} Outliers tend to obey less predictable distributions, making them harder to fit during training. This irregularity results in poor gradient cohesion, i.e., inconsistent gradient directions of outliers.
\end{itemize}
In the case of \textit{cover}, outliers only take up $0.96\%$ of the overall dataset. This makes the model utilize far more gradients from inlier values than those from outliers during parameter updates, with inlier gradients exhibiting stronger cohesion and more consistent direction. As a result, the total gradient magnitude for inlier values is often larger, which can be empirically verified in the middle plot of Fig. \ref{fig:training-dynamics-sub1}. At the early stages of training AE on \textit{cover}, inliers almost dominate the parameter updates. However, the \textit{average} gradient magnitude of outliers often significantly exceeds that of inliers during the whole training process, shown in the right plot of Fig. \ref{fig:training-dynamics-sub1}. This phenomenon is due to, first, the inherent difficulty in reconstructing outliers; second, outliers less frequently obey a specific, regular distribution, leading to their gradients being less aligned directionally, and thus counteract each other, making it harder for the model to fit (thus reduce) them during training. Similar training dynamics are also observed in learning in class imbalance environments\cite{analysis-class-imbalance}. Since OD can also be viewed as a two-class classification task with two classes severely imbalanced \cite{survey-elm}, these training dynamics occur reasonably.

Therefore, on the individual gradient level, the initial values of outliers are often larger; on the whole distribution level, outliers are more difficult to fit during training. Thus, heuristically, throughout the entire training process, the individual gradients of outliers remain larger than those of inliers, as observed in the right plot of Fig. \ref{fig:training-dynamics-sub1}. This internal process is reflected in the loss level as inlier priority, as shown in the left plot of Fig. \ref{fig:training-dynamics-sub1}.

Furthermore, we theoretically demonstrate the basic mechanism of inlier priority and its relationship with the training dynamics. Specifically, when the cohesion of the sample gradients of inliers and outliers meets certain conditions, we can derive a lower bound for the difference in the rate of loss decrease between inliers and outliers. This ensures the occurrence of inlier priority, thus ensuring the effectiveness of the OD task.

The theoretical analysis in Appendix \ref{proof} provides a solid foundation for understanding how and why inlier priority, which is essential for the effectiveness of the UOD task, can be maintained during the training process and observed with training dynamics. It also supports the practical utility of the following metrics and methods proposed in our GradStop algorithm. Here we only bring up Theorem \ref{inlier-priority-theorem}:

\begin{theorem}\label{inlier-priority-theorem}
With certain assumptions and $r_t > \cos{\theta_t}R + \sqrt{\cos^2{\theta_t}R^2 + 2R + 1}$, we have loss decreasing speed gap $\tilde{\triangle}^f_t>0$, which means inlier priority strengthens at epoch $t$.
\end{theorem}
For the details of the assumptions, notations, and further explanation, please refer to Appendix \ref{proof}.

\subsection{Gradient-based sample method GradSample}
In this subsection, we propose a simple yet effective sampling method \textit{GradSample} as part of GradStop. Without any label, GradSample samples two small sets of size $k$ from the training data points in each training epoch, where the samples are respectively more likely to be inliers and outliers, reflecting label information on a collective level. After each epoch, the gradients of the samples in an evaluation batch $B_{eval}$ are extracted to calculate their magnitudes, and gradient vectors with the largest-$k$ and smallest-$k$ magnitudes are selected to form two sets, $G_{i}^{\text{top}}$ and $G_{i}^{\text{last}}$, where $i$ denotes the epoch number. Consequently, as observed in \ref{metho-connection}, each data point corresponding to a gradient vector in \( G_{i}^{\text{top}} \) \textbf{has a higher probability of being an outlier, and vice versa}. Given the model at $t$-th epoch $M_t$, evaluation batch $B_{eval}$ and sample number $k$, we have:
\begin{equation}
    G_{t}^{\text{top}}, G_{t}^{\text{last}} = \mathbf{GradSample}(M_t, B_{eval}, k) 
\end{equation}
$\mathbf{GradSample}(M_t, B_{eval}, k)$ is detailed in Appendix \ref{Appx:Algorithm}.

\subsection{Gradient-based Metric: Cohesion and Divergence}
With these two sets, $G_{t}^{\text{top}}$ and $G_{t}^{\text{last}}$, we can gain insights into the current training dynamics of the model by analyzing them, to check whether it still aligns with the outlier assumptions. Since the sets do not strictly reflect the ground-truth labels, we cannot compute based on labels for each gradient vector; however, because they exhibit distribution tendencies similar to inliers and outliers on a collective level, we can adopt measures to reflect the overall characteristics of them. Furthermore, these metrics should be capable of reflecting the outlier assumptions. Therefore, we have designed two metrics to respectively reflect the \textbf{inner cohesion} and \textbf{inter-divergence} of the inlier and outlier classes. These are determined based on the distribution assumptions of inliers and outliers: the gradients of inliers should exhibit better inner cohesion, thereby facilitating better learning of inliers. Beneficial parameter updates should demonstrate this, proving that the model can effectively exhibit inlier priority to benefit the OD task. Also, the overall gradient direction of inliers and outliers can reflect the features currently learned by the model: if the directions are relatively consistent, it indicates that the model is learning the common features of inliers and outliers; if the directions are relatively inconsistent, it suggests that the loss of inliers and outliers cannot be reduced simultaneously, indicating that the model has already reached the ceiling. The contribution to the OD task is much smaller in the latter phase compared to the former.

Given a set of gradient vectors $G=\{g_1, \cdots, g_k\}$, the cohesion metric $\mathbf{C}$ is defined in equation \ref{cohesion}.

\begin{equation}\label{cohesion}
    \mathbf{C}(G) = \frac{\|\Sigma_1^k g_i\|}{\Sigma_1^k\|g_i\|} \in [0,1]
\end{equation}
Specifically, when all vectors in $G$ completely counteract each other, $\mathbf{C}(G)$ equals 0; when all vectors in $G$ are in the same direction, $\mathbf{C}(G)$ equals 1.

Given two sets of gradient vectors $G^{1}=\{g^1_1, \cdots, g^1_k\}$ and $G^{2}=\{g^2_1, \cdots, g^2_k\}$. The divergence metric $\mathbf{D}$ is defined in equation \ref{divergence}.
\begin{equation}\label{divergence}
    \mathbf{D}(G^{1}, G^{2}) = \theta_t=\angle(\Sigma_1^k g^1_i, \Sigma_1^k g^2_i)
\end{equation}
in which $\angle(u, v)$ means the angle value between vectors $u$ and $v$. $\mathbf{C}$ and $\mathbf{D}$ are also associated with inlier priority theoretically according to Appendix \ref{proof}.

\RestyleAlgo{ruled}
\SetKwComment{Comment}{/* }{ */}
\SetKwInput{kwInput}{Input}
\SetKwInput{kwOutput}{Output}
\SetKwInput{kwReturn}{Return}
\SetKw{Break}{break}


\subsection{GradStop: Automated Early Stopping Algorithm}
Based on the two metrics, we design \textit{GradStop}, an early stopping algorithm that can dynamically assess whether the current training state of the model aligns with the outlier assumptions through the lens of training dynamics. If not, it is inferred that the model has reached its optimal performance on the OD task, and training should be halted to prevent toxic training caused by goal misalignment.

We determine stopping with a sliding window mechanism. Window size $w$ is similar to the patience parameter for searching the optimal iteration in \cite{EntropyStop}, and $R_{down}$ sets the requirement for the smooth degree of downtrend. A larger $w$ usually improves accuracy at the expense of longer training time. GradStop does the following:
\begin{itemize}
    \item At the beginning of training, check if $\mathbf{D}(G_{\text{top}}, G_{\text{last}})$ is large ($>t_{\mathbf{D}}$), if so, choose the initial model parameter. This suggests that inliers and outliers are inherently distinguished by the model with its model structure, and probably learning is useless for certain algorithm-dataset pairs. In this circumstance, performance is likely to degrade consistently. The random initialized model may exhibit the best performance. This phenomenon is empirically observed when training AE and DeepSVDD on many datasets, as training DeepSVDD on \textit{vowels} in Fig.\ref{Fig:loss-auc-alg}
    \item At every training epoch, calculate $\mathbf{C_\Delta} = \mathbf{C}(G_{\text{last}}) - \mathbf{C}(G_{\text{top}})$. If it is small ($<t_{\mathbf{C}b}$) and has not been increasing for $w$ epochs, indicating useless learning is going on, halt the training. 
\end{itemize}

Furthermore, a significance threshold $t_{\mathbf{C}s}$ is introduced to deal with the cold start problem. In some cases, inlier priority in which GradSample is grounded needs some training to exhibit. We apply an optimistic mechanism that deems training beneficial by default. Only when an evident disparity exists between $\mathbf{C}(G_{\text{last}})$ and $\mathbf{C}(G_{\text{top}})$, i.e. $|\mathbf{C_\Delta}|>t_{\mathbf{C}s}$, the algorithm further decides whether training is useless. The algorithm is detailed in Appendix \ref{Appx:Algorithm}. With GradStop, we can terminate training when the outlier assumptions no longer hold and select the epoch that the training dynamics best conform to the outlier assumptions, improving the model's final performance on the OD task.

\section{Experiments}
\definecolor{red-c}{rgb}{1, 0, 0}
\definecolor{blue-c}{rgb}{0, 0, 1}
\newcommand\best[1]{\textbf{\textcolor{red-c}{#1}}}
\newcommand\second[1]{\underline{{#1}}}

The experiment consists of: experiment settings, GradAE performance, and improvements on other deep UOD models. In GradAE performance, we apply GradStop to the AutoEncoder (AE) model, a widely used deep UOD method, to gain GradAE, achieving comparable performance with other SOTA UOD baselines. Then, improvements on other deep UOD Models are evaluated to show the generalization ability and robustness of GradStop along with the limitation of the early stopping approach.

\subsection{Experiment Settings}
All experiments adopt a transductive setting, where the training set equals the test set, which is common in Unsupervised OD \cite{robod,randnet}. 

\textbf{Dataset.}
Experiments are carried out on 47 widely-used real-world tabular datasets\footnote{https://github.com/Minqi824/ADBench/} collected by \cite{ADbench}, which cover many application domains, including healthcare, image processing, finance, etc. The details of the 47 datasets are shown in Appendix \ref{Appx:Exp}. Our codes are available at 
\url{https://anonymous.4open.science/r/gradAE-879F}.

\textbf{Evaluation Metrics.}
We evaluate performance with AUC, a widely used evaluation metric in the field of OD, defined in Equation \ref{eq: auc}. Computing AUC does not need any threshold for outlier scores outputted by the model, as it is a ranking-based metric. 

\textbf{Computing Infrastructures.}
All experiments are conducted on 12th Gen Intel(R) Core(TM) i5-12400F CPU, and NVIDIA GeForce RTX 3060 Ti (8GB GPU memory) GPU, CUDA Version 12.2.

\subsection{GradAE Performance Study}
We first study how much improvement can be achieved by employing GradStop on AE model. VanillaAE denotes the simplest form of AE without any additional technique. We apply our early stopping method to VanillaAE to gain \textbf{\textit{GradAE}}. Then, GradAE is compared with two ensemble AEs, the recent SOTA hyper-ensemble ROBOD \cite{robod} and the widely-used RandNet \cite{randnet}.  
We also choose EntropyAE with EntropyStop\cite{EntropyStop} as one of the baselines, since it is the pioneering work applying early stopping to deep UOD. Another family of UOD is traditional methods including IF \cite{IsolationForest}, ECOD \cite{ecod}, KNN \cite{knn}, CBLOF \cite{CBLOF} and GMM \cite{dagmm}. Compared to deep methods, traditional methods may lack some potential for generalization and dealing with high-dimensional data, but still, they are very competitive in UOD and perform well on certain datasets.

\subsubsection{\textbf{Detection Performance Result on AE}}
The result is in Table \ref{tab:AE-ensemble-compare} and details can be found in Appendix \ref{Appx:exp-details}. GradAE significantly outperforms VanillaAE and also gains superior performance over other SOTA methods even ensemble AE. For each dataset, we ranked the AUC performance of the ten algorithms and listed the average ranking of each algorithm. GradStop greatly improved VanillaAE's score from 0.758 to 0.775 by approximately 2.25\% and VanillaAE's average ranking from a nearly bottom 5.809 to a second 5.021 across 10 UOD methods. By employing early stopping, GradAE effectively mitigates the problem of goal misalignment thus improving the overall performance.


\begin{table}[tb]
  \centering
  \caption{Detection performance of models from AE family and other SOTA UOD methods. }
    \vskip 0.15in
    \resizebox{0.8\columnwidth}{!}{
        \begin{tabular}{c|c|c|c|c|c}
        \toprule
        \toprule
        \multicolumn{1}{r}{} & \multicolumn{5}{c}{\textbf{Family of AE models}} \\
        \midrule
              & VanillaAE & GradAE (Ours)  & EntropyAE & RandNet & ROBOD \\
        \midrule
        $\overline{AUC}$   & 0.758$\pm$0.004 & \best{0.775$\pm$0.003} & \second{0.769$\pm$0.005} & 0.736$\pm$0.00 &  0.744$\pm$0.00 \\
        \midrule
        $\overline{Rank}_{AUC}$ & 5.809    & \second{5.021}     & 5.043     & 5.574     & 5.638 \\
        \midrule
        \midrule
        \multicolumn{1}{c}{} & \multicolumn{5}{c}{\textbf{Traditional UOD methods}} \\
        \midrule
              & Isolation Forest & ECOD  & KNN   & CBLOF & GMM \\
        \midrule
        $\overline{AUC}$   & 0.762$\pm$0.00 & 0.742$\pm$0.00 & 0.720$\pm$0.00 & 0.748$\pm$0.00 & 0.758$\pm$0.00 \\
        \midrule
        $\overline{Rank}_{AUC}$ & \best{4.872}	& 5.872	& 5.894	& 5.489 &	5.532 \\
        \midrule
        \bottomrule
        \end{tabular}%
    }
  \label{tab:AE-ensemble-compare}%
\vspace{-5mm}
\end{table}%





\subsection{Improvements on other Deep UOD Models}
In this subsection, we apply GradStop to other Deep UOD models, including VAE \cite{vae}, DeepSVDD \cite{deep-svdd}, and RDP \cite{RDP} to validate its generalization capability. Experiments show that GradStop significantly enhances the performance of unsupervised DeepSVDD, followed by AE and RDP, with VAE showing the least improvement.

\begin{table}[tb]
  \centering
  \caption{Detection performance of deep UOD models and their GradStop and EntropyStop versions(with respectively -G and -E suffix in the model name). Patience parameters are set to $20$.}
  \vskip 0.15in
  \resizebox{0.8\columnwidth}{!}{
    \begin{tabular}{c|c|c|c|c|c}
    \toprule
    \toprule
 & \multicolumn{3}{c|}{\textbf{Model Performance}} & \multicolumn{2}{c}{\textbf{Improvements}} \\
    \midrule
          & DeepSVDD & DeepSVDD-G & DeepSVDD-E & \multicolumn{2}{c}{DeepSVDD-G} \\
    \midrule
    AUC   & 0.502 & \best{0.648} & \second{0.51}  & 29.08\% & 38/47 \\
    \midrule
          & AE & GradAE & AE-E & \multicolumn{2}{c}{GradAE} \\
    \midrule
    AUC   & 0.758 & \best{0.775} & \second{0.769}  & 2.25\% & 24/47 \\
    \midrule
          & RDP   & RDP-G & RDP-E & \multicolumn{2}{c}{RDP-G} \\
    \midrule
    AUC   & 0.742 & \best{0.747} & \second{0.744} & 0.67\% & 22/47 \\
    \midrule
          & VAE   & VGradAE & VAE-E & \multicolumn{2}{c}{VGradAE} \\
    \midrule
    AUC   & \second{0.746} & \best{0.747} & \second{0.746} & 0.13\% & 17/47 \\
    \bottomrule
    \bottomrule
    \end{tabular}%
    }
  \label{tab:uod-models-performance}%
\vskip -0.2in
\end{table}%

\subsubsection{\textbf{Detection Performance Result on deep UOD models}}
The result of the experiment is in Table \ref{tab:uod-models-performance}, listing 12 models, DeepSVDD, AE, RDP, and VAE along with their GradStop and EntropyStop \cite{EntropyStop} versions. ``Improvements'' denotes the overall percentage of improvement and number of datasets on which GradStop successfully mitigates performance degradation among all 47.

From Table \ref{tab:uod-models-performance}, we can see that GradStop significantly improves DeepSVDD, increasing its AUC from $0.502$ to $0.648$ by $29.08\%$, and enhancing performance on a majority 38 out of 47 of the datasets. For AE, performance on half of the datasets is improved, enabling it to outperform other SOTA methods. In contrast, the improvements on RDP and VAE are smaller, with RDP's AUC increasing from $0.742$ to $0.747$, and VAE's AUC from $0.746$ to $0.747$; respectively, performance only on 22 and 17 out of 47 datasets are improved. Additionally, compared to EntropyStop, GradStop performs better on all four deep UOD models, demonstrating its stronger generalization ability across different algorithm-dataset pairs and robustness to complex training dynamics. For details on individual datasets, please refer to Appendix \ref{Appx:Exp}.

\subsubsection{\textbf{Discussions}}
The performance improvement of the GradStop algorithm on deep UOD models decreases in the order of DeepSVDD, AE, RDP, and VAE. Following we briefly analyze the reasons.

First, AE is the simplest and generally the best-performing model. Its optimization goal is reconstruction loss, which can be affected by goal misalignment, leading to trends of AUC rising, falling, or fluctuating across different datasets. AUC decreases when AE fits better to outliers. GradStop can detect this phenomenon and halt training, thus providing a performance boost for AE. For a more intuitive and detailed understanding, please refer to the case study in Appendix \ref{exp-case-study}.

DeepSVDD is more susceptible to the impact of outliers compared to AE, making it more prone to goal misalignment and performance degradation. This is due to its optimization goal, which minimizes the hypersphere radius in the latent space that encloses all data point representations. As a result, the quantitative advantage of inliers is partly weakened, and the model is more likely to learn outliers far from the center of the hypersphere. 

On the other hand, VAE and RDP impose stronger constraints on the latent space. VAE enforces a regularization constraint to make the latent variables follow a normal distribution, while RDP specifies an additional distance function and requires the distance space of the latent vectors to fit a randomly projected distance space. These constraints help mitigate the problem of goal misalignment since the distribution pattern of representations is fixed. Performance degradation of RDP and VAE is only observed in several datasets. Therefore, the early stopping mechanism loses its room for maneuvering. However, these constraints also limit the model's representation capacity, making it harder for the model to flexibly fit complex inliers, leading to performances (0.747 for both RDP and VAE) lower than and GradAE (0.775), which imposes no constraints on the representation space. 


\section{Conclusion}

In deep Unsupervised Outlier Detection (UOD), there exists an inherent misalignment between the model optimization goal and the goal of the UOD task due to the lack of label guidance. In this work, inspired by the characteristics of outlier distributions and inlier priority\cite{inlier-priority}, we elucidate the connection between the UOD goal, outlier assumptions, and UOD training dynamics empirically and theoretically. A label-free gradient-based sampling method \textbf{\textit{GradSample}} and two metrics, inner cohesion and inter-divergence, are designed to measure the outlier assumptions during training. \textbf{\textit{GradStop}} is designed to halt the training before the outlier assumptions are violated. Experiments show that applying GradStop on AutoEncoder can significantly enhance its performance, surpassing existing SOTA UOD baselines. Furthermore, GradStop applies to different algorithm-dataset pairs and handles various training dynamics. Experiments on DeepSVDD, RDP, and VAE indicate that it can effectively mitigate performance degradation on various dataset-algorithm pairs.

Future work involves exploring the potential training dynamics, especially in unsupervised scenarios where label guidance is unavailable. GradSample might be used to generate pseudo labels that could transform unsupervised learning into weakly supervised learning. Moreover, a better approach is to integrate the proposed metrics into the optimization goal rather than as auxiliary early-stopping metrics, since early stopping meets its limitation when the representations space is constrained. Further delving into UOD training dynamics can help us understand the mechanism of deep UOD methods better.

\bibliographystyle{icml2024}
\bibliography{GradStop}

\begin{thebibliography}{40}
\providecommand{\natexlab}[1]{#1}
\providecommand{\url}[1]{\texttt{#1}}
\expandafter\ifx\csname urlstyle\endcsname\relax
  \providecommand{\doi}[1]{doi: #1}\else
  \providecommand{\doi}{doi: \begingroup \urlstyle{rm}\Url}\fi

\bibitem[Arpit et~al.(2017)Arpit, Jastrz{\k{e}}bski, Ballas, Krueger, Bengio, Kanwal, Maharaj, Fischer, Courville, Bengio, et~al.]{ES_noise_3}
Arpit, D., Jastrz{\k{e}}bski, S., Ballas, N., Krueger, D., Bengio, E., Kanwal, M.~S., Maharaj, T., Fischer, A., Courville, A., Bengio, Y., et~al.
\newblock A closer look at memorization in deep networks.
\newblock In \emph{International conference on machine learning}, pp.\  233--242. PMLR, 2017.

\bibitem[Bai et~al.(2021)Bai, Yang, Han, Yang, Li, Mao, Niu, and Liu]{ES_noise_1}
Bai, Y., Yang, E., Han, B., Yang, Y., Li, J., Mao, Y., Niu, G., and Liu, T.
\newblock Understanding and improving early stopping for learning with noisy labels.
\newblock In \emph{Advances in Neural Information Processing Systems}, volume~34, pp.\  24392--24403, 2021.

\bibitem[Bradley(1997)]{auc}
Bradley, A.~P.
\newblock The use of the area under the roc curve in the evaluation of machine learning algorithms.
\newblock \emph{Pattern recognition}, 30\penalty0 (7):\penalty0 1145--1159, 1997.

\bibitem[Breunig et~al.(2000)Breunig, Kriegel, Ng, and Sander]{LOF}
Breunig, M.~M., Kriegel, H.-P., Ng, R.~T., and Sander, J.
\newblock Lof: identifying density-based local outliers.
\newblock In \emph{Proceedings of the 2000 ACM SIGMOD international conference on Management of data}, pp.\  93--104, 2000.

\bibitem[Chalapathy \& Chawla(2019)Chalapathy and Chawla]{deep-od-survey-2019}
Chalapathy, R. and Chawla, S.
\newblock Deep learning for anomaly detection: A survey.
\newblock \emph{arXiv preprint arXiv:1901.03407}, 2019.

\bibitem[Chandola et~al.(2009)Chandola, Banerjee, and Kumar]{od-survey}
Chandola, V., Banerjee, A., and Kumar, V.
\newblock Anomaly detection: A survey.
\newblock \emph{ACM computing surveys (CSUR)}, 41\penalty0 (3):\penalty0 1--58, 2009.

\bibitem[Chen et~al.(2017)Chen, Sathe, Aggarwal, and Turaga]{randnet}
Chen, J., Sathe, S., Aggarwal, C., and Turaga, D.
\newblock Outlier detection with autoencoder ensembles.
\newblock In \emph{Proceedings of the 2017 SIAM international conference on data mining}, pp.\  90--98. SIAM, 2017.

\bibitem[Ding et~al.(2024)Ding, Zhao, and Akoglu]{robod}
Ding, X., Zhao, L., and Akoglu, L.
\newblock Hyperparameter sensitivity in deep outlier detection analysis and a scalable hyper-ensemble solution.
\newblock In \emph{Proceedings of the 36th International Conference on Neural Information Processing Systems}, NIPS '22, 2024.

\bibitem[Dou et~al.(2020)Dou, Liu, Sun, Deng, Peng, and Yu]{financial-example}
Dou, Y., Liu, Z., Sun, L., Deng, Y., Peng, H., and Yu, P.~S.
\newblock Enhancing graph neural network-based fraud detectors against camouflaged fraudsters.
\newblock In \emph{Proceedings of the 29th ACM International Conference on Information \& Knowledge Management}, pp.\  315--324, 2020.

\bibitem[Francazi et~al.(2023)Francazi, Baity-Jesi, and Lucchi]{analysis-class-imbalance}
Francazi, E., Baity-Jesi, M., and Lucchi, A.
\newblock A theoretical analysis of the learning dynamics under class imbalance.
\newblock In Krause, A., Brunskill, E., Cho, K., Engelhardt, B., Sabato, S., and Scarlett, J. (eds.), \emph{Proceedings of the 40th International Conference on Machine Learning}, volume 202 of \emph{Proceedings of Machine Learning Research}, pp.\  10285--10322. PMLR, 2023.

\bibitem[Han et~al.(2022)Han, Hu, Huang, Jiang, and Zhao]{ADbench}
Han, S., Hu, X., Huang, H., Jiang, M., and Zhao, Y.
\newblock Adbench: Anomaly detection benchmark.
\newblock \emph{arXiv preprint arXiv:2206.09426}, 2022.

\bibitem[Hawkins(1980)]{uod-definition}
Hawkins, D.~M.
\newblock \emph{Identification of outliers}, volume~11.
\newblock Springer, 1980.

\bibitem[He et~al.(2003)He, Xu, and Deng]{CBLOF}
He, Z., Xu, X., and Deng, S.
\newblock Discovering cluster-based local outliers.
\newblock \emph{Pattern Recogn. Lett.}, 24\penalty0 (9–10):\penalty0 1641–1650, 2003.

\bibitem[Huang et~al.(2024)Huang, Zhang, Wang, Zhang, and Lin]{EntropyStop}
Huang, Y., Zhang, Y., Wang, L., Zhang, F., and Lin, X.
\newblock Entropystop: Unsupervised deep outlier detection with loss entropy.
\newblock In \emph{Proceedings of the 30th ACM SIGKDD Conference on Knowledge Discovery and Data Mining}, 2024.

\bibitem[Kiani et~al.(2024)Kiani, Jin, and Sheng]{survey-elm}
Kiani, R., Jin, W., and Sheng, V.~S.
\newblock Survey on extreme learning machines for outlier detection.
\newblock \emph{Mach. Learn.}, 2024.

\bibitem[Kim et~al.(2024)Kim, Hwang, Lee, Kim, and Kim]{inlier-memorization}
Kim, D., Hwang, J., Lee, J., Kim, K., and Kim, Y.
\newblock Odim: outlier detection via likelihood of under-fitted generative models.
\newblock In \emph{Proceedings of the 41st International Conference on Machine Learning}, ICML'24. JMLR.org, 2024.

\bibitem[Kingma \& Welling(2022)Kingma and Welling]{vae}
Kingma, D.~P. and Welling, M.
\newblock Auto-encoding variational bayes, 2022.
\newblock URL \url{https://arxiv.org/abs/1312.6114}.

\bibitem[Lai et~al.(2021)Lai, Zha, Xu, Zhao, Wang, and Hu]{Ts-benchmark}
Lai, K.-H., Zha, D., Xu, J., Zhao, Y., Wang, G., and Hu, X.
\newblock Revisiting time series outlier detection: Definitions and benchmarks.
\newblock In \emph{Thirty-fifth Conference on Neural Information Processing Systems Datasets and Benchmarks Track (Round 1)}, 2021.

\bibitem[Li et~al.(2020)Li, Soltanolkotabi, and Oymak]{ES_noise_0}
Li, M., Soltanolkotabi, M., and Oymak, S.
\newblock Gradient descent with early stopping is provably robust to label noise for overparameterized neural networks.
\newblock In \emph{International conference on artificial intelligence and statistics}, pp.\  4313--4324. PMLR, 2020.

\bibitem[Li et~al.(2022)Li, Zhao, Hu, Botta, Ionescu, and Chen]{ecod}
Li, Z., Zhao, Y., Hu, X., Botta, N., Ionescu, C., and Chen, G.
\newblock Ecod: Unsupervised outlier detection using empirical cumulative distribution functions.
\newblock \emph{IEEE Transactions on Knowledge and Data Engineering}, 2022.

\bibitem[Liu et~al.(2008)Liu, Ting, and Zhou]{IsolationForest}
Liu, F.~T., Ting, K.~M., and Zhou, Z.-H.
\newblock Isolation forest.
\newblock In \emph{2008 eighth ieee international conference on data mining}, pp.\  413--422. IEEE, 2008.

\bibitem[Liu et~al.(2022)Liu, Dou, Zhao, Ding, Hu, Zhang, Ding, Chen, Peng, Shu, et~al.]{god-benchmark}
Liu, K., Dou, Y., Zhao, Y., Ding, X., Hu, X., Zhang, R., Ding, K., Chen, C., Peng, H., Shu, K., et~al.
\newblock Bond: Benchmarking unsupervised outlier node detection on static attributed graphs.
\newblock In \emph{Thirty-sixth Conference on Neural Information Processing Systems Datasets and Benchmarks Track}, 2022.

\bibitem[Liu et~al.(2019)Liu, Li, Zhou, Jiang, Sun, Wang, and He]{gan-ensemble}
Liu, Y., Li, Z., Zhou, C., Jiang, Y., Sun, J., Wang, M., and He, X.
\newblock Generative adversarial active learning for unsupervised outlier detection.
\newblock \emph{IEEE Transactions on Knowledge and Data Engineering}, 32\penalty0 (8):\penalty0 1517--1528, 2019.

\bibitem[Morgan \& Bourlard(1989)Morgan and Bourlard]{ES_1st}
Morgan, N. and Bourlard, H.
\newblock Generalization and parameter estimation in feedforward nets: Some experiments.
\newblock In Touretzky, D. (ed.), \emph{Advances in Neural Information Processing Systems}, volume~2. Morgan-Kaufmann, 1989.

\bibitem[Pang et~al.(2021)Pang, Shen, Cao, and Hengel]{deep-od-survey-2021}
Pang, G., Shen, C., Cao, L., and Hengel, A. V.~D.
\newblock Deep learning for anomaly detection: A review.
\newblock \emph{ACM Computing Surveys (CSUR)}, 54\penalty0 (2):\penalty0 1--38, 2021.

\bibitem[Qiu et~al.(2021)Qiu, Pfrommer, Kloft, Mandt, and Rudolph]{NTL}
Qiu, C., Pfrommer, T., Kloft, M., Mandt, S., and Rudolph, M.
\newblock Neural transformation learning for deep anomaly detection beyond images.
\newblock In \emph{International Conference on Machine Learning}, pp.\  8703--8714. PMLR, 2021.

\bibitem[Qiu et~al.(2022)Qiu, Li, Kloft, Rudolph, and Mandt]{LOE}
Qiu, C., Li, A., Kloft, M., Rudolph, M., and Mandt, S.
\newblock Latent outlier exposure for anomaly detection with contaminated data.
\newblock In \emph{International Conference on Machine Learning}, pp.\  18153--18167. PMLR, 2022.

\bibitem[Ramaswamy et~al.(2000)Ramaswamy, Rastogi, and Shim]{knn}
Ramaswamy, S., Rastogi, R., and Shim, K.
\newblock Efficient algorithms for mining outliers from large data sets.
\newblock In \emph{Proceedings of the 2000 ACM SIGMOD international conference on Management of data}, pp.\  427--438, 2000.

\bibitem[Ruff et~al.(2018)Ruff, Vandermeulen, Goernitz, Deecke, Siddiqui, Binder, M{\"u}ller, and Kloft]{deep-svdd}
Ruff, L., Vandermeulen, R., Goernitz, N., Deecke, L., Siddiqui, S.~A., Binder, A., M{\"u}ller, E., and Kloft, M.
\newblock Deep one-class classification.
\newblock In \emph{International conference on machine learning}, pp.\  4393--4402. PMLR, 2018.

\bibitem[Ruff et~al.(2021)Ruff, Kauffmann, Vandermeulen, Montavon, Samek, Kloft, Dietterich, and M{\"u}ller]{deep-od-survey-3}
Ruff, L., Kauffmann, J.~R., Vandermeulen, R.~A., Montavon, G., Samek, W., Kloft, M., Dietterich, T.~G., and M{\"u}ller, K.-R.
\newblock A unifying review of deep and shallow anomaly detection.
\newblock \emph{Proceedings of the IEEE}, 109\penalty0 (5):\penalty0 756--795, 2021.

\bibitem[Schlegl et~al.(2017)Schlegl, Seeb{\"o}ck, Waldstein, Schmidt-Erfurth, and Langs]{AnoGAN}
Schlegl, T., Seeb{\"o}ck, P., Waldstein, S.~M., Schmidt-Erfurth, U., and Langs, G.
\newblock Unsupervised anomaly detection with generative adversarial networks to guide marker discovery.
\newblock In \emph{International conference on information processing in medical imaging}, pp.\  146--157. Springer, 2017.

\bibitem[Shenkar \& Wolf(2021)Shenkar and Wolf]{ICL}
Shenkar, T. and Wolf, L.
\newblock Anomaly detection for tabular data with internal contrastive learning.
\newblock In \emph{International Conference on Learning Representations}, 2021.

\bibitem[Wang et~al.(2019{\natexlab{a}})Wang, Pang, Shen, and Ma]{RDP}
Wang, H., Pang, G., Shen, C., and Ma, C.
\newblock Unsupervised representation learning by predicting random distances.
\newblock \emph{arXiv preprint arXiv:1912.12186}, 2019{\natexlab{a}}.

\bibitem[Wang et~al.(2019{\natexlab{b}})Wang, Zeng, Liu, Zhu, Yin, Xu, and Kloft]{inlier-priority}
Wang, S., Zeng, Y., Liu, X., Zhu, E., Yin, J., Xu, C., and Kloft, M.
\newblock Effective end-to-end unsupervised outlier detection via inlier priority of discriminative network.
\newblock \emph{Advances in neural information processing systems}, 32, 2019{\natexlab{b}}.

\bibitem[Weller-Fahy et~al.(2014)Weller-Fahy, Borghetti, and Sodemann]{security-example}
Weller-Fahy, D.~J., Borghetti, B.~J., and Sodemann, A.~A.
\newblock A survey of distance and similarity measures used within network intrusion anomaly detection.
\newblock \emph{IEEE Communications Surveys \& Tutorials}, 17\penalty0 (1):\penalty0 70--91, 2014.

\bibitem[Xia et~al.(2020)Xia, Liu, Han, Gong, Wang, Ge, and Chang]{ES_noise_2}
Xia, X., Liu, T., Han, B., Gong, C., Wang, N., Ge, Z., and Chang, Y.
\newblock Robust early-learning: Hindering the memorization of noisy labels.
\newblock In \emph{International conference on learning representations}, 2020.

\bibitem[Xia et~al.(2015)Xia, Cao, Wen, Hua, and Sun]{outlier_refine_2015}
Xia, Y., Cao, X., Wen, F., Hua, G., and Sun, J.
\newblock Learning discriminative reconstructions for unsupervised outlier removal.
\newblock In \emph{Proceedings of the IEEE international conference on computer vision}, pp.\  1511--1519, 2015.

\bibitem[Yoon et~al.(2021)Yoon, Sohn, Li, Arik, Lee, and Pfister]{outlier_refine_2021}
Yoon, J., Sohn, K., Li, C.-L., Arik, S.~O., Lee, C.-Y., and Pfister, T.
\newblock Self-trained one-class classification for unsupervised anomaly detection.
\newblock \emph{arXiv e-prints}, pp.\  arXiv--2106, 2021.

\bibitem[Zhou \& Paffenroth(2017)Zhou and Paffenroth]{RDA}
Zhou, C. and Paffenroth, R.~C.
\newblock Anomaly detection with robust deep autoencoders.
\newblock In \emph{Proceedings of the 23rd ACM SIGKDD international conference on knowledge discovery and data mining}, pp.\  665--674, 2017.

\bibitem[Zong et~al.(2018)Zong, Song, Min, Cheng, Lumezanu, Cho, and Chen]{dagmm}
Zong, B., Song, Q., Min, M.~R., Cheng, W., Lumezanu, C., Cho, D., and Chen, H.
\newblock Deep autoencoding gaussian mixture model for unsupervised anomaly detection.
\newblock In \emph{International conference on learning representations}, 2018.

\end{thebibliography}

\newpage
\appendix

\onecolumn

\section{Algorithm details}\label{Appx:Algorithm}
The details of algorithms GradSample and GradStop are shown in Algorithm \ref{alg:sample} and Algorithm \ref{alg:GradStop}, respectively. The related notations are in Appendix \ref{Appx:GradStopNotations}. In Algorithm \ref{alg:GradStop}, an additional threshold $t_{\mathbf{C}s}$ is applied. $|C^{\Delta}[t]|$ should be larger than a significance threshold $t_{\mathbf{C}s}$ to stop training, which assures that the model would not stop training when there is no significant difference between $\mathbf{C}(G_{\text{top}})$ and $\mathbf{C}(G_{\text{last}})$, which usually happens when current model parameter is unable to distinguish inliers and outliers effectively and further training is needed. For the configurations of the hyperparameters, please refer to Appendix \ref{exp-hp}.

\begin{algorithm}[tbhp]\label{GradSample}
\small
\caption{GradSample: An sampling algorithm based on training dynamics when training deep UOD model}
\label{alg:sample}
\kwInput{Model $M_t$ with learnable parameters $\Theta_t$, evaluation set $B_{eval}=\{x_1, \dots, x_n\}$, sampling number $k$}
\kwOutput{Two sets of gradient vectors, $G_{t}^{\text{top}}, G_{t}^{\text{last}}$}
Calculate $G = \{g_i \mid x_i \in B_{eval}\}$ with $g_i$ being the gradient vector of $M_t(x_i)$\;
Calculate $G_{norm} = \{\|g_i\| \mid g_i \in G\}$\;
$i_{last}$ = $\mathbf{argsort}(G_{norm})[1:k]$\;
$i_{top}$ = $\mathbf{argsort}(G_{norm})[n-k+1:n]$
\Comment*[r]{argsort($l$) returns indices of list $l$ in value-ascending order.}
$G^{\text{top}}=G[i_{top}]$\;
$G^{\text{last}}=G[i_{last}]$\;
\kwReturn{$G^{\text{top}}, G^{\text{last}}$}
\end{algorithm}

\begin{algorithm}[tbhp]\label{GradStop}
\small
\caption{GradStop: An early stopping algorithm for deep UOD model based on training dynamics}
\label{alg:GradStop}
\kwInput{ Model $M$ with learnable parameters $\Theta$, downtrend threshold $R_{down}$, dataset $\mathcal{D}$, iterations $T$, sampling number $k$, evaluation batch number $n$, sliding window size $w$}
\kwOutput{Outlier score list $\textbf{O}$}
Initialize the parameter $\Theta_0$ of Model $M$\; Random sample $n$ instances from $\mathcal{D}$ as the evaluation batch $B_{eval}$ \;
$\Theta_{best}\gets \Theta_0;$ $H \gets 0$ \Comment*[r]{Model Training} 
\For {$t:= 1 \rightarrow T$}{
    $\mathcal{L}_{t}=f(\Theta_{t-1}, \mathcal{D})$ \;
    $\Theta_t \gets \textbf{optimizer}(\Theta_{t-1}, \mathcal{L}_{t})$ \;
    $G_{t}^{\text{top}}, G_{t}^{\text{last}} = \mathbf{GradSample}(M_t, B_{eval}, k)$ \;
    $C^{\Delta}[t] \gets \mathbf{C}(G_{t}^{\text{last}})-\mathbf{C}(G_{t}^{\text{top}})$ \;
    $\mathbf{D}[t] = \mathbf{D}(G_{t}^{\text{top}}, G_{t}^{\text{last}})$ \;
    $H \gets H + (C^{\Delta}[t] - C^{\Delta}[t-1])$ \;
    $C^{\Delta}_{max} \gets max(C^{\Delta}[t-w+1:t]$ \;
    \If{$\Big({H}^{-1}\big(C^{\Delta}[t]-C^{\Delta}_{max}\big) > R_{down}$ or $C^{\Delta}[t]>t_{\mathbf{C}b}\Big)$ or $|C^{\Delta}[t]|<t_{\mathbf{C}s}$}{
        $\Theta_{best} \gets \Theta_t$; $H \gets 0$ \Comment*[r]{beneficial training or no evident disparity}
    }
    \Else{
        \If{$\Theta_{best}=\Theta_{t-w}$}
            {Break  \Comment*[r]{evident useless training for $w$ epochs}}
    }
}
\If{$\textbf{avg}(\mathbf{D}[:w]) >t_{\mathbf{D}}$}{
$\Theta_{best} \gets \Theta_0 $\;
}
\kwReturn{$ \{f_M(x) | x\in \mathcal{D}, \Theta=\Theta_{best}\}$}
\end{algorithm}

\subsection{Notations of GradStop}\label{Appx:GradStopNotations}
We summarize the notations of the GradStop algorithm here. For clarity, we simplified the notations in a full batch gradient descent setting. At any epoch $t$, we denote current training dynamics as:
\begin{itemize}
    \item $t$ : time, i.e., the number of training epoch.
    \item $\mathcal{D}$ : the training dataset.
    \item $M$ : a UOD model. $M_t$ denotes model at epoch $t$.
    \item $k$ : the number of data points when performing $top-k$ and $last-k$ sampling in GradSample.
    \item $B_{eval} = \{x_1, \dots, x_n\}$ : the evalutation batch on which we perform GradStop.
    \item $G=G_{B_{eval}} = \{g_1, \dots, g_n\}$ : a set of gradient vectors generated by model $M_t$ on $B_{eval}$. Usually, $n$ equals to evaluation batch size $\|B_{eval}\|$ with $g_i$'s dimensionality equals to the dimensionality of parameter space $\|M\|$.
    \item $G^\text{top}$, $G^\text{last}$ : sets of gradient vectors generated by GradSample on $G$.
    \item $\mathbf{C}(G^\text{top})$, $\mathbf{C}(G^\text{last})$ : the cohesion metric reflecting the cohesion degree of gradient set $G^\text{top}$ or $G^\text{last}$.
    \item $\mathbf{D}(G_{\text{top}}, G_{\text{last}})$ : the divergence metric reflecting the divergence degree between gradient set $G_{\text{top}}$ and $G_{\text{last}}$.
    \item $t_{\mathbf{C}s}, t_{\mathbf{C}b}$: the significance threshold and benefit threshold of cohesion metric, with $t_{\mathbf{C}s} < t_{\mathbf{C}b}$.
    \item $t_\mathbf{D}$ : the stopping threshold of divergence metric.
    \item $w$: the size of the sliding window of the early stopping algorithm.
    \item $R_{down}$ : the smooth parameter in stopping algorithm by \citeauthor{EntropyStop}
\end{itemize}

\section{Theoretical Demonstration of Inlier Priority} \label{proof}
\subsection{Notations of Proof}
The related notations in our proof are as follows.

\begin{itemize}
    \item $\mathcal{D}$ : the training dataset.
    \item $\left|C_i\right|$, $\left|C_o\right|$: the number of samples belonging to inliers and outliers, respectively.
    \item $R=\frac{\left|C_i\right|}{\left|C_{o}\right|}$: the ratio of the number of inliers to the number of outliers.
    \item $L$: Lipschitz constant, with regard to Lipschitz smooth.
    \item $t$: time of the training process (i.e., number of iterations).
\end{itemize}
At each time $t$, we have:
\begin{itemize}
    \item $\eta_t$: learning rate, a positive real number.
    \item $\omega_t$ : set of network parameters.
    \item $f(\omega_t)=
    {\textstyle\sum_{x\in{D}}^{}f_x(\omega_t)}$ : loss function summed over all samples in the dataset.
    \item $f^i(\omega_t)=
    {\textstyle\sum_{x\in{C_i}}^{}f_x(\omega_t)}$ : loss function summed over all samples belonging to inliers. Similarly, we have $f^o(\omega_t)={\textstyle\sum_{x\in{C_o}}^{}f_x(\omega_t)}$
    \item $\nabla f^i(\omega_t)$: gradient computed on the loss function summed over all samples belonging to inliers. Similarly, we have $\nabla f^o(\omega_t)$.
    \item $\theta_t$ : the angle between $\nabla f^i(\omega_t)$ and $\nabla f^{o}(\omega_t)$.
    \item $\nabla f(\omega_t) = \nabla f^i(\omega_t) + \nabla f^{o}(\omega_t)$: gradient computed on the loss function summed over all samples.
    \item $\nabla_i = \left\| \nabla f^i(\omega_t) \right\|$ : the norm of the overall gradient of inliers. For clarity, $t$ is omitted in the simplified notation $\nabla_i$ in some formulas.
    \item $r_t=\frac{\nabla_i}{\nabla_{o}}$: the ratio of the norms of the inlier gradients to outlier gradients.
    \item $\triangle^f_t = (f^i(\omega_t)-f^i(\omega_{t+1}))-
    (f^{o}(\omega_t)-f^{o}(\omega_{t+1}))$
    : the difference of the decreasing speed of summed loss value between two classes.
    \item $\tilde{\triangle}^f_t = \frac{1}{\left|C_i\right|}(f^i(\omega_t)-f^i(\omega_{t+1}))-
    \frac{1}{\left|C_{o}\right|}(f^{o}(\omega_t)-f^{o}(\omega_{t+1}))$
    : \textit{loss decreasing speed gap}, the difference of the decreasing speed of averaged loss value between inliers and outliers. \textbf{If $\tilde{\triangle}^f_t>0$, inlier priority strengthens.}
\end{itemize}

\subsection{Derivation Process}
The main purpose of the proof is to demonstrate the connection between inlier priority, a crucial outlier assumption for the OD tasks, and certain training dynamics. Then, we can utilize these dynamics to monitor whether inlier priority is strengthening or weakening, and infer OD performance.
\par First, we would like to recap inlier priority and its importance. In outlier detection scenarios, the decreasing speed of the averaged loss value of inliers differs from one of the outliers due to their different distribution. To be specific, initially, average inlier reconstruction loss decreases rapidly, while average outlier loss decreases rather slowly, generating a gap between the two average losses after a short period. This gap is utilized by current OD methods to distinguish outliers. However, with the proceeding of learning, the gap is gradually mended, since both losses converge to 0. The aforementioned OD methods will fail after the gap is closed. Therefore, it is quite meaningful to apply an early stop to the training process, keeping the useful loss gap for afterward OD methods.\cite{EntropyStop} \par
With the following proof, we derive a lower bound for $\tilde{\triangle}^f_t$, the loss decreasing speed gap by showing a sufficient condition of $\tilde{\triangle}^f_t>\epsilon_{lower}>0$, which means inlier priority strengthens. \par


\begin{assumption}\label{lSmoothAssumption}
Assume that for each class $j$, $f^j(\omega_t)$ is $L-Smooth$.
\end{assumption}
\begin{assumption}\label{lrAssumption}
At each time $t$, there exists a learning rate $\eta_t>0$ being sufficiently small to guarantee the decrease of both $f^i$ and $f^{o}$.
\end{assumption}



\begin{proof}
In each iteration, the gradient descent algorithm uses the gradient summed over all samples to update the parameters:
\begin{align}
\omega^{t+1}=\omega^{t}-\eta_t\nabla f(\omega_t)
\end{align}
\noindent Since each $f^i(\omega_t)$ is $L-Smooth$, we have
\begin{align}
f^i(\omega_{t+1}) &= f^i(\omega_t) + \nabla f^i(\omega_t)^{\mathrm{T}}(\omega_{t+1} - \omega_{t}) \label{applyTaylor}
\\&\qquad\qquad+ \frac{1}{2} (\omega_{t+1} - \omega_{t})^{\mathrm{T}} \nabla^2f(u)(\omega_{t+1} - \omega_{t}) \notag
\\ \le f^i(\omega_t) &+ \nabla f^i(\omega_t)^{\mathrm{T}}(\omega_{t+1} - \omega_{t}) + \frac{L}{2} \left \| \omega_{t+1} - \omega_{t} \right \| ^2 \label{applySmoothUpper},
\end{align}

\noindent and
\begin{align}
f^i(\omega_{t+1}) \ge f^i(\omega_t) &+ \nabla f^i(\omega_t)^{\mathrm{T}}(\omega_{t+1} - \omega_{t}) - \frac{L}{2} \left \| \omega_{t+1} - \omega_{t} \right \| ^2 \label{applySmoothLower},
\end{align}

\noindent In \eqref{applyTaylor}, $f(u)$ is some convex combination of $\omega_{t+1}$ and $\omega_{t}$ with respect to multivariate Taylor expansion. We have \eqref{applySmoothUpper} and \eqref{applySmoothLower} by applying Lipschitz smooth, which ensures the term $(\omega_{t+1} - \omega_{t})^{\mathrm{T}} \nabla^2f(u)(\omega_{t+1} - \omega_{t})$ is at most $L \left \| \omega_{t+1} - \omega_{t} \right \| ^2$ and at least $-L \left \| \omega_{t+1} - \omega_{t} \right \|$. We first focus on the upper bound guaranteed by \eqref{applySmoothUpper}.

\begin{align}
f^i(\omega_{t+1}) &\le f^i(\omega_t) - \eta_t\nabla f^i(\omega_t)^{\mathrm{T}}\nabla f(\omega_t) + \frac{\eta_t^2 L}{2} \left \| \nabla f(\omega_t) \right \| ^2 \notag
\\ &= f^i(\omega_t) - \eta_t(\left \| \nabla f^i(\omega_t) \right \|^2 + \nabla f^i(\omega_t)^\mathrm{T}\nabla f^{o}(\omega_t)) + \frac{\eta_t^2 L}{2} \left \| \nabla f^i(\omega_t) + \nabla f^{o}(\omega_t) \right \|^2
\end{align}
\noindent From this, we can derive a lower bound of the decreasing speed of averaged loss: 
\begin{align}
\frac{1}{\left|C_i\right|}(f^i(\omega_t)-f^i(\omega_{t+1})) \ge &-\frac{\eta_t^2 L}{2\left|C_i\right|} (\left \| \nabla f^i(\omega_t) \right \|^2 + \left \| \nabla f^{o}(\omega_t) \right \|^2 \notag
\\ &+ 2\cos{\theta_t}\left \| \nabla f^i(\omega_t) \right \| \left \| \nabla f^{o}(\omega_t) \right \|) + \frac{\eta_t}{\left|C_i\right|}(\left \| \nabla f^i(\omega_t) \right \|^2 + \nabla f^i(\omega_t)^\mathrm{T}\nabla f^{o}(\omega_t))
\end{align}
\noindent Note that $\theta_t=\angle(\nabla f^i(\omega_t), \nabla f^{o}(\omega_t))$. It can reflect the degree of divergence between the inlier gradients and outlier gradients. A larger $\theta_t$ means a stronger rival between inliers and outliers, which may drive the model parameters towards two opposite directions respectively. For clarity, we use a simplified notation $\nabla_i = \left\| \nabla f^i(\omega_t) \right\|$:
\begin{align}
\frac{1}{\left|C_i\right|}(f^i(\omega_t)-f^i(\omega_{t+1})) \ge & -\frac{\eta_t^2 L}{2\left|C_i\right|} (\nabla_i^2 +\nabla_{o}^2 + 2\cos{\theta_t}\nabla_i\nabla_{o})+ \frac{\eta_t}{\left|C_i\right|}(\nabla_i^2 + \cos{\theta_t} \nabla_i \nabla_{o}) \label{lowerBound}
\end{align}
Now we get a lower bound of $f^i(\omega_t)-f^i(\omega_{t+1})$ with \eqref{applySmoothUpper}. With \eqref{applySmoothLower} and a similar procedure, we can derive an upper bound of $\frac{1}{\left|C_i\right|}(f^i(\omega_t)-f^i(\omega_{t+1}))$:
\begin{align}
&\frac{1}{\left|C_i\right|}f^i(\omega_t)-f^i(\omega_{t+1}) \le \frac{\eta_t^2 L}{2\left|C_i\right|} (\nabla_i^2 +\nabla_{o}^2 + 2\cos{\theta_t}\nabla_i\nabla_{o}) + \frac{\eta_t}{\left|C_i\right|}(\nabla_i^2 + \cos{\theta_t} \nabla_i \nabla_{o}) \label{upperBound}
\end{align}

\noindent \noindent Note that the previous formulas also apply to the other class $1-i$, since we assume that for each $l$, $f^l(\omega_t)$ is $L-Smooth$. With \ref{lrAssumption} and the inequation \eqref{lowerBound}, we have:
\\ \emph{For each class $j$,}
\begin{align}
-\frac{\eta_t^2 L}{2} (\nabla_j^2 +\nabla_{1-j}^2 &+ 2\cos{\theta_t}\nabla_i\nabla_{1-j}) +\eta_t(\nabla_j^2 + \cos{\theta_t} \nabla_j \nabla_{1-j}) > 0 \notag
\end{align}
\noindent Hence, 
\begin{gather}
\left\{
\begin{array}{l}
\eta_t L < \frac{2(\min(\nabla_i^2, \nabla_{o}^2) + \cos{\theta_t}\nabla_i \nabla_{o})}{\nabla_i^2 + 2\cos{\theta_t}\nabla_i\nabla_{o} + \nabla_{o}^2} \\
\theta_t<\cos^{-1}(-\frac{1}{r})
\end{array}
\right.
\end{gather}


\noindent Now we would like to derive a lower bound for $\triangle^f_t$ first. From \eqref{lowerBound} we can infer that a sufficient condition for $\triangle^f_t>0$ is:
\begin{align}
\eta_t(\nabla_i^2-\nabla_{o}^2)-\eta_t^2L(\nabla_i^2+2\cos{\theta_t}\nabla_i\nabla_{o}+\nabla_{o}^2) &> 0
\end{align}

\noindent Then the sufficient condition for $\triangle^f_t>0$ can be written as:
\begin{gather}
\nabla_i^2 - \nabla_{o}^2 - 2(\min(\nabla_i^2, \nabla_{o}^2) + \cos{\theta_t}\nabla_i \nabla_{o}) \ge 0 \notag
\end{gather}
\noindent With $r_t=\frac{\nabla_i}{\nabla_{o}}$, we have 2 cases: $r_t>=1$ and $r_t<1$.
\\ \emph{Case 1, $r_t>=1.$}
\begin{gather} 
(r_t^2-1)\nabla_{o}^2 - 2(\nabla_{o}^2+2r_t\cos{\theta_t}\nabla_{o}^2) \ge 0 \notag
\\ r_t^2-2r_t\cos{\theta_t} - 3 \ge 0  \notag
\\ r_t \ge \sqrt{\cos^2{\theta_t} + 3} + \cos{\theta_t}  \notag
\end{gather}
\\ \emph{Case 2, $r_t<1.$}
\begin{gather}
(r_t^2-1)\nabla_{o}^2 - 2(r_t^2\nabla_{o}^2+2r_t\cos{\theta_t}\nabla_{o}^2) \ge 0 \notag
\\ -r_t^2-2r_t\cos{\theta_t} - 1 \ge 0 \notag
\end{gather}
\noindent Combining the two cases, we gain a sufficient condition to ensure $\triangle^f_t > 0$:
\begin{equation}
r_t \ge \sqrt{\cos^2{\theta_t} + 3} + \cos{\theta_t} \label{sufficientCondition}
\end{equation}
\noindent Additionally, we also want to inspect the expectation of \emph{sample} loss, i.e., $\tilde{\triangle}^f_t$, instead of the \emph{summed} loss of each class, since sample loss is directly utilized for OD algorithms. With a similar derivation procedure, we can derive a sufficient condition for $\tilde{\triangle}^f_t>0$:
\\ \emph{Case 1, $r_t>=1.$}
\begin{gather} 
\left|C_{o}\right| r_t^2 - 2\cos{\theta_t}\left|C_i\right| - 2(\left|C_i\right|+\left|C_{o}\right|) \ge 0 \notag
\end{gather}
\\ \emph{Case 2, $r_t<1.$}
\begin{gather}
-\left|C_i\right|(r_t^2+2\cos{\theta_t}r_t+1) \ge 0 \notag
\end{gather}
\noindent Combining the two cases, we gain a sufficient condition to ensure $\tilde{\triangle}^f_t>0$:
\begin{gather}
r_t \ge \frac{\cos{\theta_t}\left|C_i\right| + \sqrt{\cos^2{\theta_t}\left|C_{i}\right|^2+2\left|C_i\right| \left|C_{o}\right| + \left|C_{o}\right|^2}}{\left|C_{o}\right|} \notag
\\ r_t > \cos{\theta_t}R + \sqrt{\cos^2{\theta_t}R^2 + 2R + 1} \label{sufficientConditionNormed}
\end{gather}
\noindent Finally, we gain a sufficient condition that ensures the decreasing speed of sample-wise loss of inlier class $i$ to be faster than the one of outlier class $o$, i.e. inequation \eqref{sufficientConditionNormed}. A larger $r_t$ ensures a larger lower bound for $\tilde{\triangle}^f_t>0$, indicating a stronger alignment with the OD assumption inlier priority. Note that, $\cos{\theta_t}$ is the divergence metric $\mathbf{D}(\nabla f^i(\omega_t), \nabla f^o(\omega_t))$, and $r_t=\frac{\mathbf{C}(\nabla f^i(\omega_t)){\textstyle\sum_{x\in{C_{i}}}^{} \|\nabla f_x(\omega_t)\|}}{\mathbf{C}(\nabla f^o(\omega_t)){\textstyle\sum_{x\in{C_{o}}}^{} \|\nabla f_x(\omega_t)\|}}$, which are the theoretical enlightenments of designing cohesion metric $\mathbf{C}$ and divergence metric $\mathbf{D}$, respectively. $\log{(r_t)}=\mathbf{C}(\nabla f^i(\omega_t))-\mathbf{C}(\nabla f^o(\omega_t)) + \tilde{R}$ where $\tilde{R}$ is linear with $R$ during training. Therefore, the stopping indicator $\mathbf{C}(G^\text{last}) - \mathbf{C}(G^\text{top})$ is used to approximate $\mathbf{C}(\nabla f^i(\omega_t))-\mathbf{C}(\nabla f^o(\omega_t))$ thus reflecting $\log{(r_t)}$. We left exploring more elaborate metric designs being closer to the theoretical conclusion as our future work.

\end{proof}

\begin{table}[htbp]
  \centering
  \caption{Real-world dataset pool}
  \vskip 0.1in
    \resizebox{0.668\columnwidth}{!}{
    \begin{tabular}{cl|rrr}
    \toprule
          & \textbf{Dataset} & \textbf{Num Pts} & \textbf{Dim} & \textbf{\% Outlier} \\
    \midrule
    1     & ALOI  & 49534 & 27    & 3.04  \\
    2     & annthyroid & 7200  & 6     & 7.42  \\
    3     & backdoor & 95329 & 196   & 2.44  \\
    4     & breastw & 683   & 9     & 34.99  \\
    5     & campaign & 41188 & 62    & 11.27  \\
    6     & cardio & 1831  & 21    & 9.61  \\
    7     & Cardiotocography & 2114  & 21    & 22.04  \\
    8     & celeba & 202599 & 39    & 2.24  \\
    9     & census & 299285 & 500   & 6.20  \\
    10    & cover & 286048 & 10    & 0.96  \\
    11    & donors & 619326 & 10    & 5.93  \\
    12    & fault & 1941  & 27    & 34.67  \\
    13    & fraud & 284807 & 29    & 0.17  \\
    14    & glass & 214   & 7     & 4.21  \\
    15    & Hepatitis & 80    & 19    & 16.25  \\
    16    & http  & 567498 & 3     & 0.39  \\
    17    & InternetAds & 1966  & 1555  & 18.72  \\
    18    & Ionosphere & 351   & 32    & 35.90  \\
    19    & landsat & 6435  & 36    & 20.71  \\
    20    & letter & 1600  & 32    & 6.25  \\
    21    & Lymphography & 148   & 18    & 4.05  \\
    22    & magic & 19020 & 10    & 35.16  \\
    23    & mammography & 11183 & 6     & 2.32  \\
    24    & mnist & 7603  & 100   & 9.21  \\
    25    & musk  & 3062  & 166   & 3.17  \\
    26    & optdigits & 5216  & 64    & 2.88  \\
    27    & PageBlocks & 5393  & 10    & 9.46  \\
    28    & pendigits & 6870  & 16    & 2.27  \\
    29    & Pima  & 768   & 8     & 34.90  \\
    30    & satellite & 6435  & 36    & 31.64  \\
    31    & satimage-2 & 5803  & 36    & 1.22  \\
    32    & shuttle & 49097 & 9     & 7.15  \\
    33    & skin  & 245057 & 3     & 20.75  \\
    34    & smtp  & 95156 & 3     & 0.03  \\
    35    & SpamBase & 4207  & 57    & 39.91  \\
    36    & speech & 3686  & 400   & 1.65  \\
    37    & Stamps & 340   & 9     & 9.12  \\
    38    & thyroid & 3772  & 6     & 2.47  \\
    39    & vertebral & 240   & 6     & 12.50  \\
    40    & vowels & 1456  & 12    & 3.43  \\
    41    & Waveform & 3443  & 21    & 2.90  \\
    42    & WBC   & 223   & 9     & 4.48  \\
    43    & WDBC  & 367   & 30    & 2.72  \\
    44    & Wilt  & 4819  & 5     & 5.33  \\
    45    & wine  & 129   & 13    & 7.75  \\
    46    & WPBC  & 198   & 33    & 23.74  \\
    47    & yeast & 1484  & 8     & 34.16  \\
    \bottomrule
    \end{tabular}%
    }
  \label{pool-dataset}
\end{table}%

\newpage

\section{Experimental details}\label{Appx:Exp}
\subsection{Datasets}
The details of the 47 datasets are shown in Table \ref{pool-dataset}.

\subsection{Implementation Details in Experiments}
All results are averaged on 3 runs with different random seeds. VanillaAE is built with only one hidden layer with size $h_{dim}=64$. The experiments of two ensemble models are based on the open-source code of ROBOD\footnote{https://github.com/xyvivian/ROBOD}. Implementations of DeepSVDD and VAE are from PyOD\footnote{https://github.com/yzhao062/pyod}, and implementation of RDP is from \cite{RDP}\footnote{https://github.com/billhhh/RDP}. We first randomly downsample datasets larger than 10,000 to a size of 10,000 before training to shorten the experiment pipeline. Downsampling almost does not influence the performance evaluation according to our observation. The size of \( B_{\text{eval}} \) is set to 400. For efficiency, we re-sample to generate \( G_{\text{top}} \) and \( G_{\text{last}} \) using GradSample every ten epochs rather than every epoch. To simplify the observation and analysis of training dynamics, we adopted a full-batch gradient descent configuration in the experiments. The default hyperparameters of deep UOD models and GradStop are in Table \ref{tab:hyperparams}. The patience parameters of EntropyStop and GradStop are set the same to be $20$, and $R_{down}$ are both $0.001$. Other hyperparameters are the same as the default hyperparameters in their original codes. 
Our codes are available at 
\url{https://anonymous.4open.science/r/gradAE-879F}.

\begin{table}[htbp]
  \centering
  \caption{Default hyperparameters of deep UOD models and GradStop}
  \vskip 0.1in
    \resizebox{0.6\columnwidth}{!}{
    \begin{tabular}{c|cccc}
    \toprule
    \toprule
          & \multicolumn{1}{c|}{AE} & \multicolumn{1}{c|}{DeepSVDD} & \multicolumn{1}{c|}{RDP} & VAE \\
    \midrule
    $\#epochs$ & \multicolumn{4}{c}{100} \\
    \midrule
    $lr$    & \multicolumn{1}{c|}{0.005} & \multicolumn{1}{c|}{0.001} & \multicolumn{1}{c|}{0.5} & 0.01 \\
    \midrule
    $k$     & \multicolumn{1}{c|}{20} & \multicolumn{1}{c|}{20} & \multicolumn{1}{c|}{20} & 10 \\
    \midrule
    $[t_{\mathbf{C}s}, t_{\mathbf{C}b}]$ & \multicolumn{1}{c|}{$[0.01, 0.05]$} & \multicolumn{1}{c|}{$[0.0, 0.1]$} & \multicolumn{1}{c|}{$[0, 0.5]$} & \multicolumn{1}{c|}{$[0.01, 0.5]$} \\
    \midrule
    $t_\mathbf{D}$ & \multicolumn{1}{c|}{1.57} & \multicolumn{1}{c|}{1.57} & \multicolumn{1}{c|}{$\infty$} & $\infty$ \\
    \midrule
    $w$     & \multicolumn{1}{c|}{20} & \multicolumn{1}{c|}{10} & \multicolumn{1}{c|}{50} & 20 \\
    \midrule
    $R_{down}$     & \multicolumn{4}{c}{0.001} \\
    \bottomrule
    \bottomrule
    \end{tabular}%
    }
    \vskip -0.05in
  \label{tab:hyperparams}%
\end{table}%

\subsection{Discussion of the Hyperparameters and ablation study}\label{exp-hp}
The GradStop algorithm introduces four new hyperparameters (HPs): the number of samples $k$ for GradSample, the interval thresholds $[t_{\mathbf{C}s}, t_{\mathbf{C}b}]$ for the difference in cohesion degree $C^{\Delta}$, and the threshold $t_\mathbf{D}$ for divergence degree. To research the utility of these HPs, we perform a grid search with the setting shown in Table \ref{tab:HP-grid}. Default HPs are in bold. Since $t_{\mathbf{C}s} < t_{\mathbf{C}b}$, this setting totally generates 30 combinations of HPs. The result is in Table \ref{tab:HP-grid-result}. ``Control Variable'' only applies ``HP setting'' and keeps other HPs default, and ``Mean'' is computed as the average AUC across all configurations in the HP grid where the ``HP setting'' is applied. From the results, we can see that GradStop generally improves VanillaAE under different HP settings, showing its insensitivity toward HPs. Following we elucidate the functionality of each HP along with the analysis of the experimental result of the grid search.

\begin{table}[htbp]
  \centering
  \caption{Value sets of HP grid search}
    \vskip 0.05in
    \resizebox{0.35\columnwidth}{!}{
    \begin{tabular}{cc}
    \toprule
    Parameter & Value Set \\
    \midrule
    $k$     & $[5, 10, \textbf{20}, 50]$ \\
    $t_\mathbf{D}$  & $[\textbf{1.57}, 1.6, \infty]$ \\
    $t_{\mathbf{C}s}$ & $[0, \textbf{0.01}, 0.05, 0.1]$ \\
    $t_{\mathbf{C}b}$ & $[\textbf{0.05}, 0.1, 0.5]$ \\
    \bottomrule
    \end{tabular}%
    }
  \label{tab:HP-grid}%
\end{table}%

$k$ represents the number of samples that GradSample selects from the evaluation batch $B_{eval}$. Experimental results show that the algorithm performs well when $k \approx 0.05|B_{eval}|$. Due to the low proportion of anomalies, a too-large $k$ can lead $G^{top}$ to include too many near-normal samples. A too-small $k$ may lower the generality of the sampled sets. Both decrease the approximation quality of retrieved sets thus decreasing the performance, as the ones with $k=5$ and $k=50$ show in Table \ref{tab:HP-grid-result}.

$t_{\mathbf{C}b}$ is used to judge whether the inlier priority holds to a certain degree (for its principle, please refer to the proof in Appendix \ref{proof} and the case studies in Appendix \ref{exp-case-study}). When $C^{\Delta}>t_{\mathbf{C}b}$, it indicates that the current iteration still satisfies the outlier assumption sufficiently, making training beneficial for the OD task. The threshold $t_{\mathbf{C}s}$ controls the confidence level of the approximation—training will stop only when $|C^{\Delta}|$ is significantly large, indicating there is a significant deviation between the sampled $G^\text{top}$ and $G^\text{last}$. This is designed to alleviate the cold start problem in some cases where inlier priority needs epochs to exhibit. Additionally, on certain algorithm-dataset pairs, inlier priority only weakly or even does not hold and GradSample may lose effectiveness. In these cases, the deep model itself is not able to perform OD on the dataset since the outlier assumption does not hold. An optimistic setting that continues training is used to enhance the robustness of early stopping. Generally, for algorithm-dataset pairs that better satisfy the outlier assumption, the accuracy of the set gained by GradSample is higher, allowing for a lower value of $t_{\mathbf{C}s}$, and vice versa. In table \ref{tab:HP-grid-result}, we carried ablation study with $t_{\mathbf{C}s}=0$. The result shows that it can slightly improve the performance especially when it is strict to decide a epoch beneficial (i.e., a large $t_{\mathbf{C}b}$).

In algorithms without constraints on the representation space (such as AE, DeepSVDD), complete toxic training from the beginning to the end is common, where anomalies can be directly identified through model structure with randomized parameters and learning leads to performance degradation. In contrast, models with constrained representation spaces (such as VAE, RDP) rarely exhibit toxic training since degradation is mitigated with regulation. Our experiments reveal that completely abandoning training when the divergence degree in the early stages of training is larger than $t_\mathbf{D}=\pi/2$ effectively prevents toxic training for unconstrained models. We leave the exploration of this intriguing phenomenon for future work. Ablation study is also carried out with $t_\mathbf{D}=\infty$, which means not using $t_\mathbf{D}$. The result shows that when $t_\mathbf{D}=\pi/2$, the performance of GradAE is improved from $0.760$ to $0.775$, showing the significant effectiveness of $t_\mathbf{D}$'s mechanism. 

\begin{table}[htbp]
  \centering
  \caption{Performance of AE with different hyperparameters (HPs). ``Control Variable'' only applies ``HP setting'' and keeps other HPs default, and ``Mean'' is computed as the average AUC across all configurations in the HP grid where the ``HP setting'' is applied.}
    \vskip 0.1in
    \begin{tabular}{ccc}
    \toprule
    HP setting & Control Variable & Mean \\
    \midrule
    $k=5$   & 0.765 & 0.759 \\
    $k=10$  & 0.762 & 0.761 \\
    $\mathbf{k=20}$  & \textbf{0.775} & \textbf{0.764} \\
    $k=50$  & 0.763 &  0.756 \\ 
    \midrule
    $\mathbf{t_\mathbf{D}=1.57}$ & \textbf{0.775} & 0.763 \\
    $t_\mathbf{D}=1.6$ & 0.772 & \textbf{0.765} \\
    $t_\mathbf{D}=\infty$ & 0.760 & 0.756 \\
    \midrule
    $[t_{\mathbf{C}s},t_{\mathbf{C}b}]=[0, 0.05]$ & 0.773 & 0.764 \\
    $[t_{\mathbf{C}s},t_{\mathbf{C}b}]=[0, 0.1]$ & 0.770 & 0.761 \\
    $[t_{\mathbf{C}s},t_{\mathbf{C}b}]=[0, 0.5]$ & 0.763 & 0.756 \\
    $\mathbf{[t_{\mathbf{C}s}, t_{\mathbf{C}b}]=[0.01, 0.05]}$ & \textbf{0.775} & \textbf{0.765} \\
    $[t_{\mathbf{C}s}$,$t_{\mathbf{C}b}]=[0.05, 0.1]$ & 0.770 & \textbf{0.765} \\
    $[t_{\mathbf{C}s}$,$t_{\mathbf{C}b}]=[0.1, 0.5]$ & 0.763 & 0.762 \\
    \bottomrule
    \end{tabular}%
  \label{tab:HP-grid-result}%
\end{table}%

\subsection{Case Studies}\label{exp-case-study}
To better understand the core algorithm principle, the gradient cohesion, we conducted three case studies on datasets \textit{glass}, \textit{wine}, and \textit{optdigits}. From this, we can intuitively see the close correspondence between the proposed metric $\mathbf{C}$ and the variation tendency of AUC. 

\begin{figure}[ht]
    \centering
    \begin{subfigure}[b]{0.3\linewidth}
        \includegraphics[width=\linewidth]{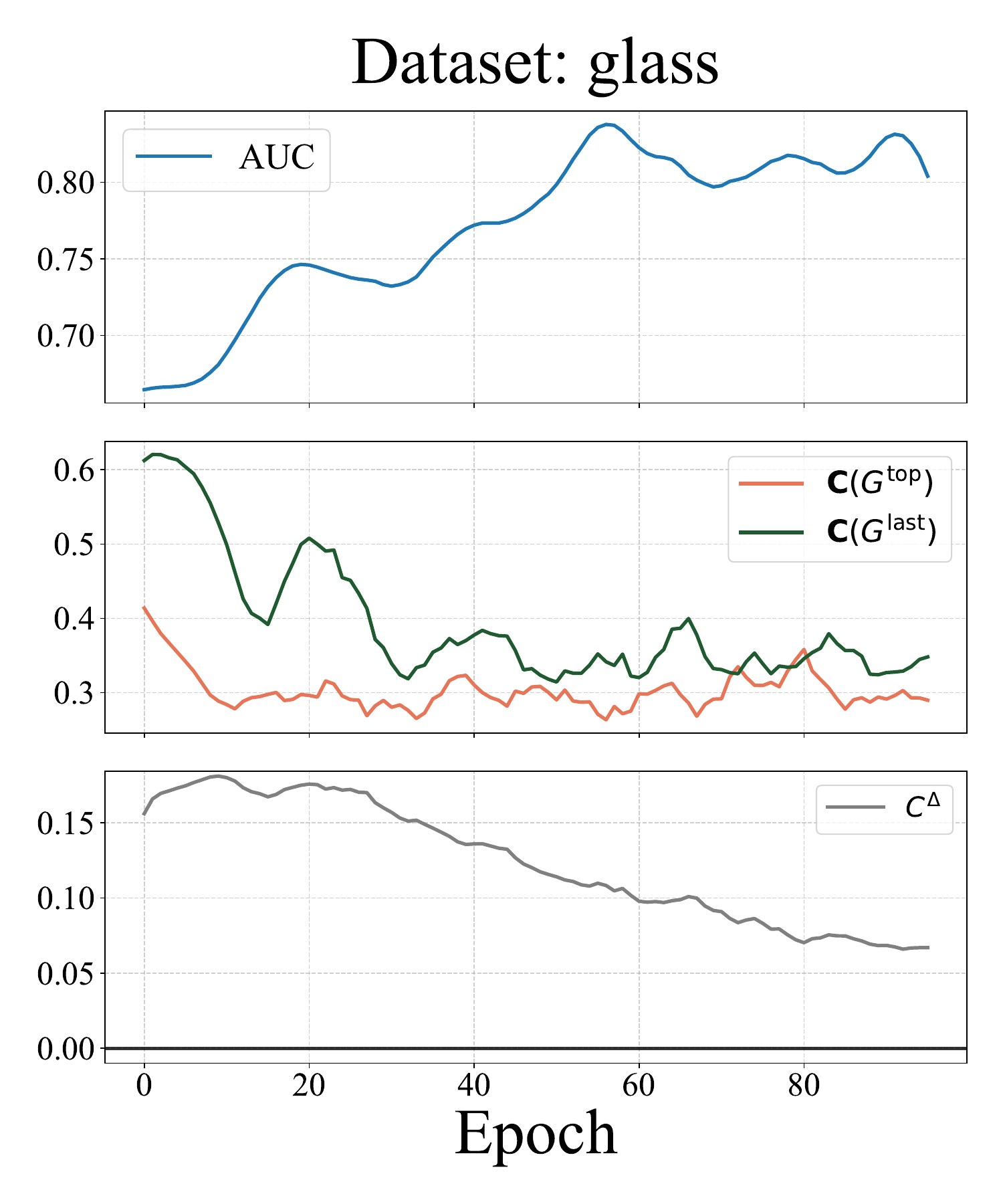}
        \caption{Case study on \textit{glass}.}
        \label{fig:case-study-glass}
    \end{subfigure}
    \hspace{-0.5em} 
    \begin{subfigure}[b]{0.3\linewidth}
        \includegraphics[width=\linewidth]{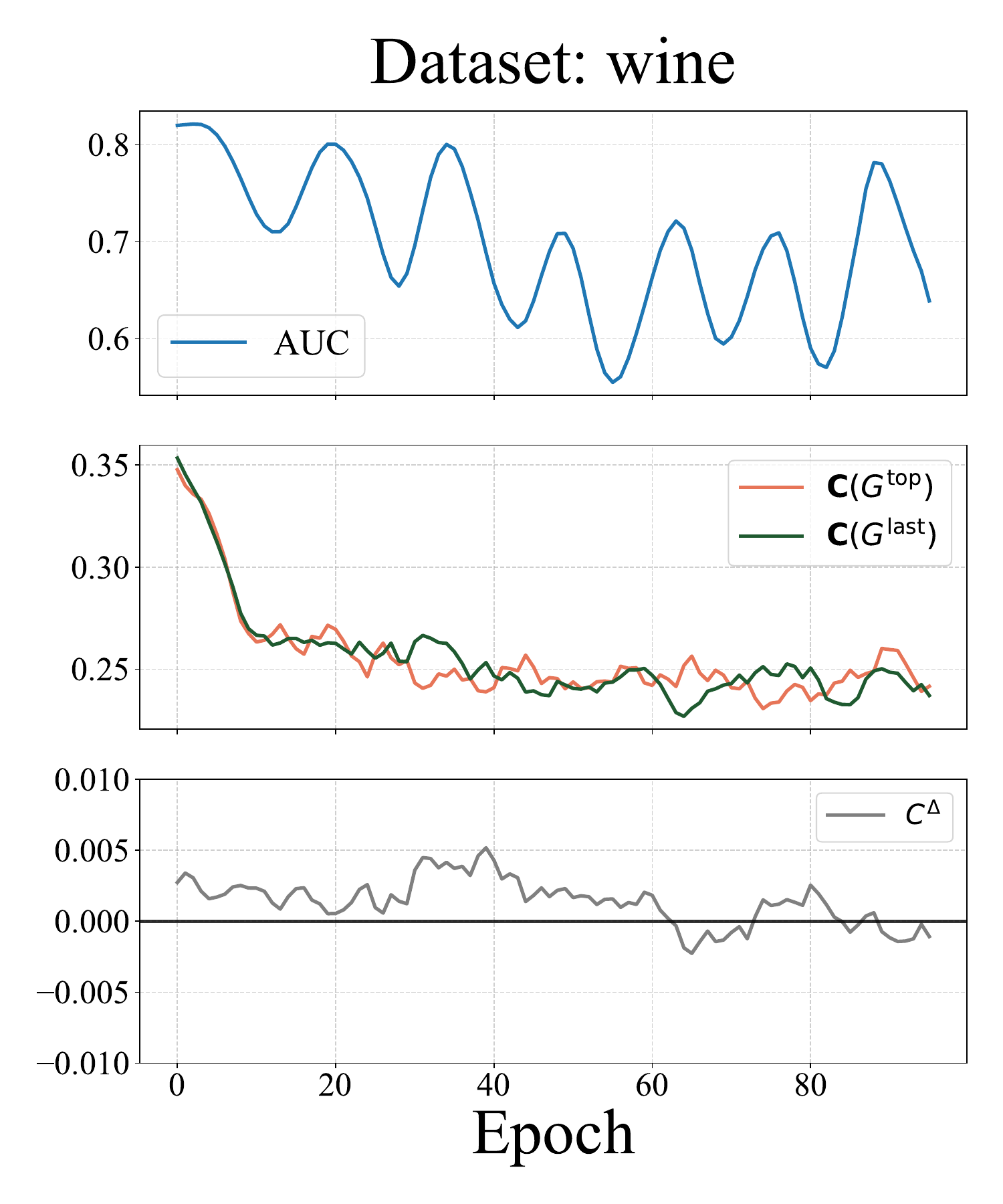}
        \caption{Case study on \textit{wine}.}
        \label{fig:case-study-wine}
    \end{subfigure}
    \hspace{-0.5em} 
    \begin{subfigure}[b]{0.3\linewidth}
        \includegraphics[width=\linewidth]{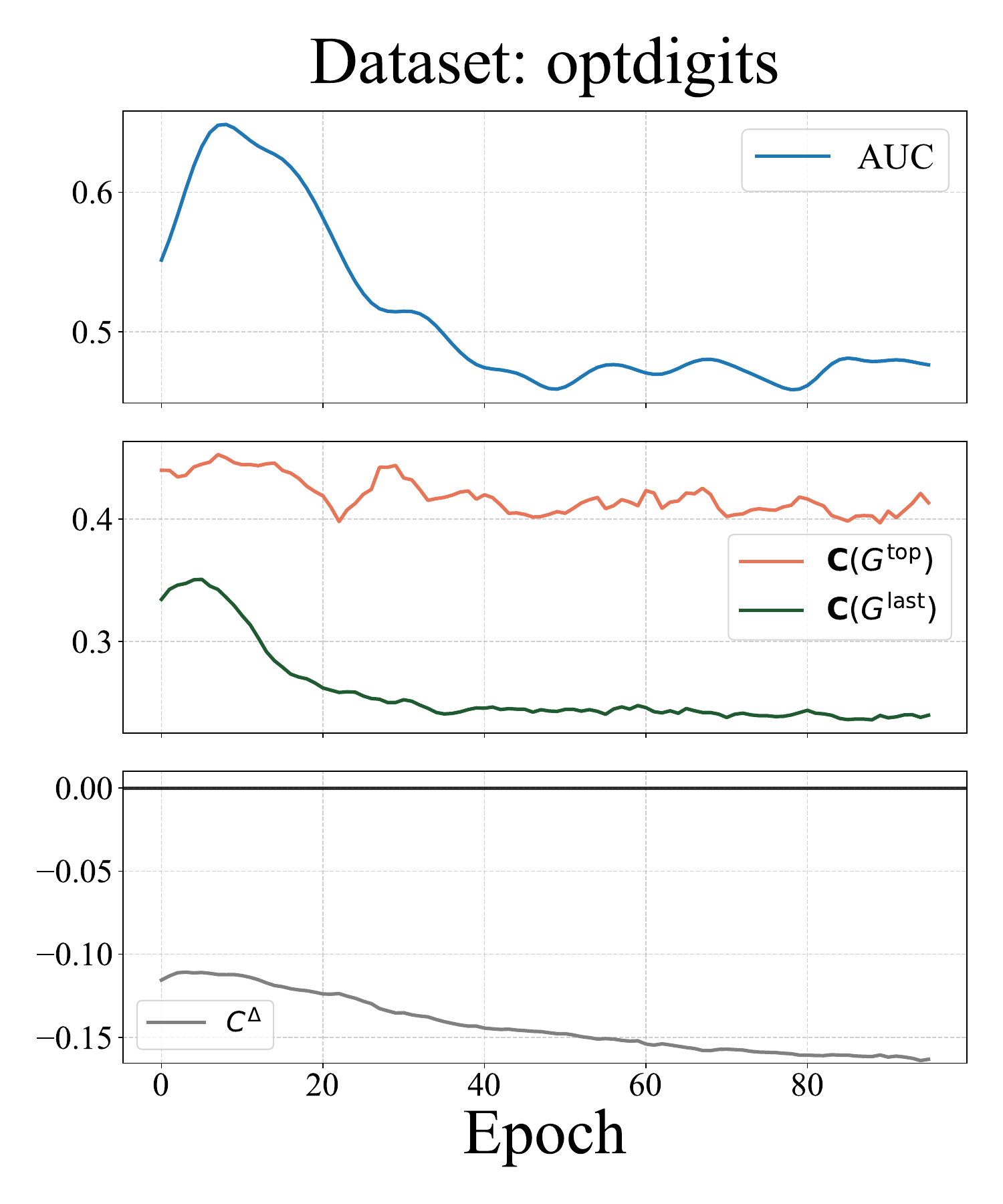}
        \caption{Case study on \textit{optdigits}.}
        \label{fig:case-study-optdigits}
    \end{subfigure}
    \vspace{-0.5em} 
    \caption{Case studies on datasets \textit{glass}, \textit{wine}, and \textit{optdigits}, showing close correspondence between the $\mathbf{C^\Delta}$ and the variation trend of AUC. Top: AUC. Middle: $\mathbf{C}(G^{\text{last}})$ and $\mathbf{C}(G^{\text{top}})$. Bottom: $C^{\Delta}=\mathbf{C}(G^{\text{last}})-\mathbf{C}(G^{\text{top}})$.}
    \label{fig:case-studies}
\end{figure}

The results on \textit{glass}, \textit{wine}, and \textit{optdigits} are respectively shown in Fig. \ref{fig:case-study-glass}, \ref{fig:case-study-wine}, and \ref{fig:case-study-optdigits}. All figures consist of three subfigures: Each of the three figures consists of three subplots, which from top to bottom are: AUC, $\mathbf{C}(G^{\text{last}})$ and $\mathbf{C}(G^{\text{top}})$, $C^{\Delta}=\mathbf{C}(G^{\text{last}})-\mathbf{C}(G^{\text{top}})$. All subplots share the same horizontal axis representing the epoch number, and the curve is mean-smoothed with adjacent values at a window size of 5 for clarity.

For \textit{glass} in Fig. \ref{fig:case-study-glass}, training is beneficial, since AUC follows a generally rising trend. This is because inlier priority holds. During the first 60 epochs, AUC rises rapidly and reaches its peak. Correspondingly, the cohesion of $G^\text{last}$ greatly outnumbered $G^\text{top}$ at first, resulting in a large $C^\Delta$. This means the model can leverage the distribution disparity of inliers and outliers to separate them in the representation space. With training proceeds, $C^\Delta$ decreases, indicating learning is becoming more and more useless for the OD task. It is reflected in the AUC curve fluctuating after around epoch 60. 

For \textit{wine} in Fig. \ref{fig:case-study-wine}, training is disordered. AUC fluctuates acutely during the whole training process. As we can see in the cohesion curves, $\mathbf{C}(G^{\text{last}})$ and $\mathbf{C}(G^{\text{top}})$ twines together and $C^\Delta$ is always around 0. This indicates that the distribution of outliers and inliers have similar impacts on training. The model cannot separate them, causing a chaotic AUC trend. In this circumstance, the inherent difficulty in reconstructing outliers contributes most to the performance, with an AUC of more than 0.8 at the first training epoch.

For \textit{wine} in Fig. \ref{fig:case-study-optdigits}, training is toxic. AUC rises temporarily at the first 5 epochs and then consistently drops. Accounting for this, $\mathbf{C}(G^{\text{last}})$ quickly decreases after epoch 5, while $\mathbf{C}(G^{\text{top}})$ remains high and unchanged during the whole training. This means that the distributions of inliers and outliers exhibit just the opposite during training: the outliers are easier to fit, and ``outlier priority'' holds instead of the inlier priority. As a result, training is harmful to the OD task. The above three case studies intuitively reveal the mechanism of the cohesion metric estimating model's convergence trend towards inlier and outlier distributions, thus estimating the AUC trend. Also, the effectiveness of GradSample is demonstrated since $G^{\text{last}}$ and $G^{\text{top}}$ are retrieved by it.




\newpage
\subsection{Detailed Experimental Results}\label{Appx:exp-details}
\begin{figure}[ht]
  \centering
\includegraphics[width=0.32\textwidth]{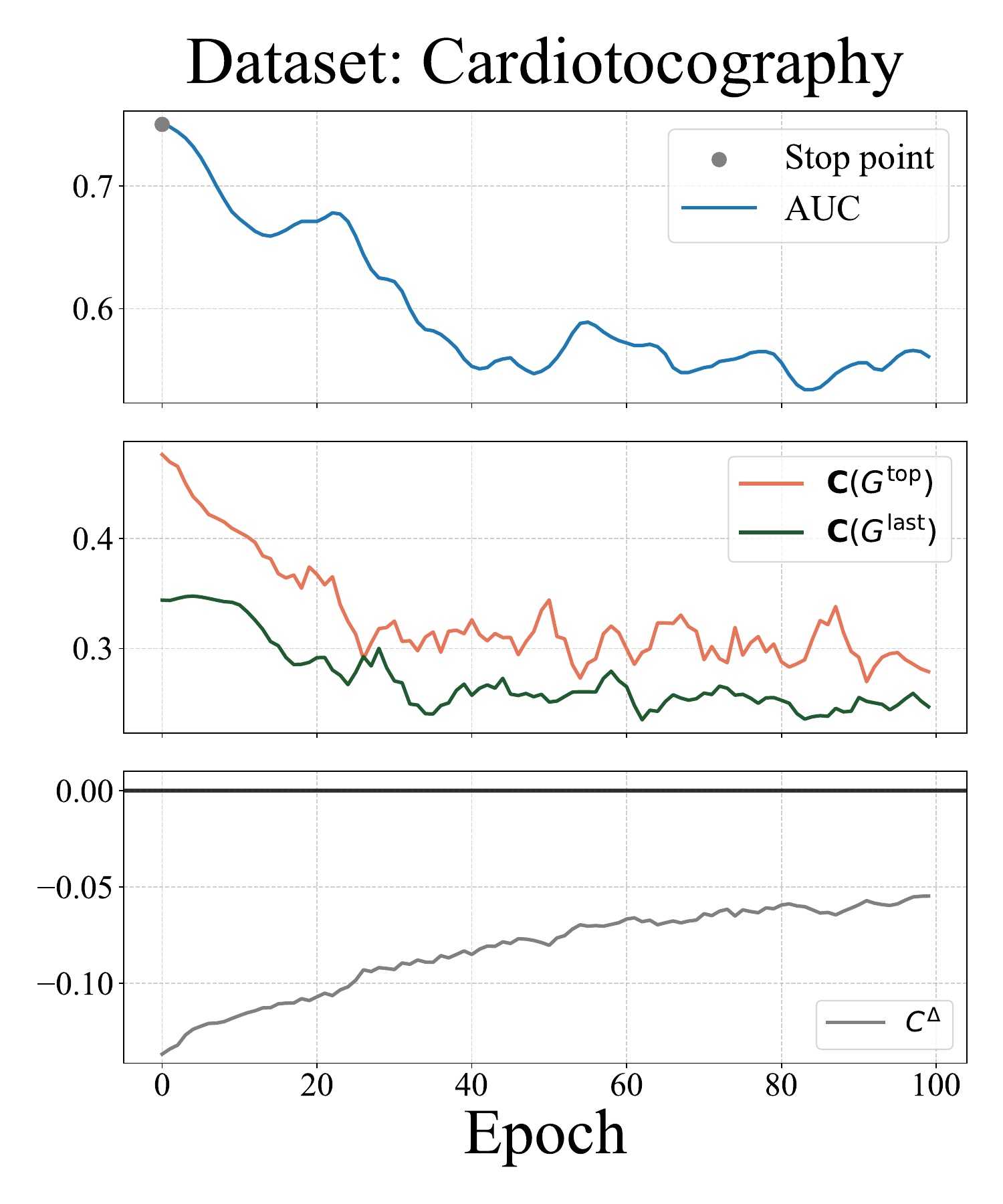}
\includegraphics[width=0.32\textwidth]{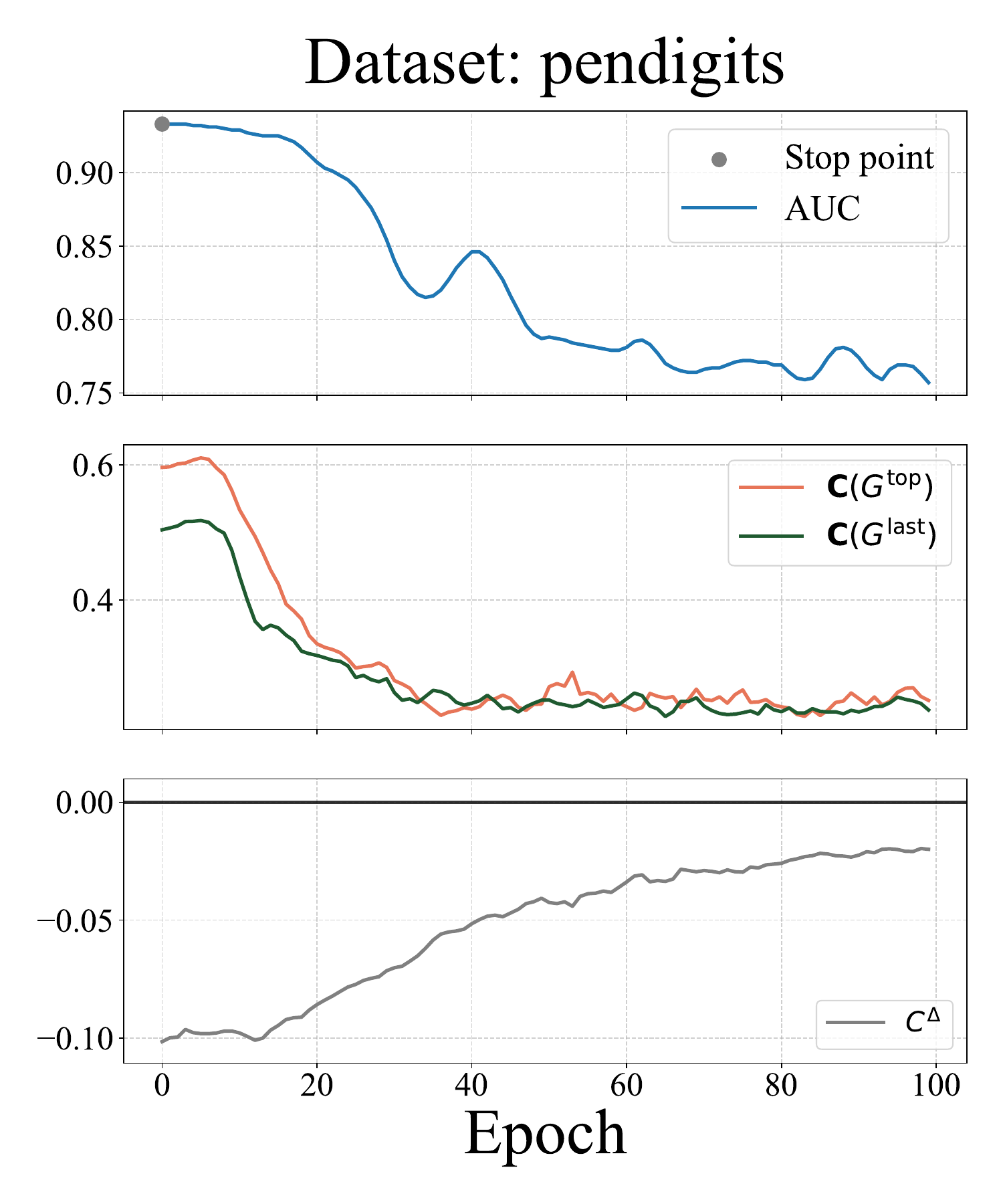}
\includegraphics[width=0.32\textwidth]{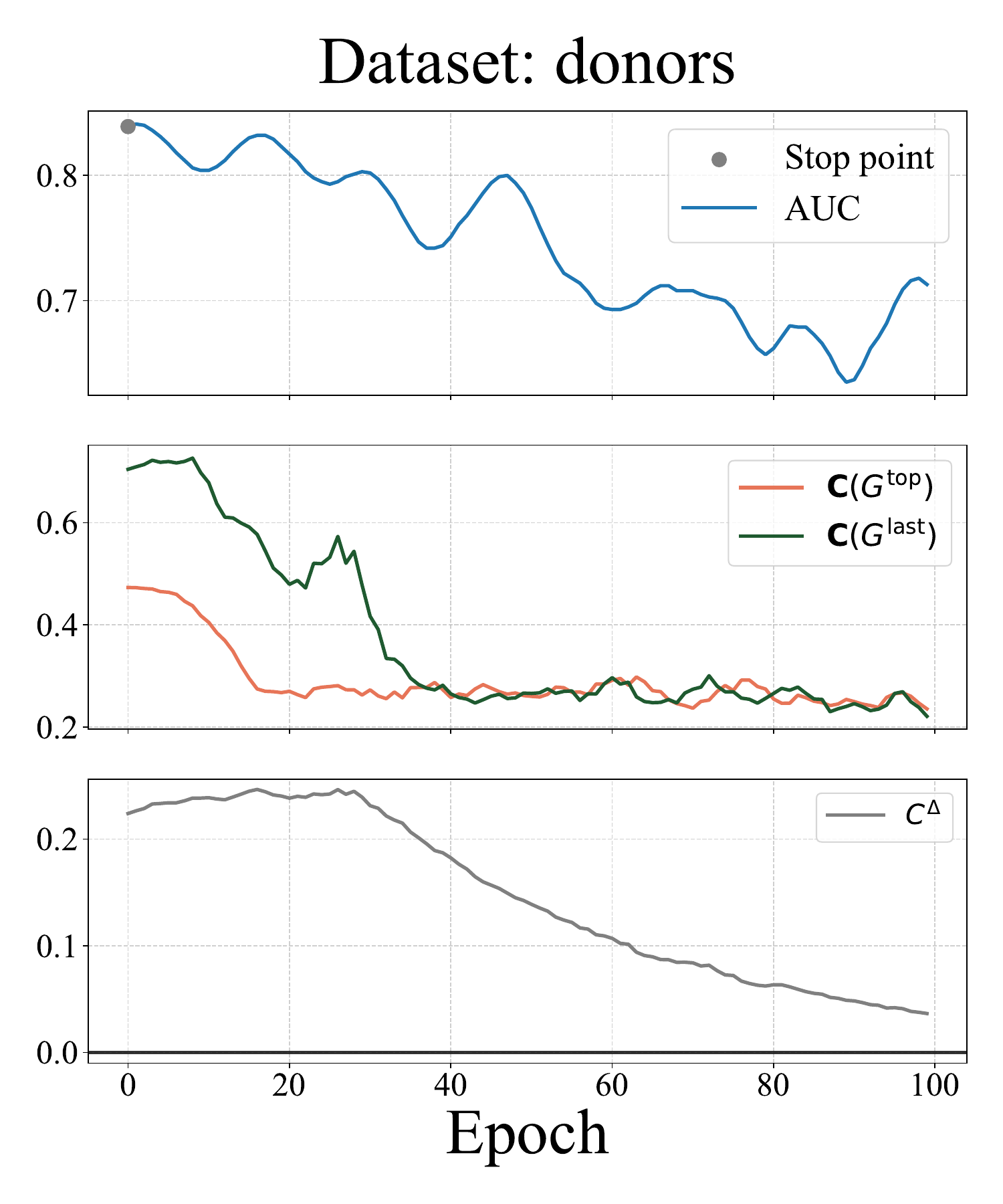}
\includegraphics[width=0.32\textwidth]{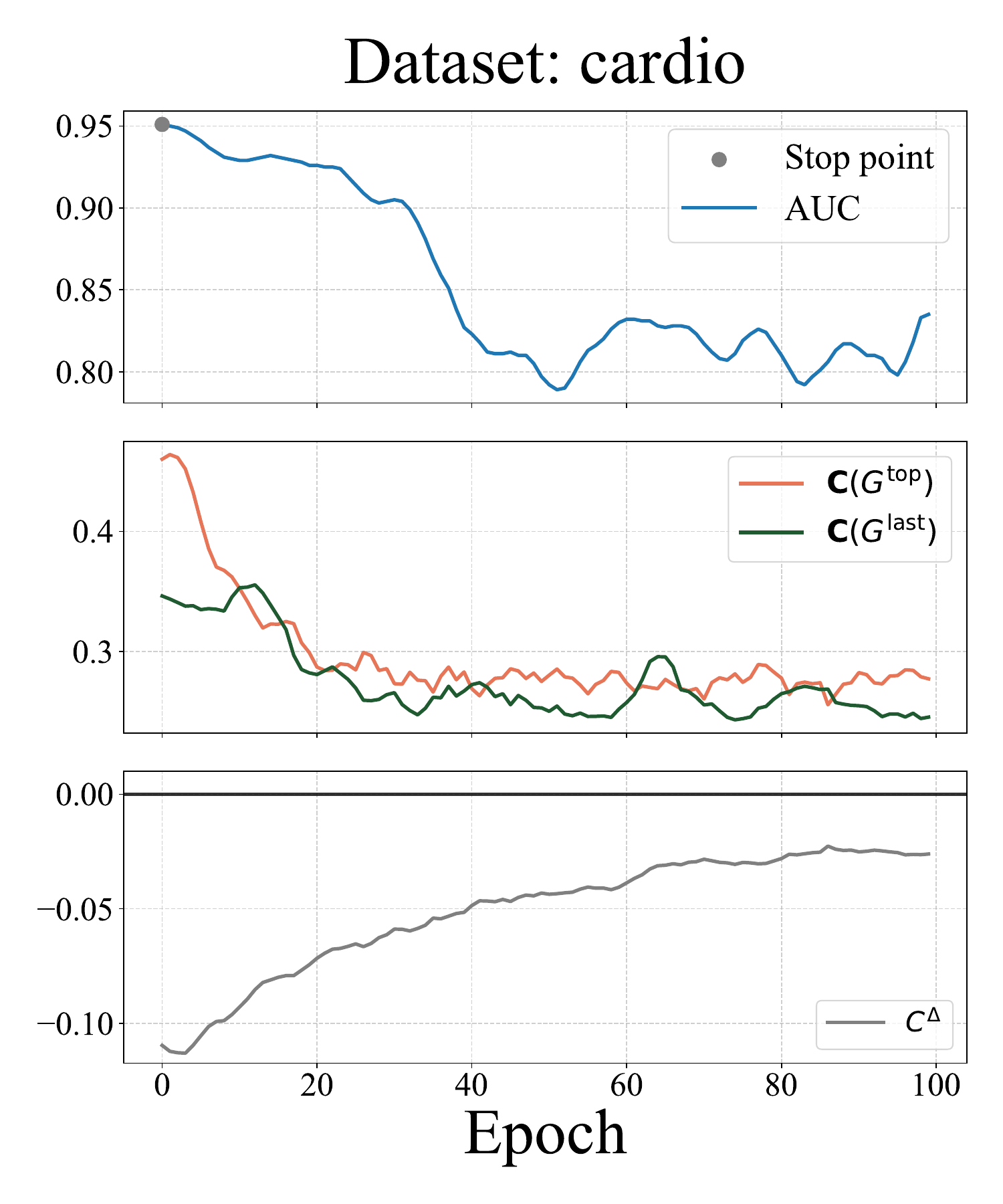}
\includegraphics[width=0.32\textwidth]{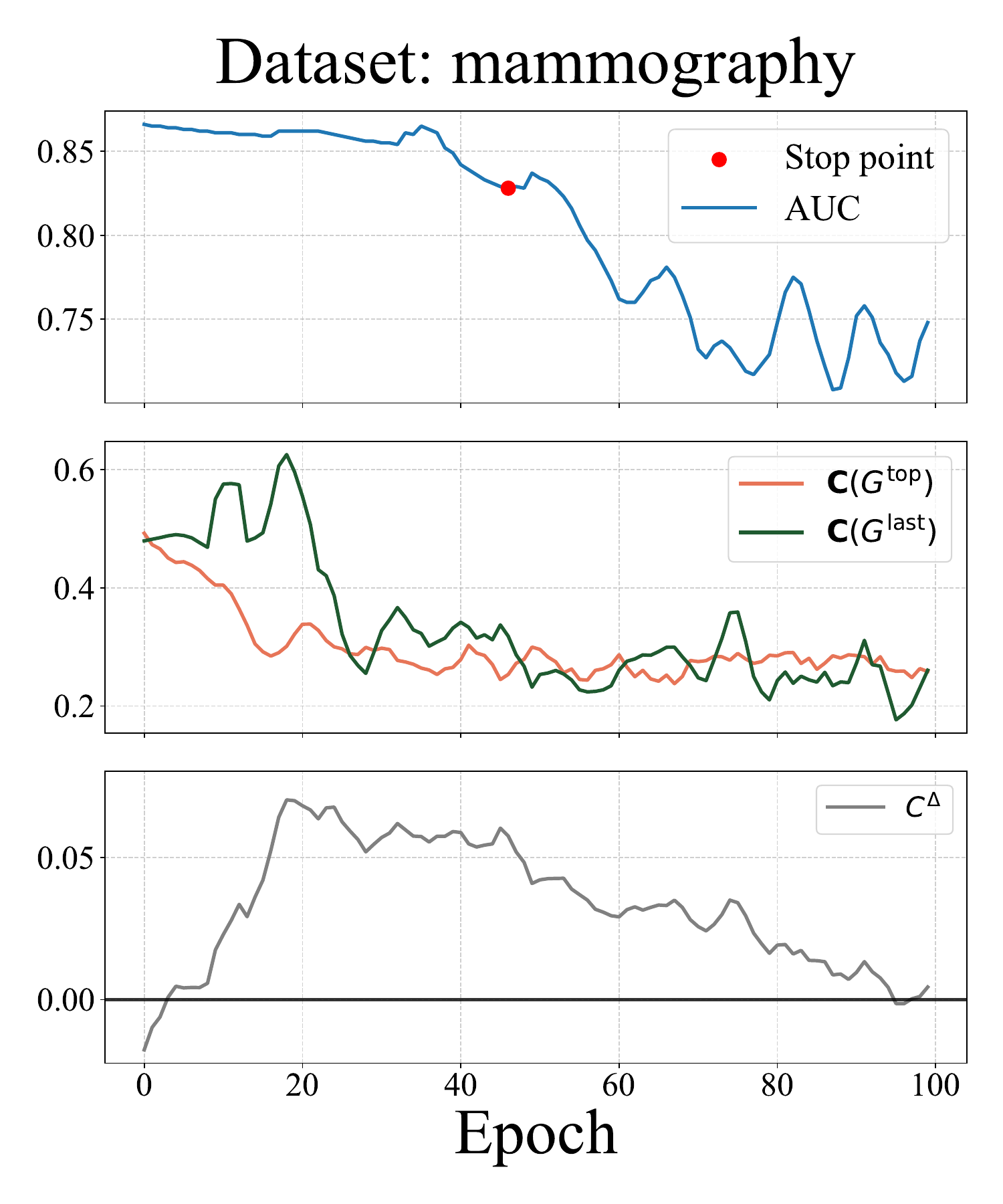}
\includegraphics[width=0.32\textwidth]{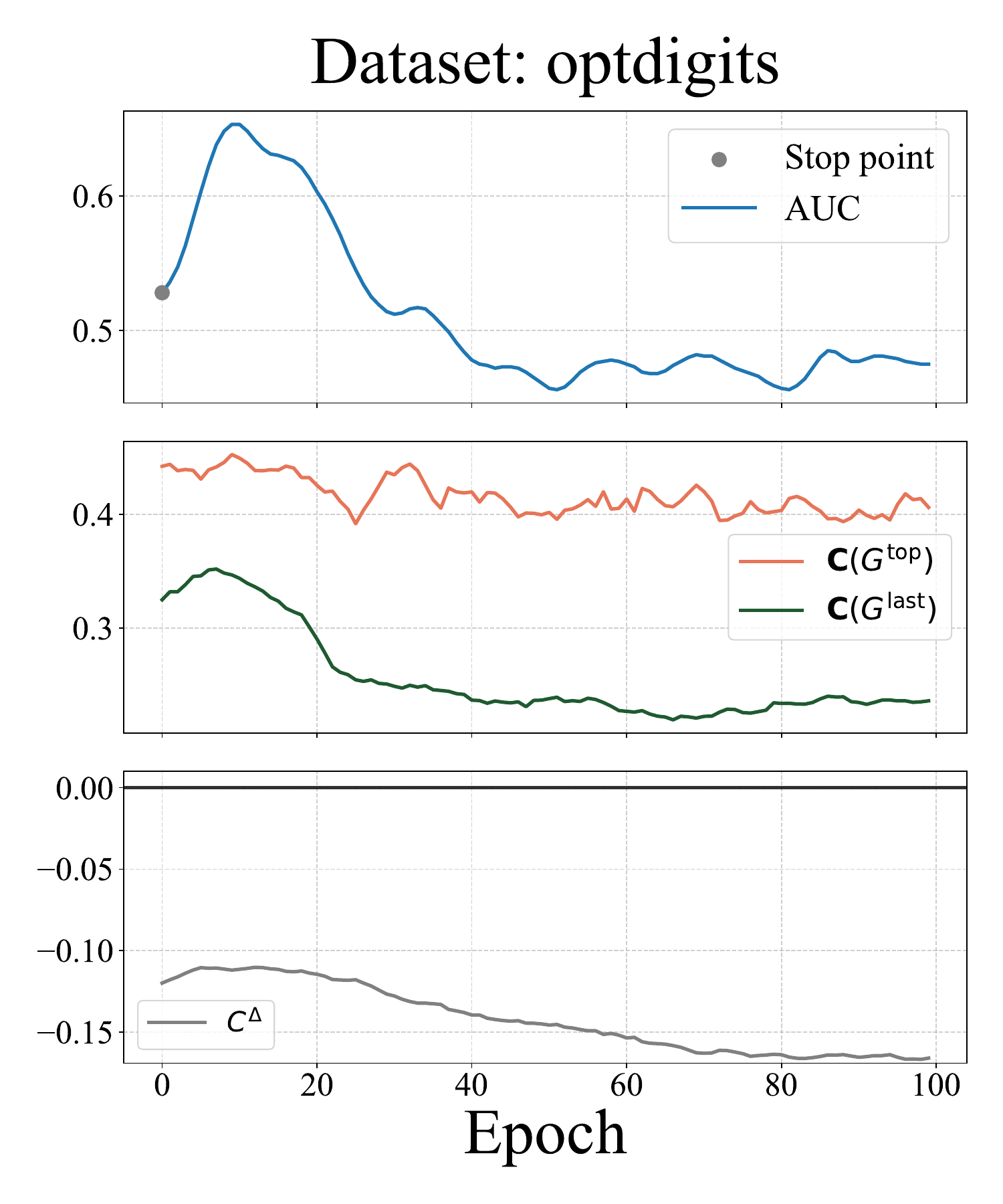}
\includegraphics[width=0.32\textwidth]{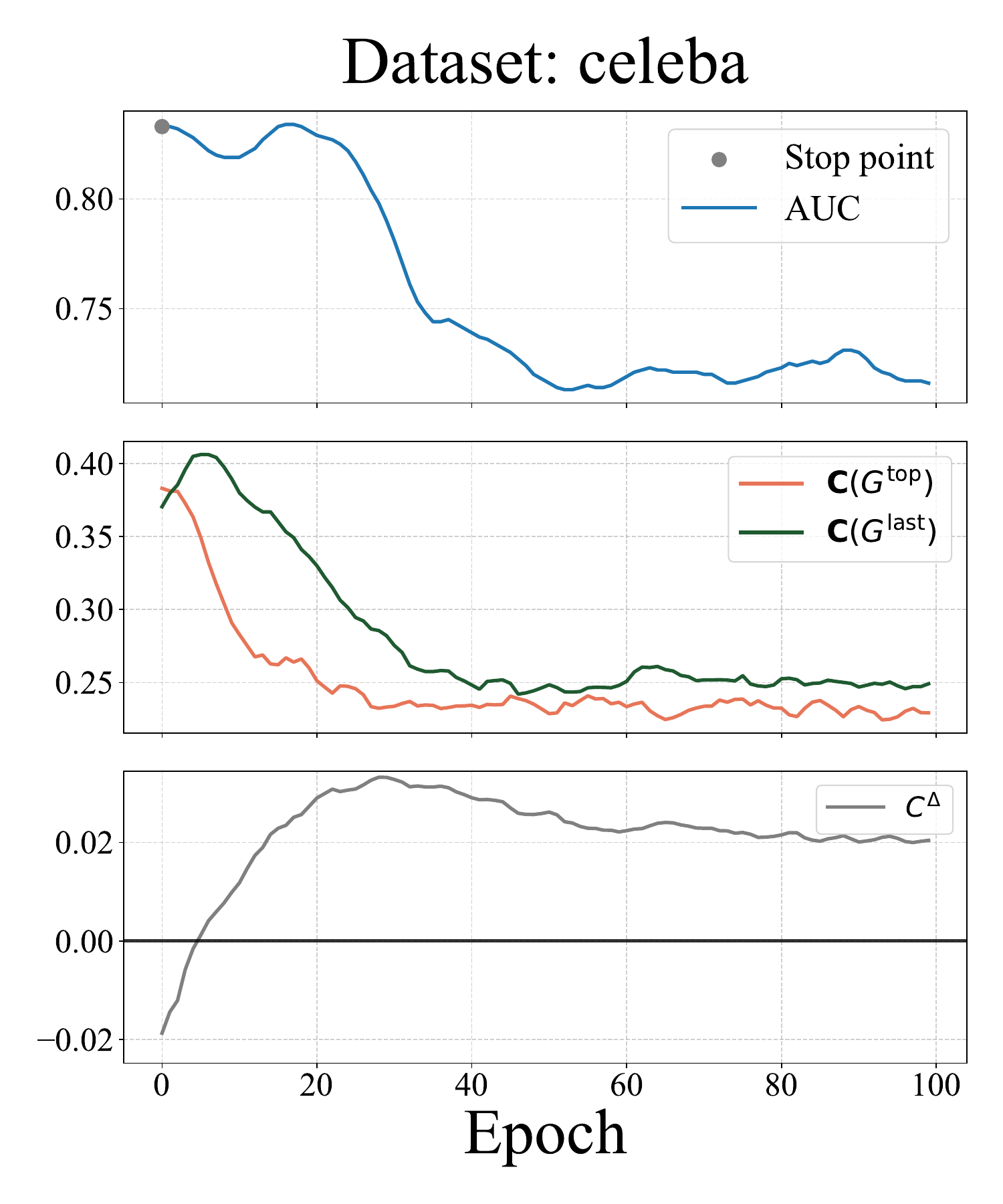}
\includegraphics[width=0.32\textwidth]{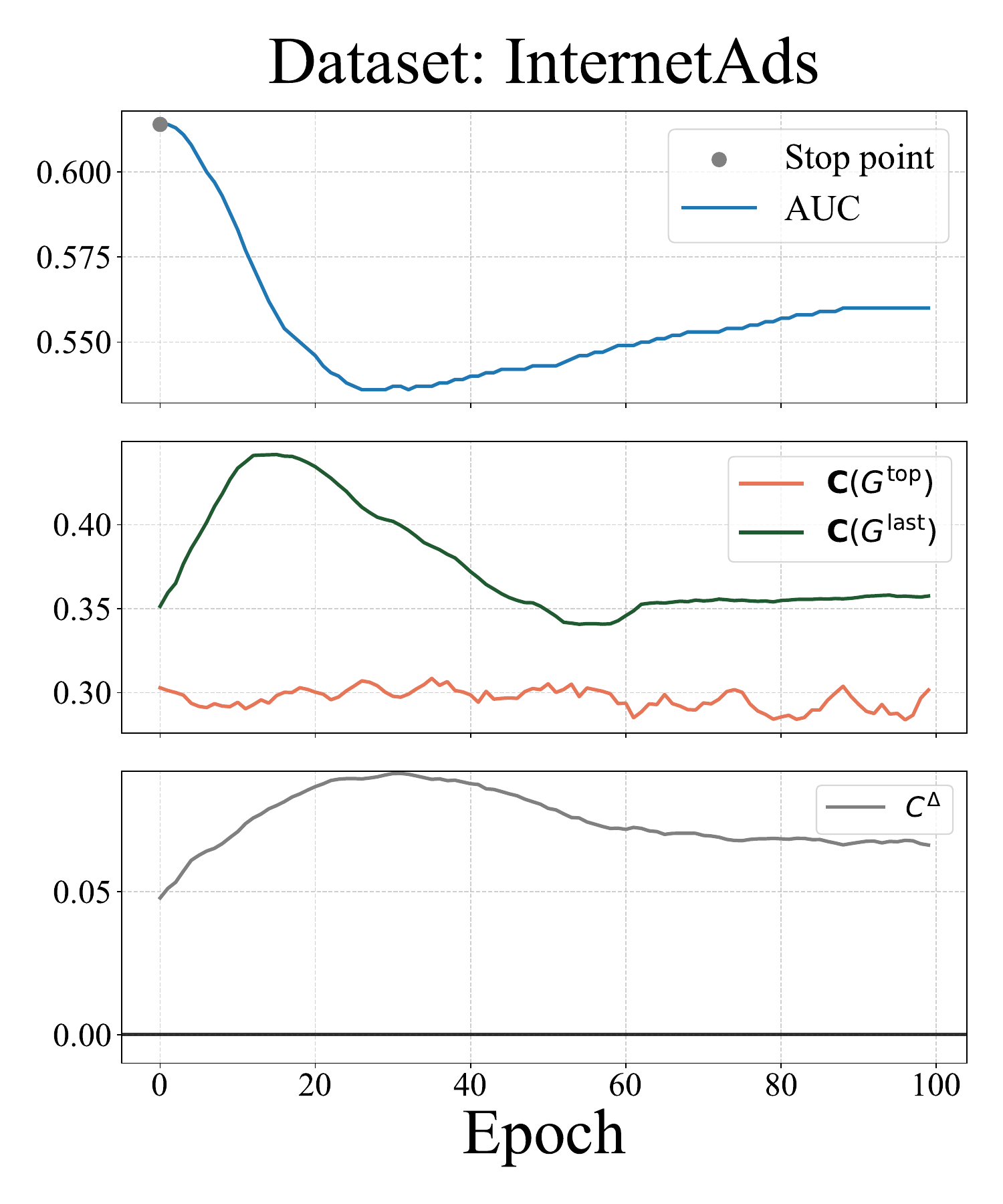}
\includegraphics[width=0.32\textwidth]{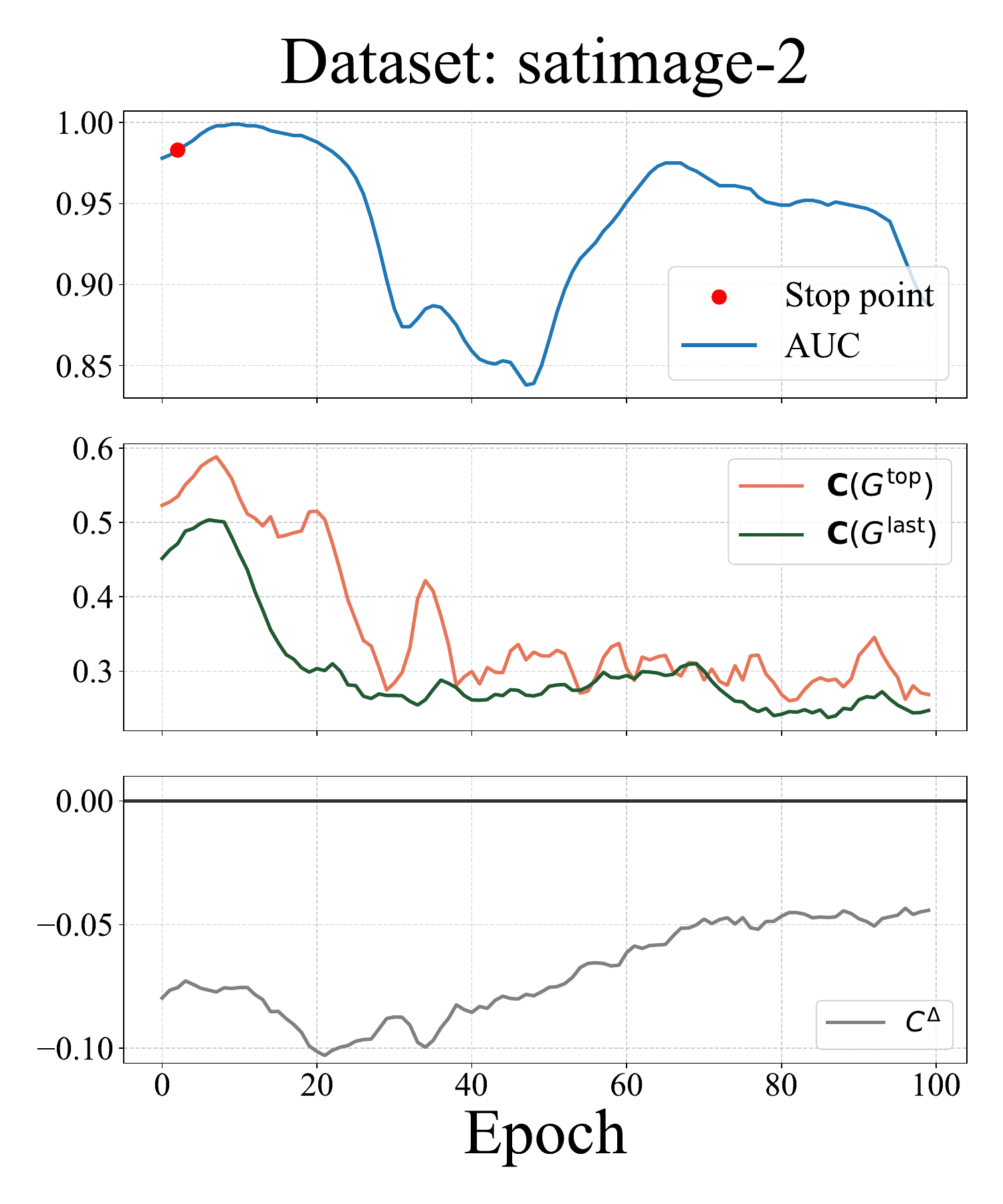}
  \caption{AE: AUC curves vs. $\mathbf{C^\Delta}$ curves. Top: AUC. Middle: $\mathbf{C}(G^{\text{last}})$ and $\mathbf{C}(G^{\text{top}})$. Bottom: $C^{\Delta}=\mathbf{C}(G^{\text{last}})-\mathbf{C}(G^{\text{top}})$.}
  \label{Fig:all-curve-1}
\end{figure}

\begin{figure}[ht]
  \centering
\includegraphics[width=0.32\textwidth]{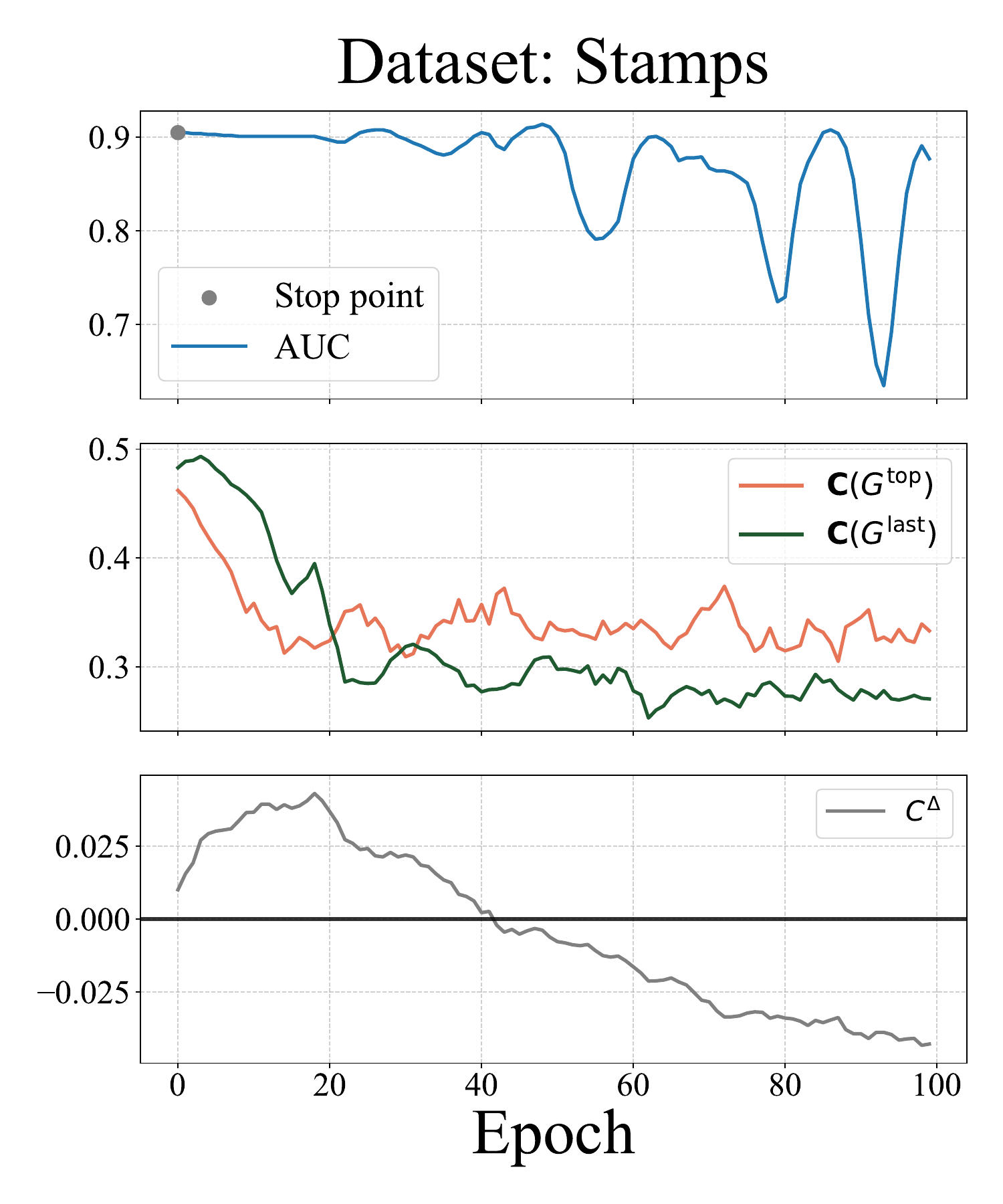}
\includegraphics[width=0.32\textwidth]{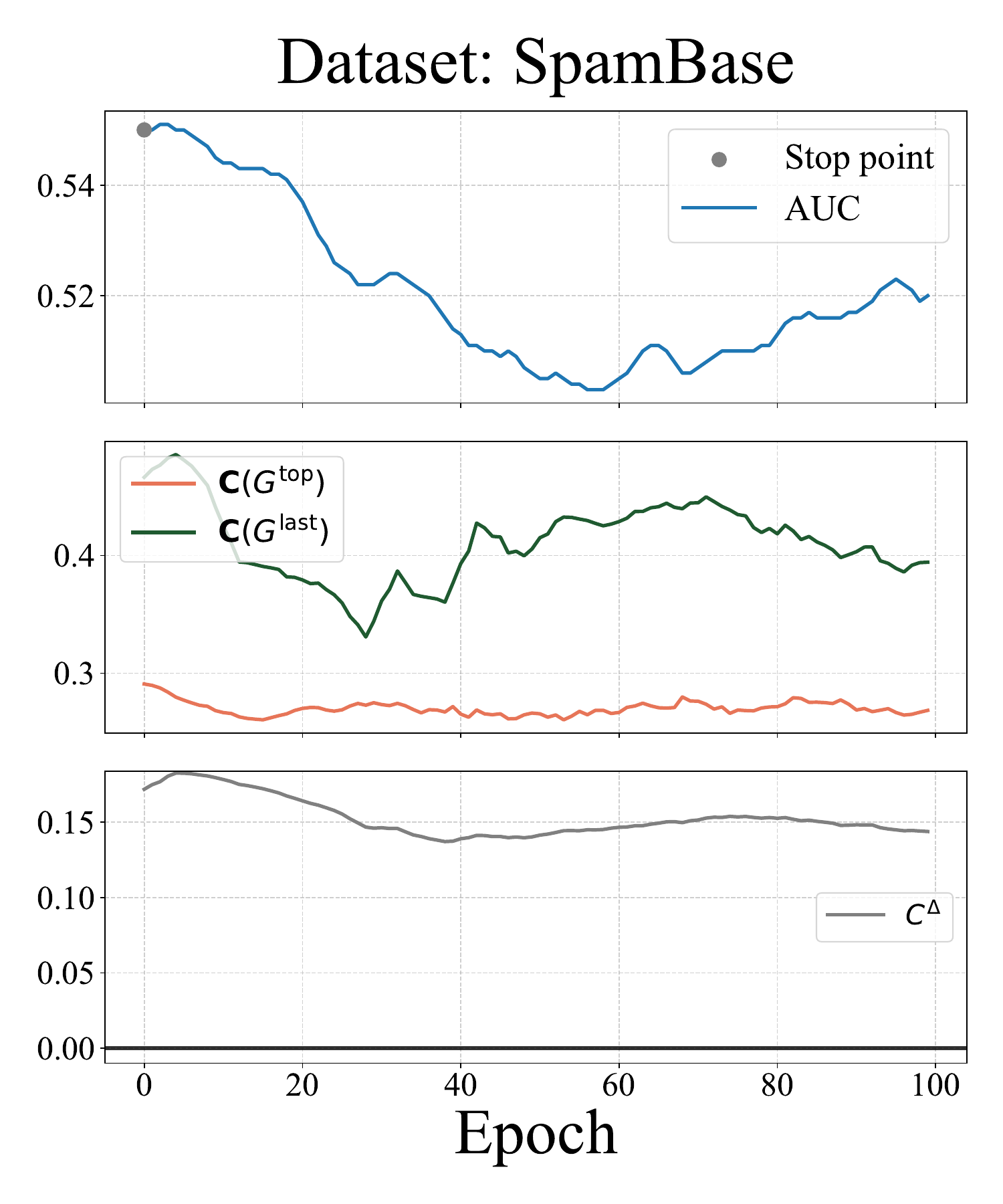}
\includegraphics[width=0.32\textwidth]{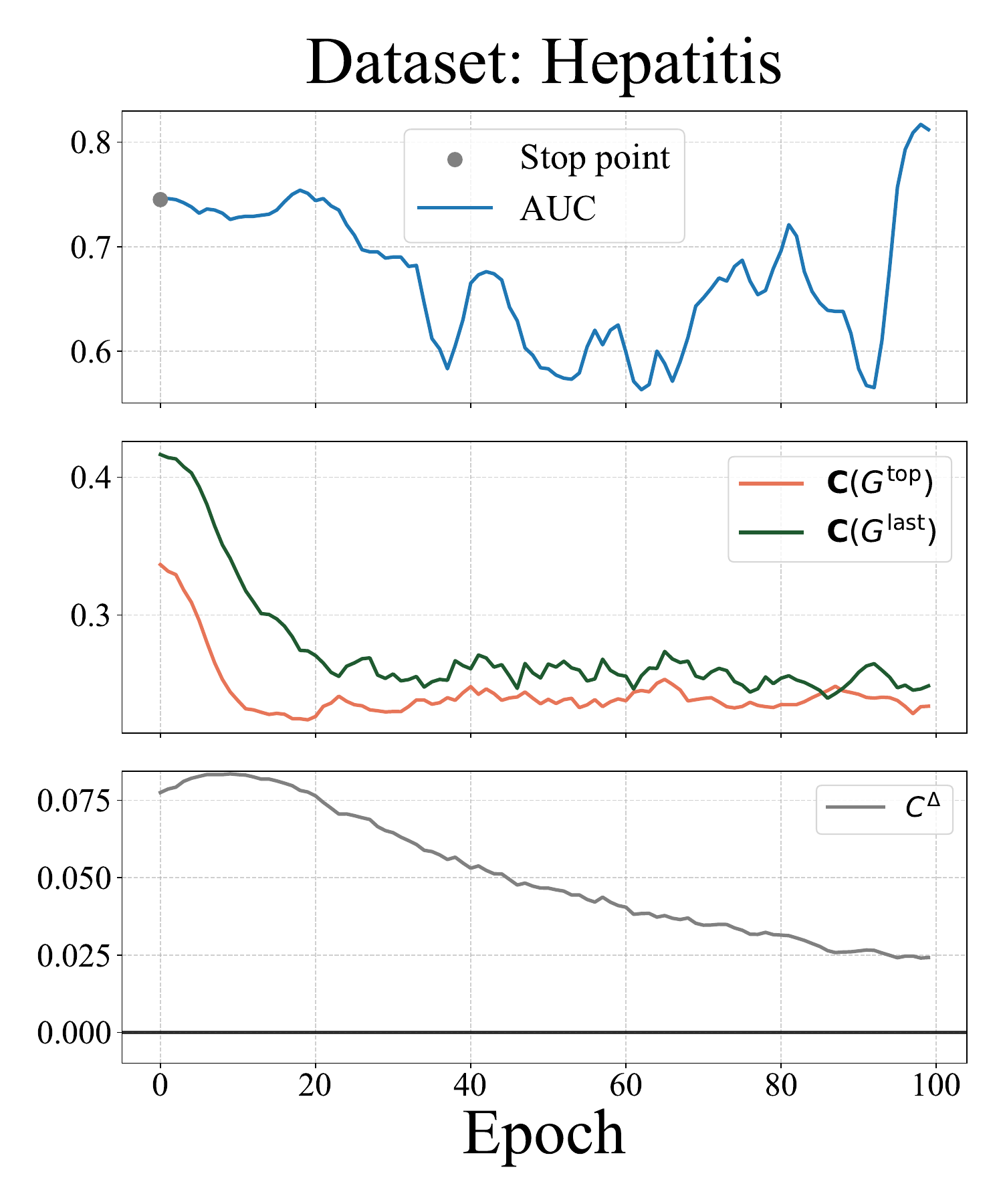}
\includegraphics[width=0.32\textwidth]{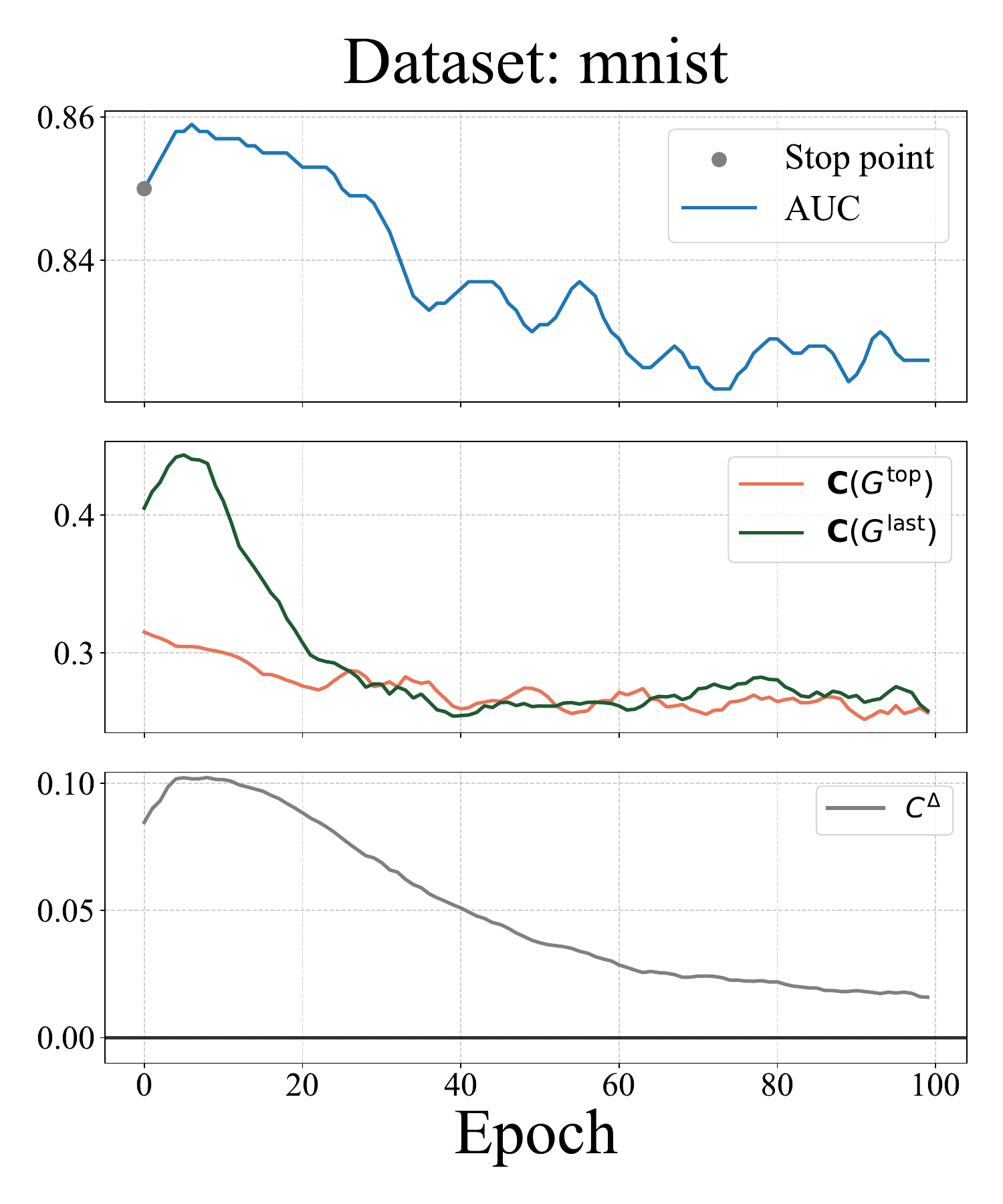}
\includegraphics[width=0.32\textwidth]{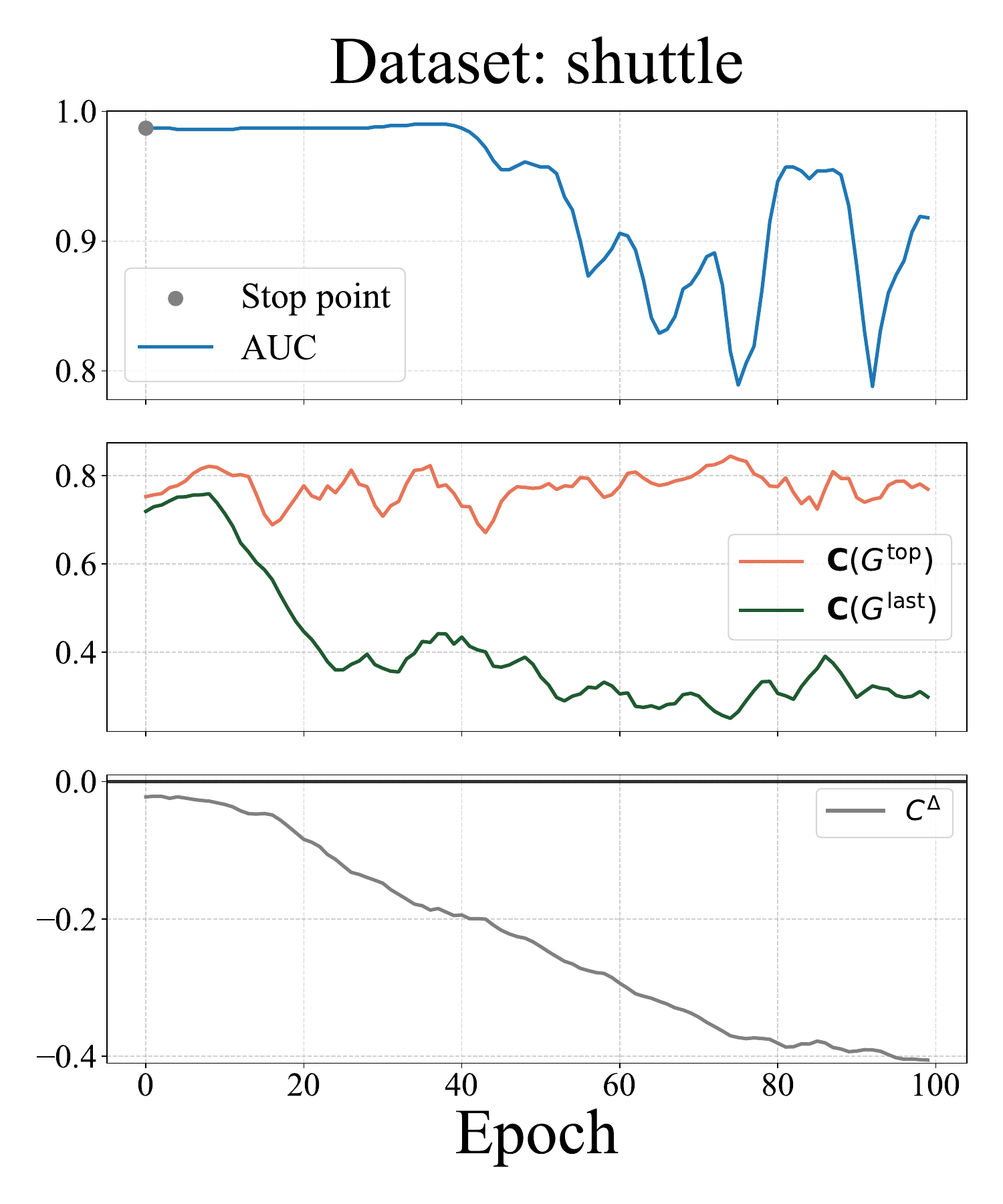}
\includegraphics[width=0.32\textwidth]{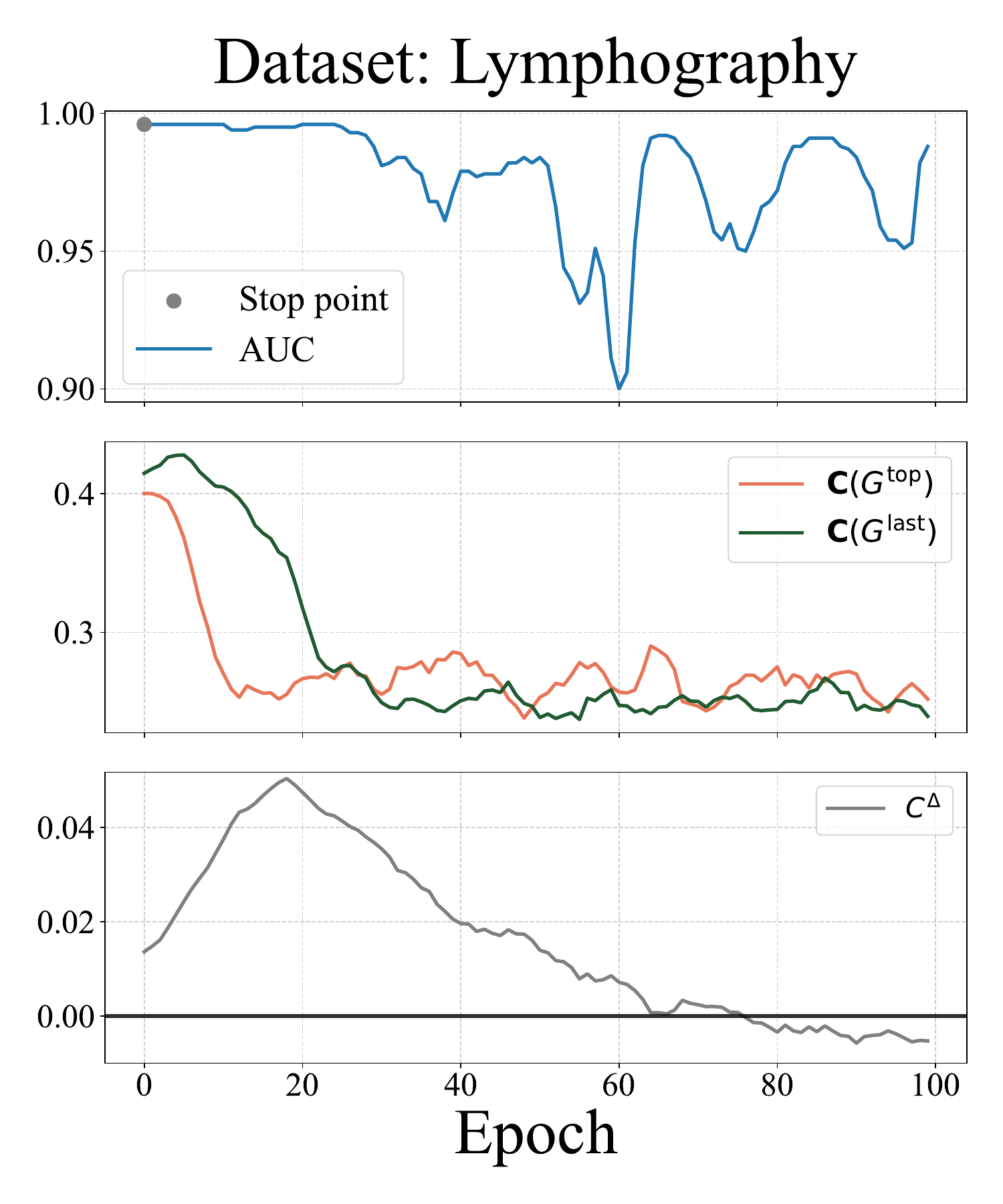}
\includegraphics[width=0.32\textwidth]{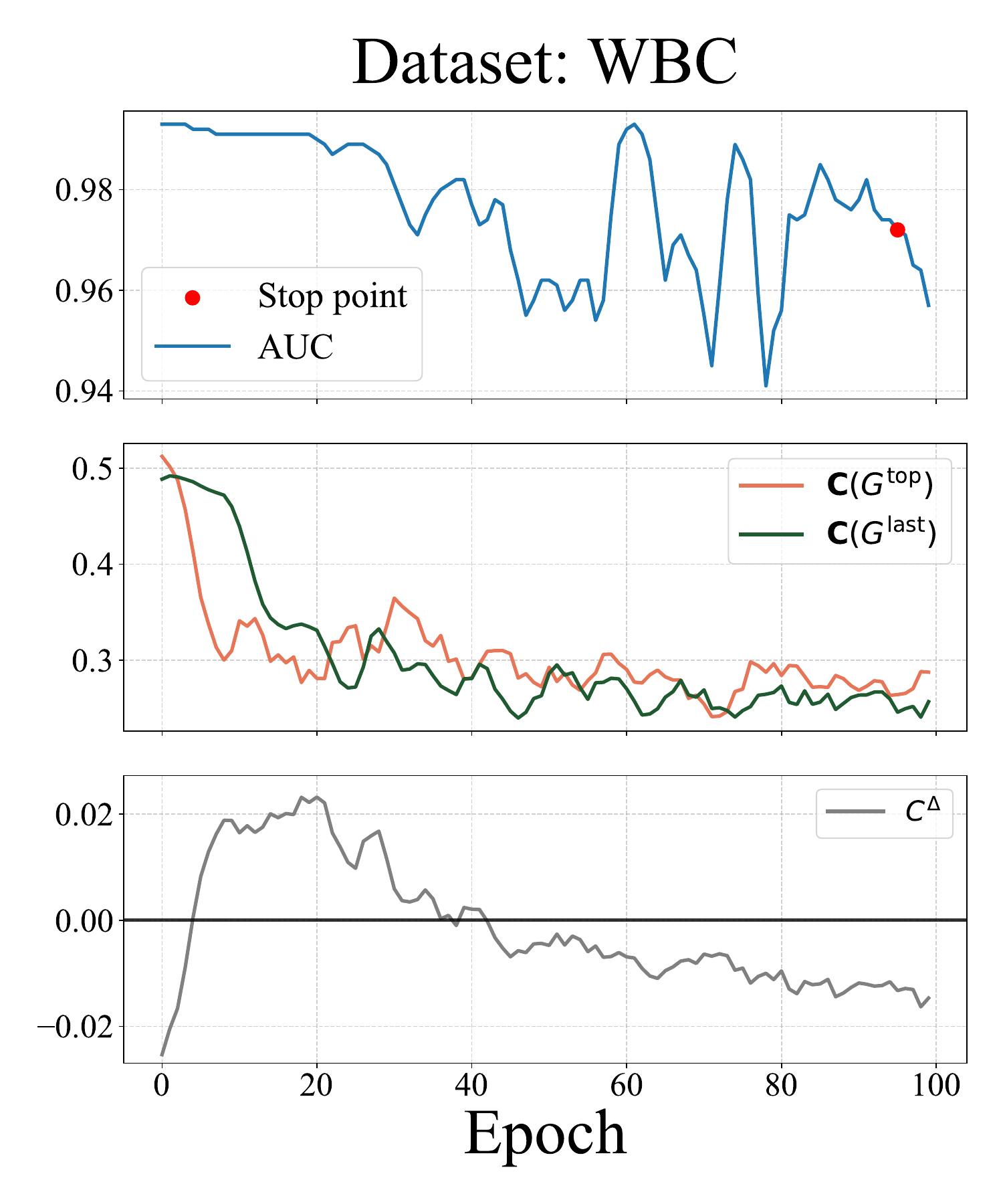}
\includegraphics[width=0.32\textwidth]{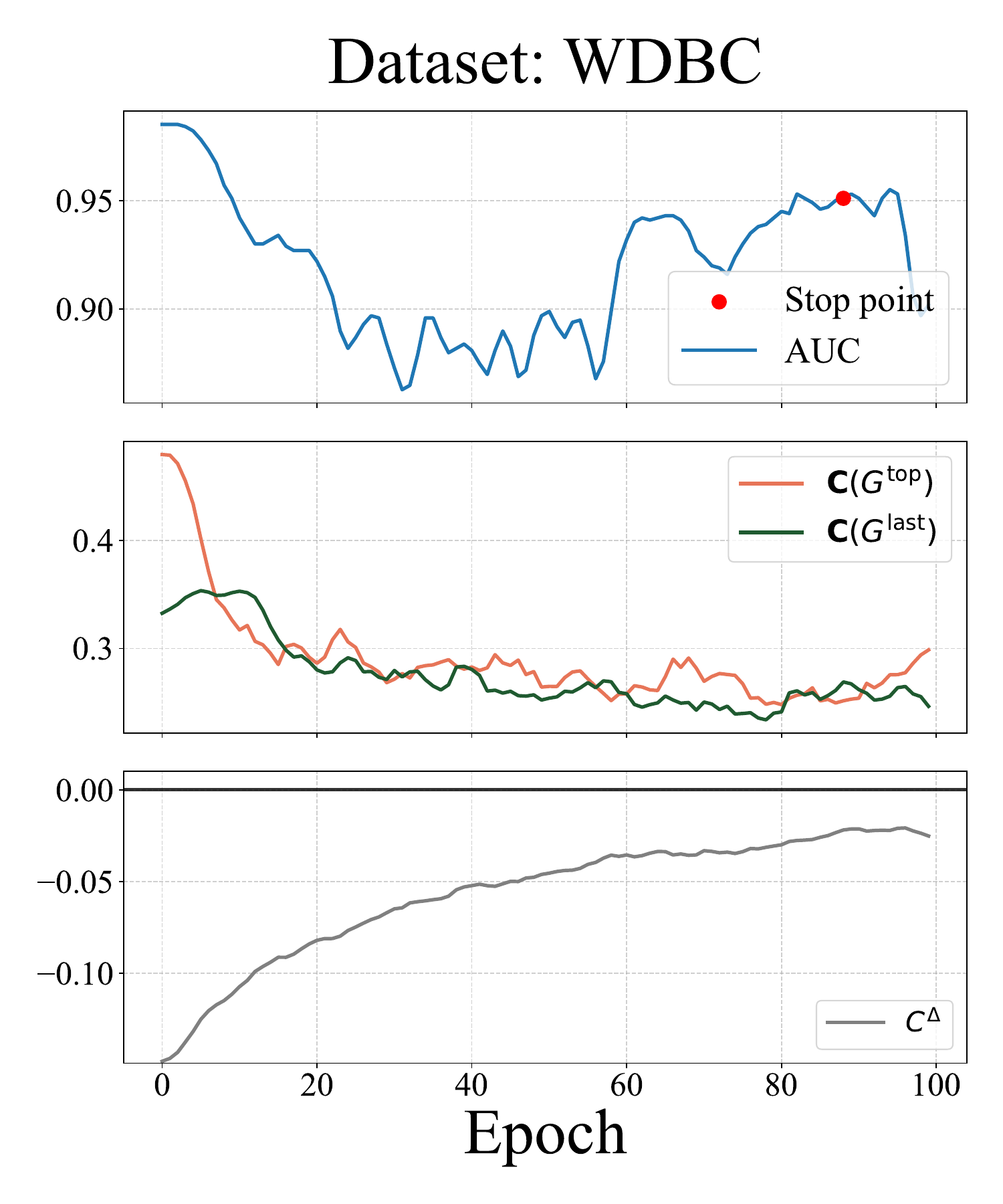}
\includegraphics[width=0.32\textwidth]{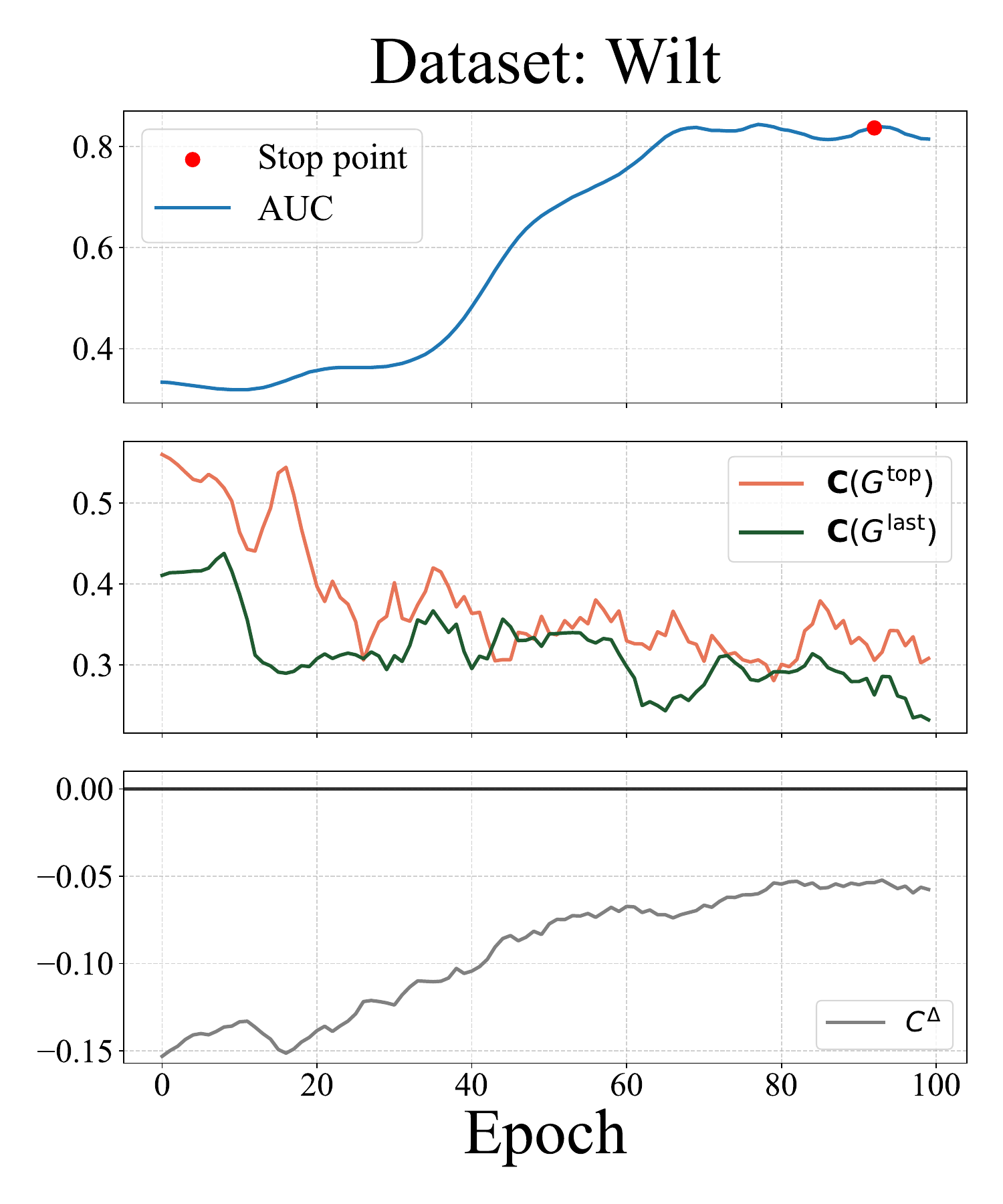}
  \caption{AE: AUC curves vs.  $\mathbf{C^\Delta}$ curves. Top: AUC. Middle: $\mathbf{C}(G^{\text{last}})$ and $\mathbf{C}(G^{\text{top}})$. Bottom: $C^{\Delta}=\mathbf{C}(G^{\text{last}})-\mathbf{C}(G^{\text{top}})$.}
  \label{Fig:all-curve-2}
\end{figure}

\begin{figure}[ht]
  \centering
\includegraphics[width=0.32\textwidth]{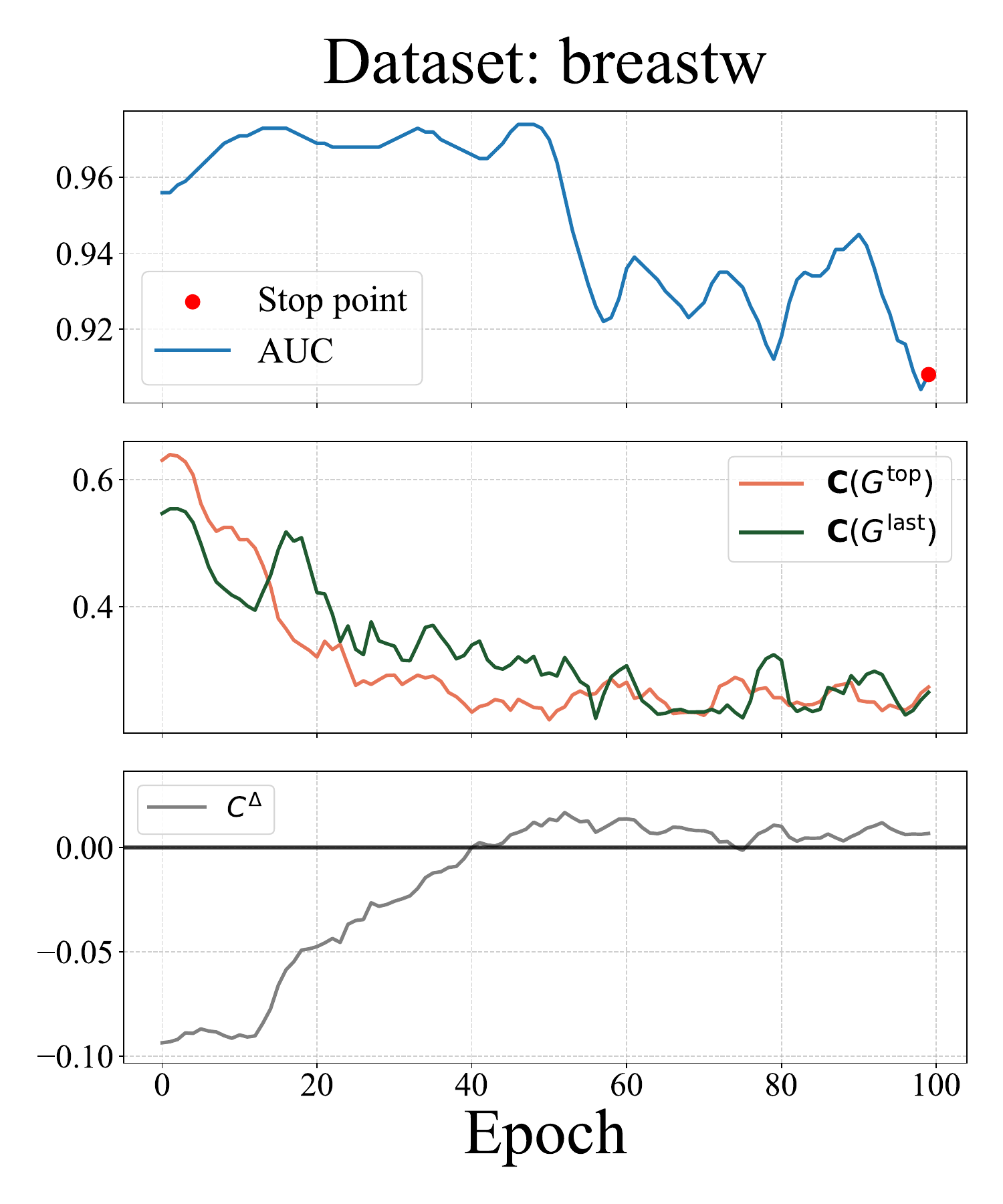}
\includegraphics[width=0.32\textwidth]{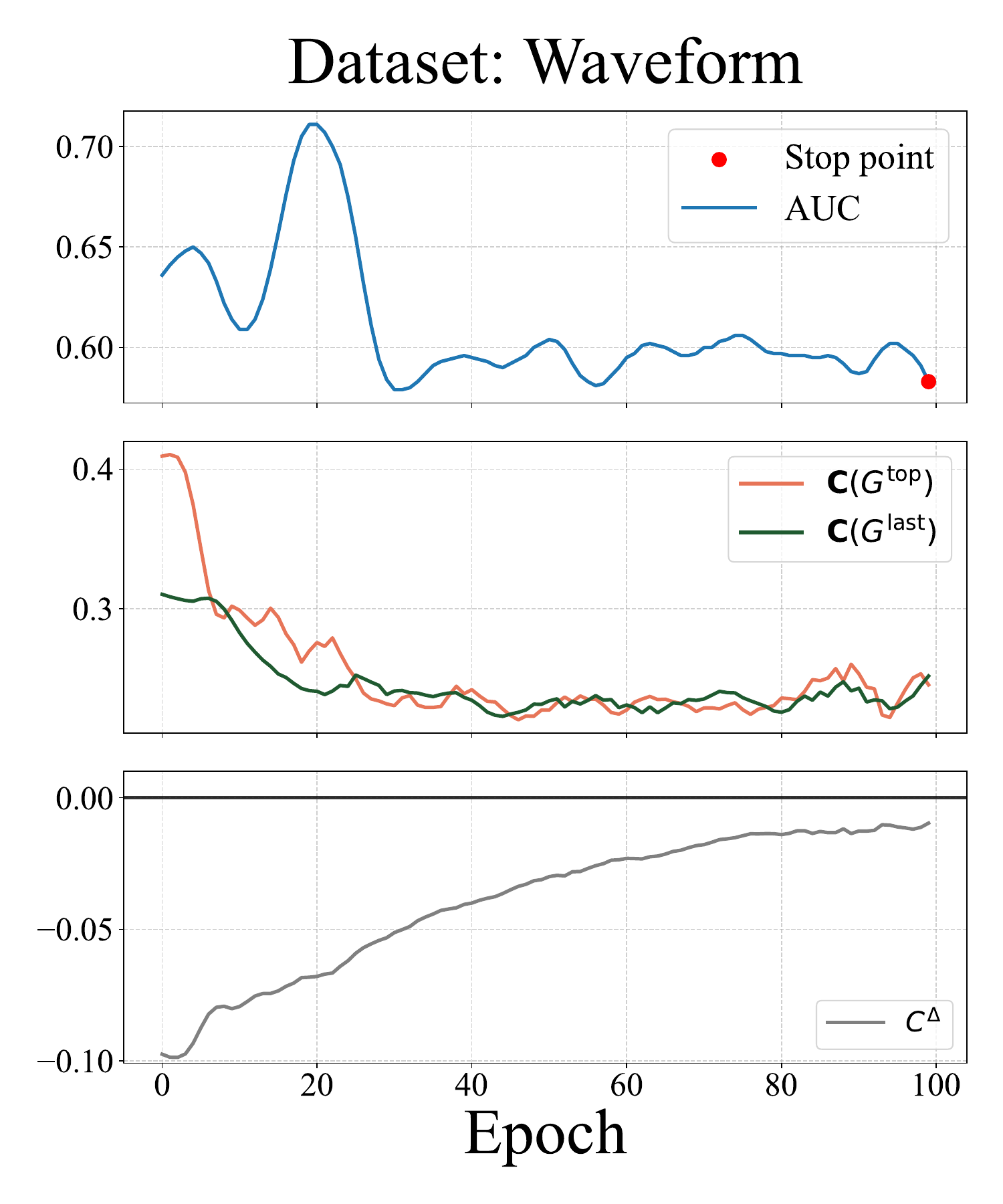}
\includegraphics[width=0.32\textwidth]{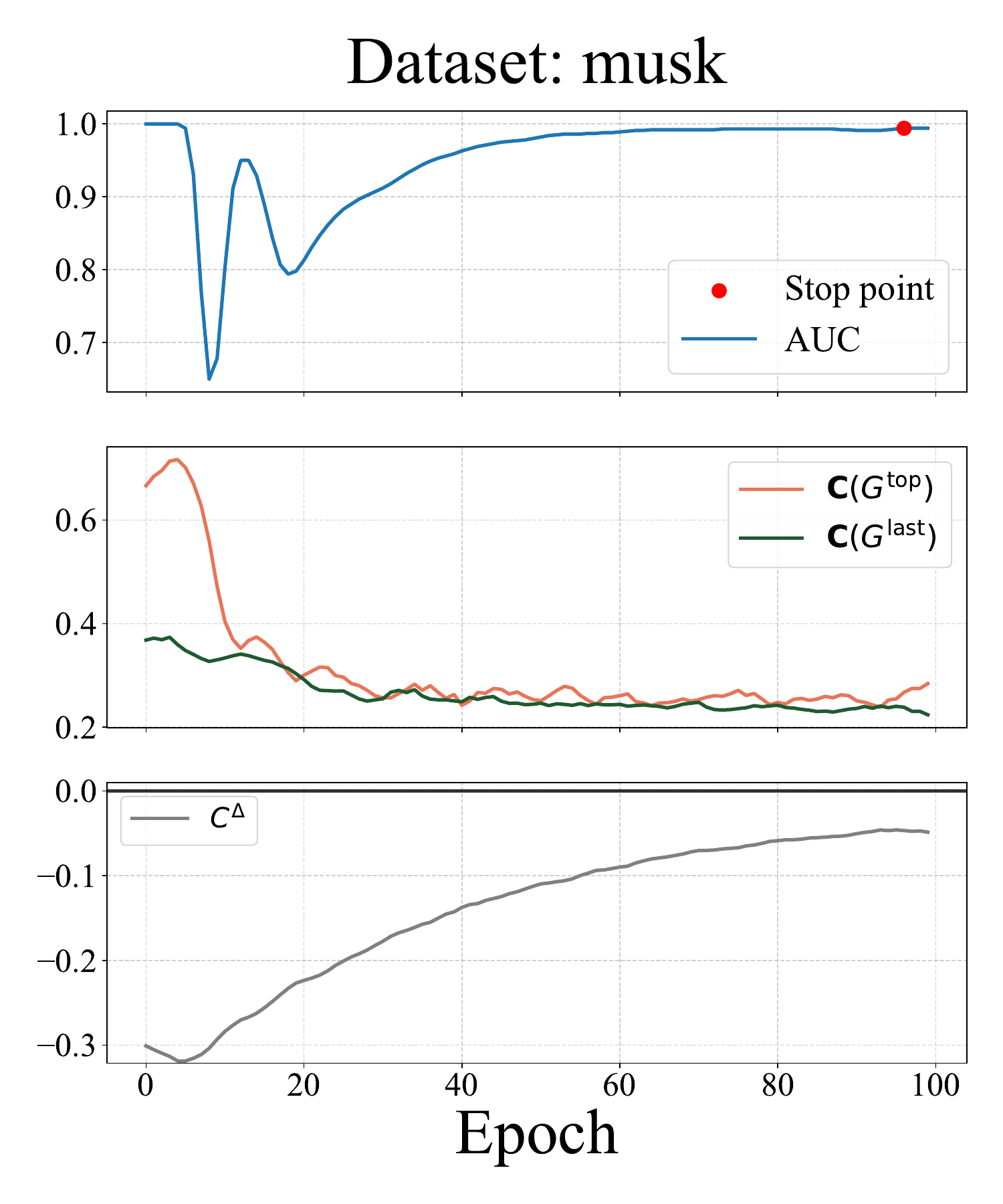}
\includegraphics[width=0.32\textwidth]{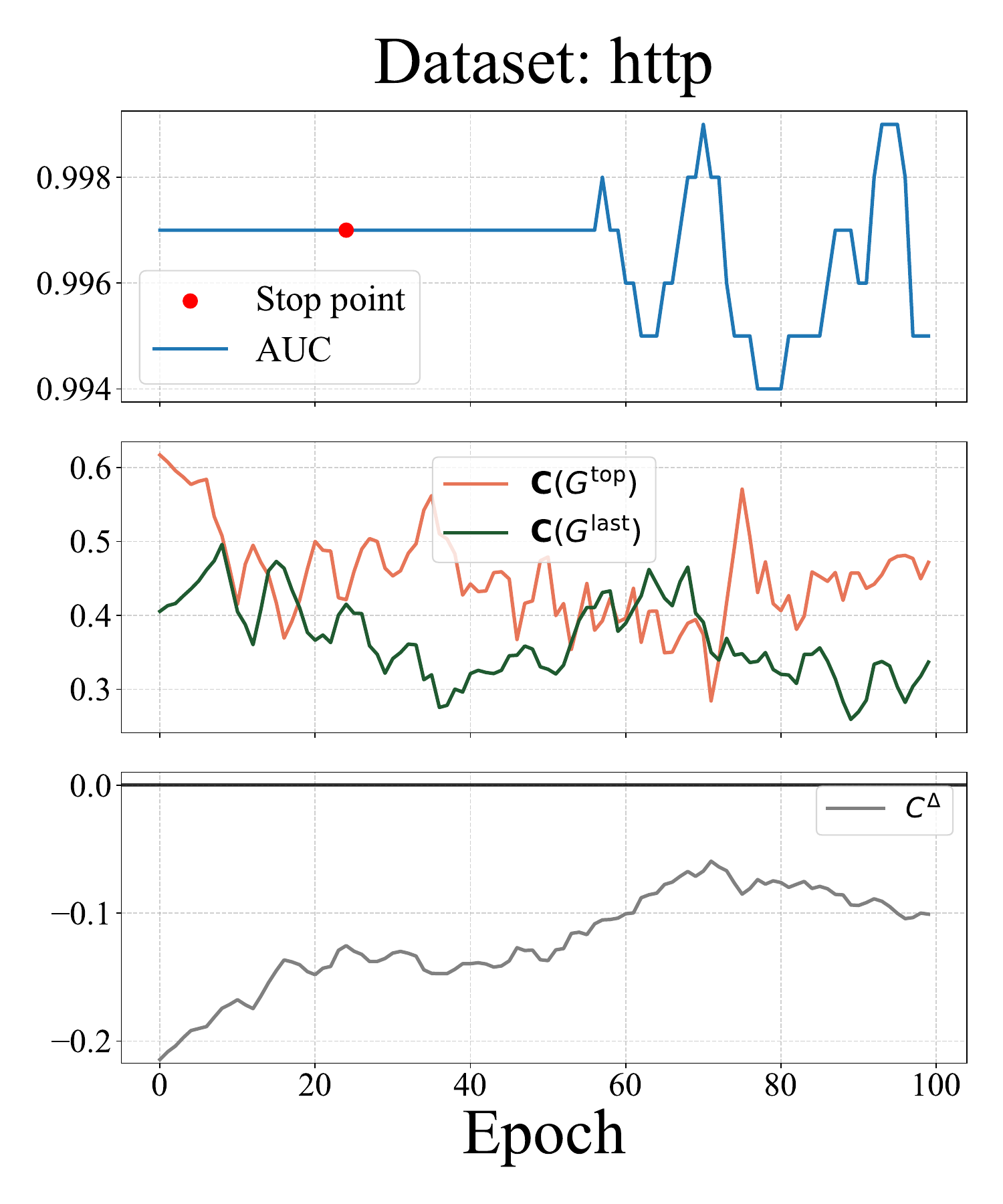}
\includegraphics[width=0.32\textwidth]{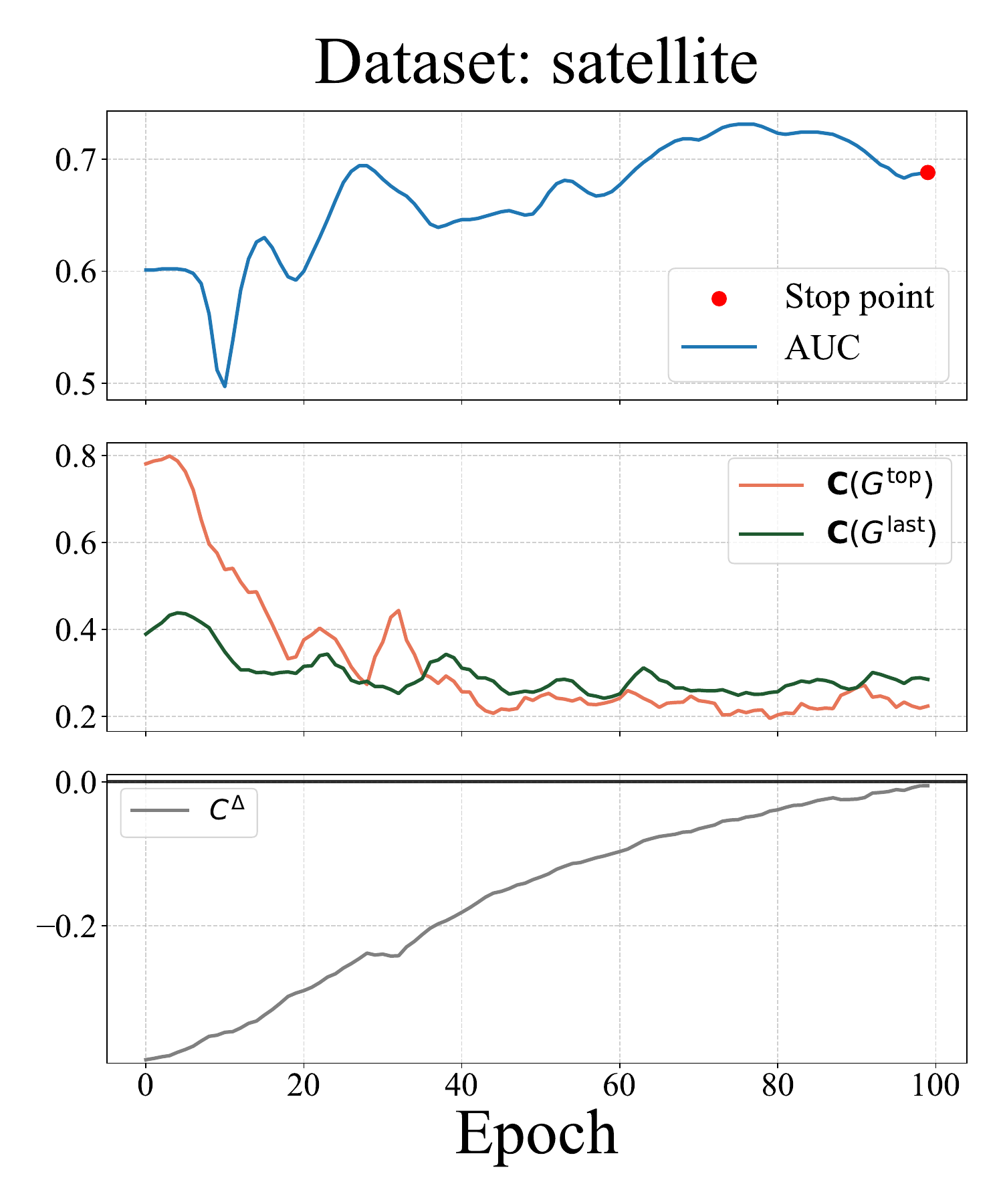}
\includegraphics[width=0.32\textwidth]{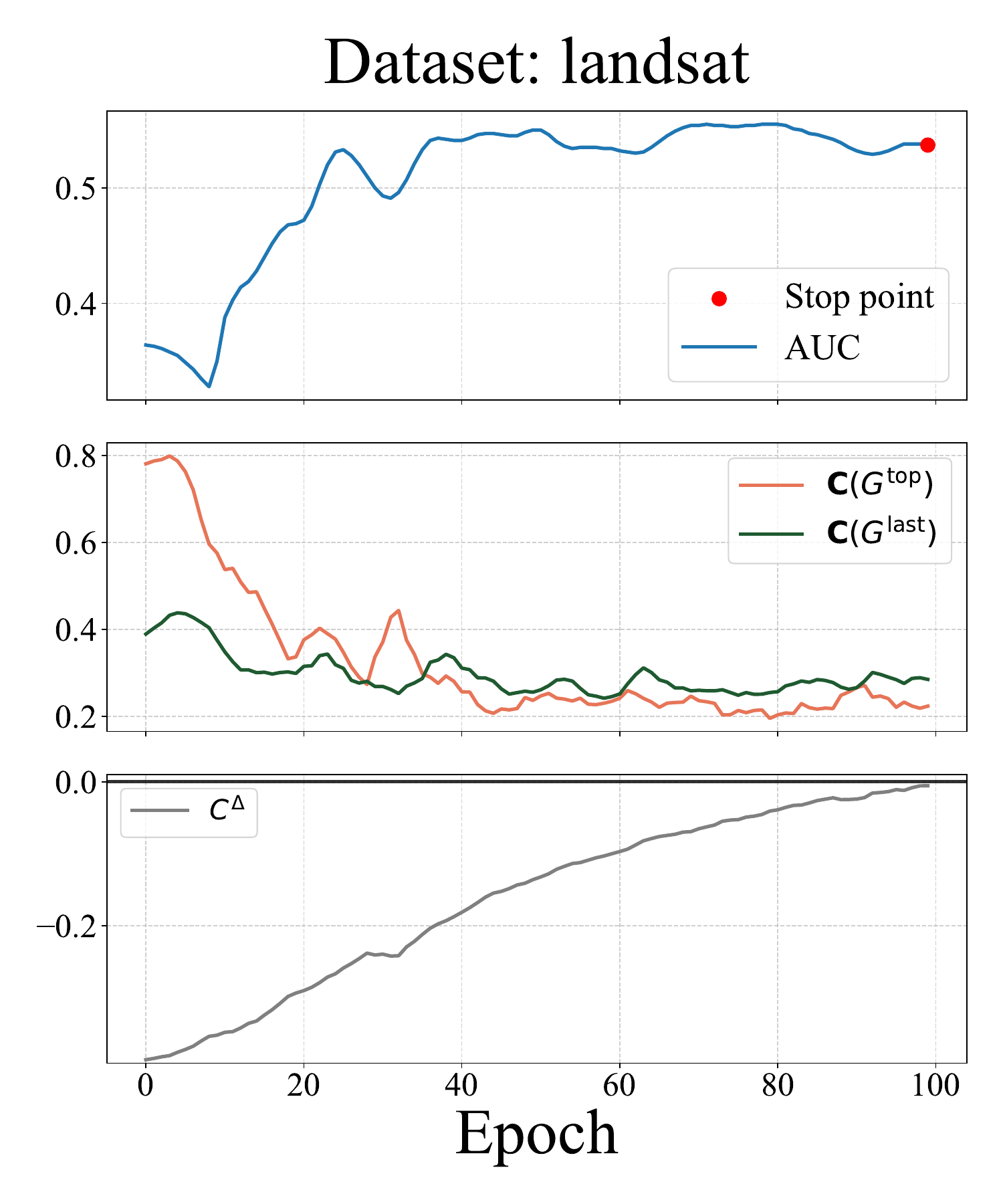}
\includegraphics[width=0.32\textwidth]{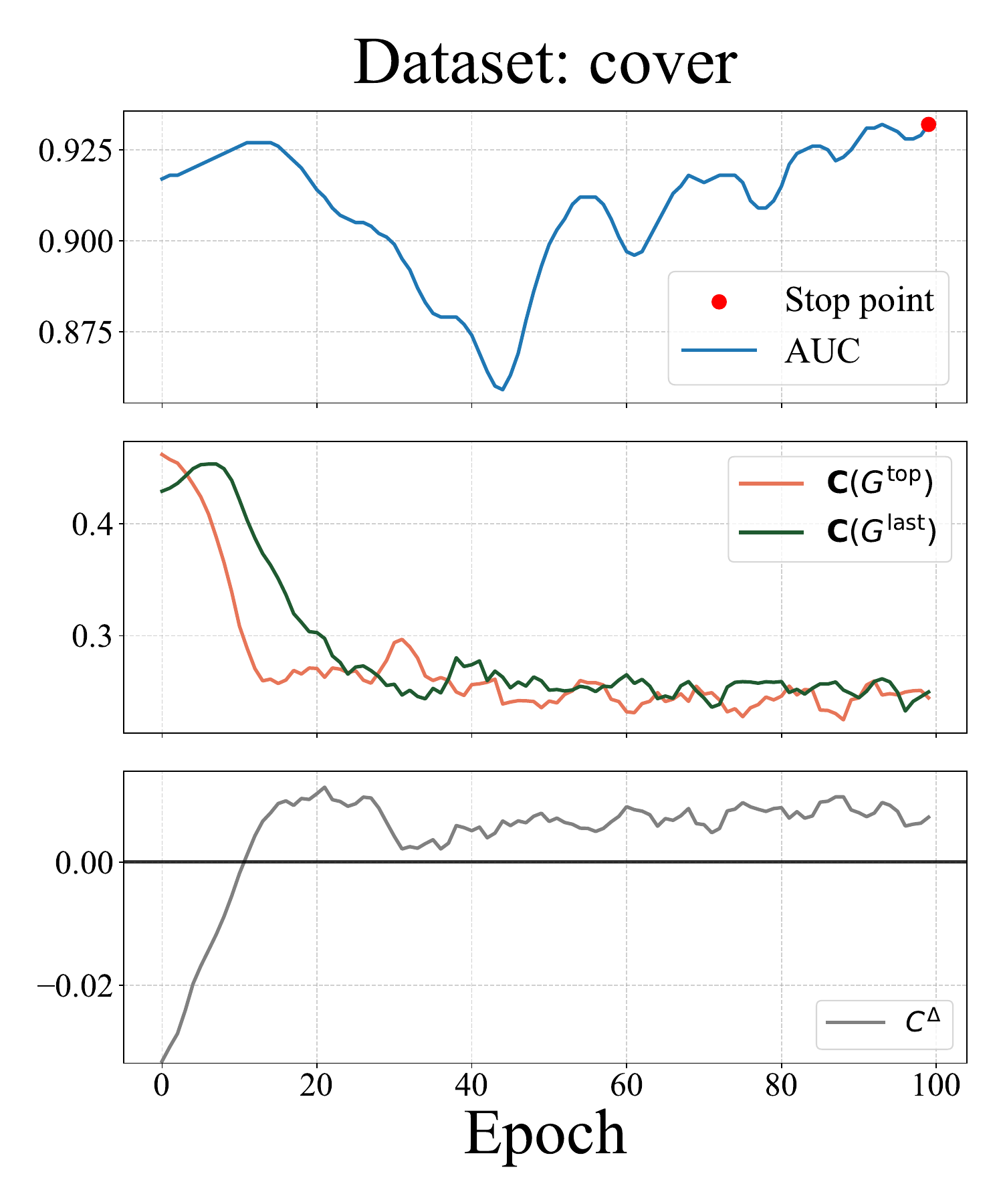}
\includegraphics[width=0.32\textwidth]{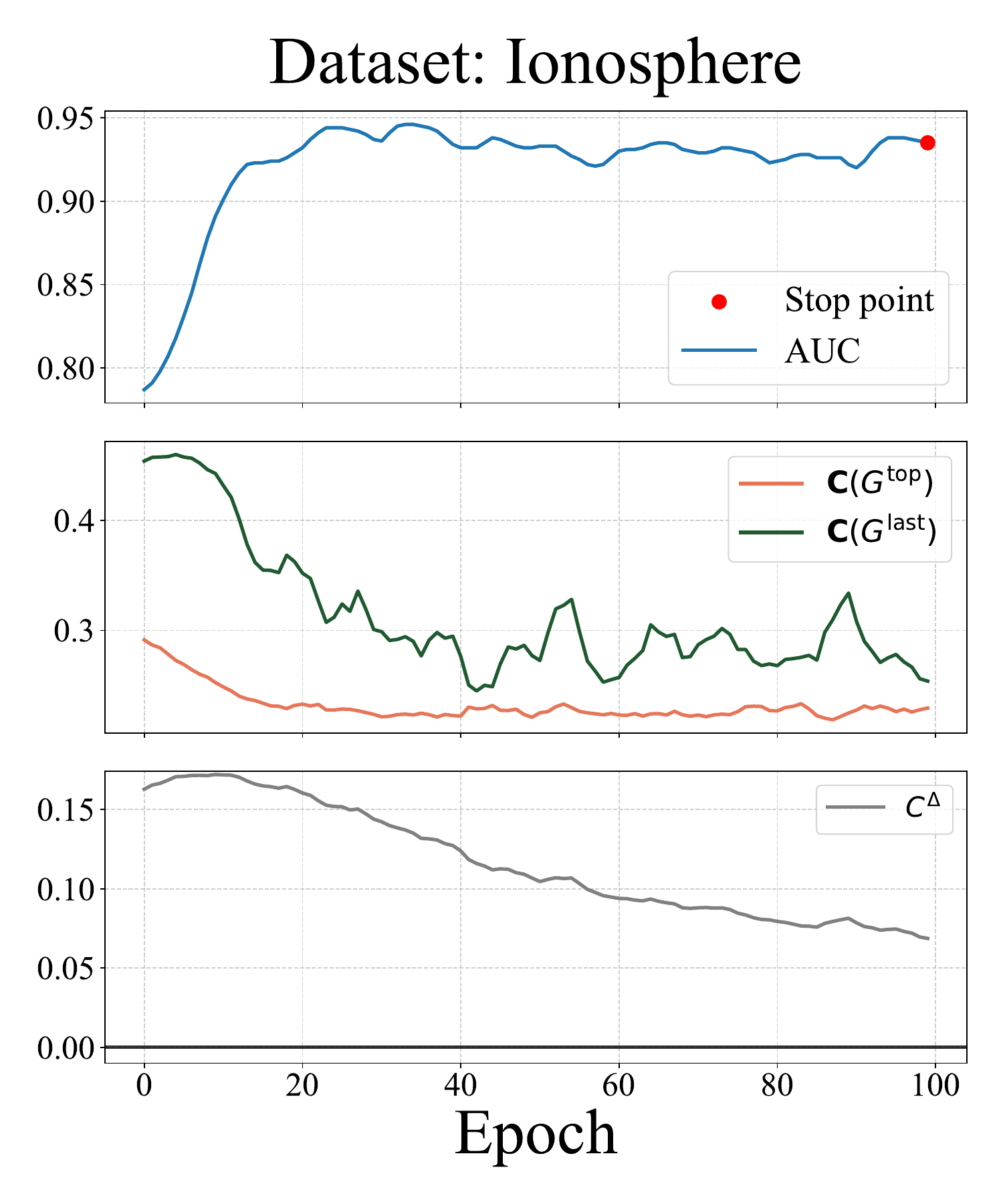}
\includegraphics[width=0.32\textwidth]{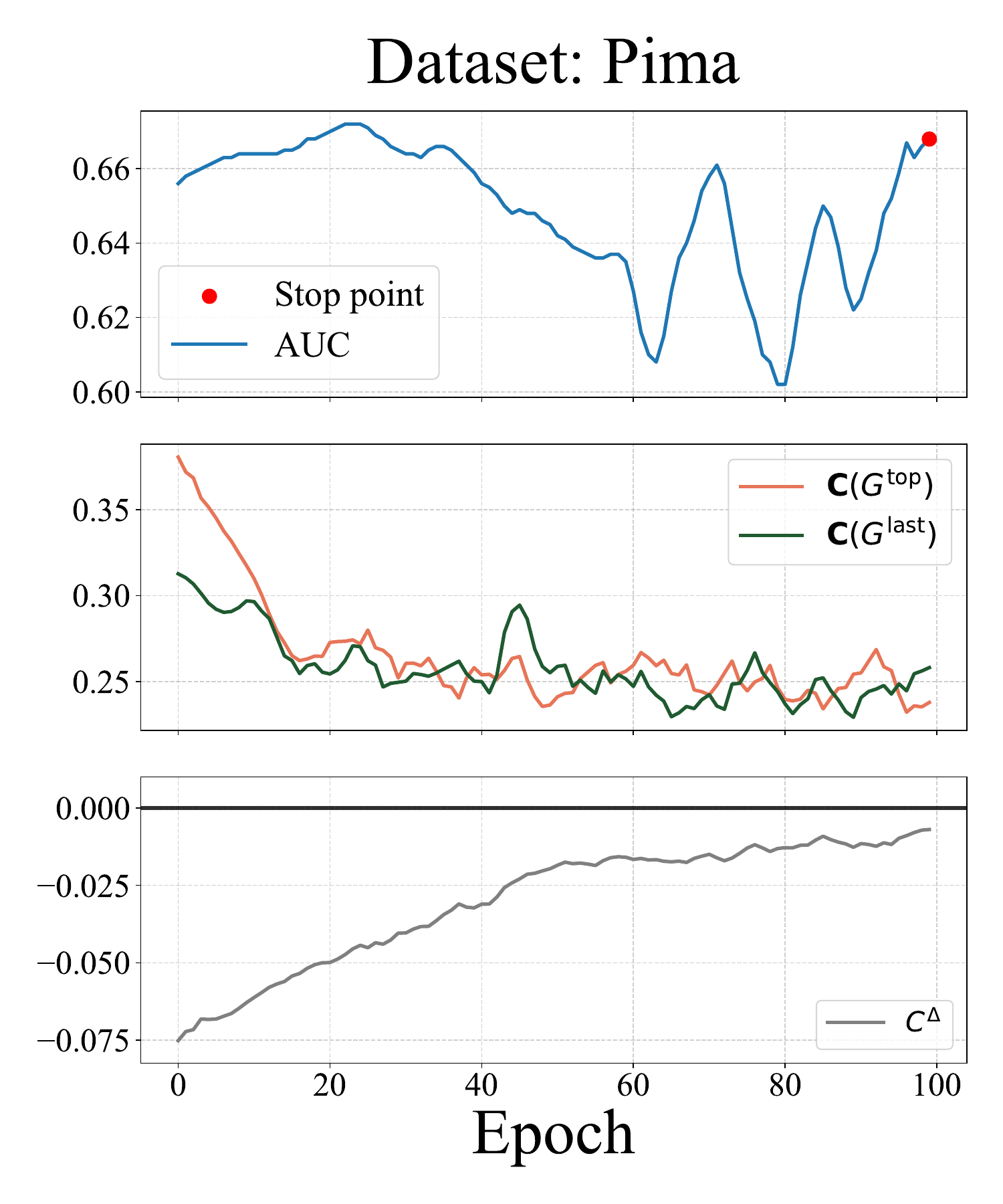}
  \caption{AE: AUC curves vs.  $\mathbf{C^\Delta}$ curves. Top: AUC. Middle: $\mathbf{C}(G^{\text{last}})$ and $\mathbf{C}(G^{\text{top}})$. Bottom: $C^{\Delta}=\mathbf{C}(G^{\text{last}})-\mathbf{C}(G^{\text{top}})$.}
  \label{Fig:all-curve-3}
\end{figure}

\begin{figure}[ht]
  \centering
\includegraphics[width=0.32\textwidth]{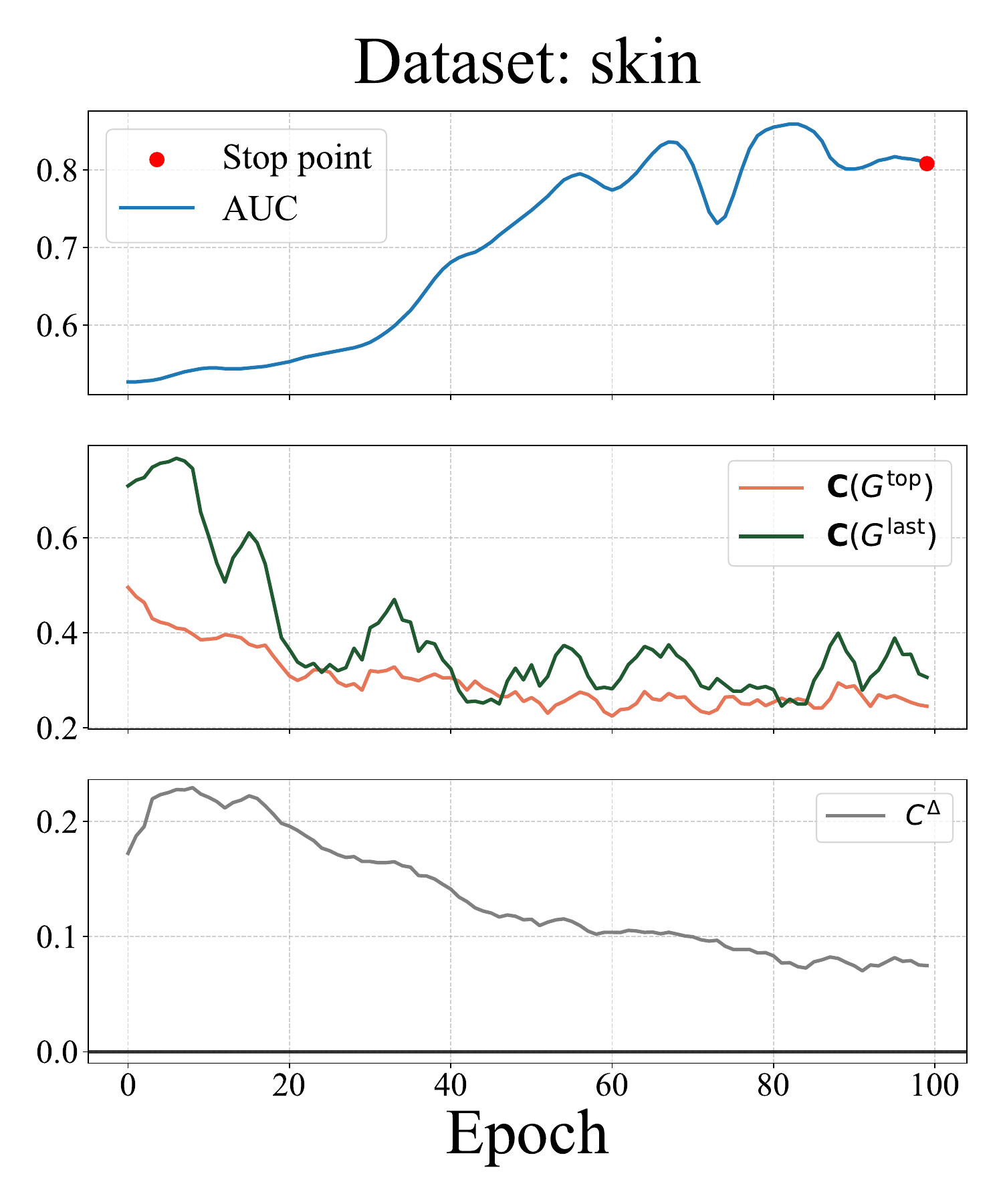}
\includegraphics[width=0.32\textwidth]{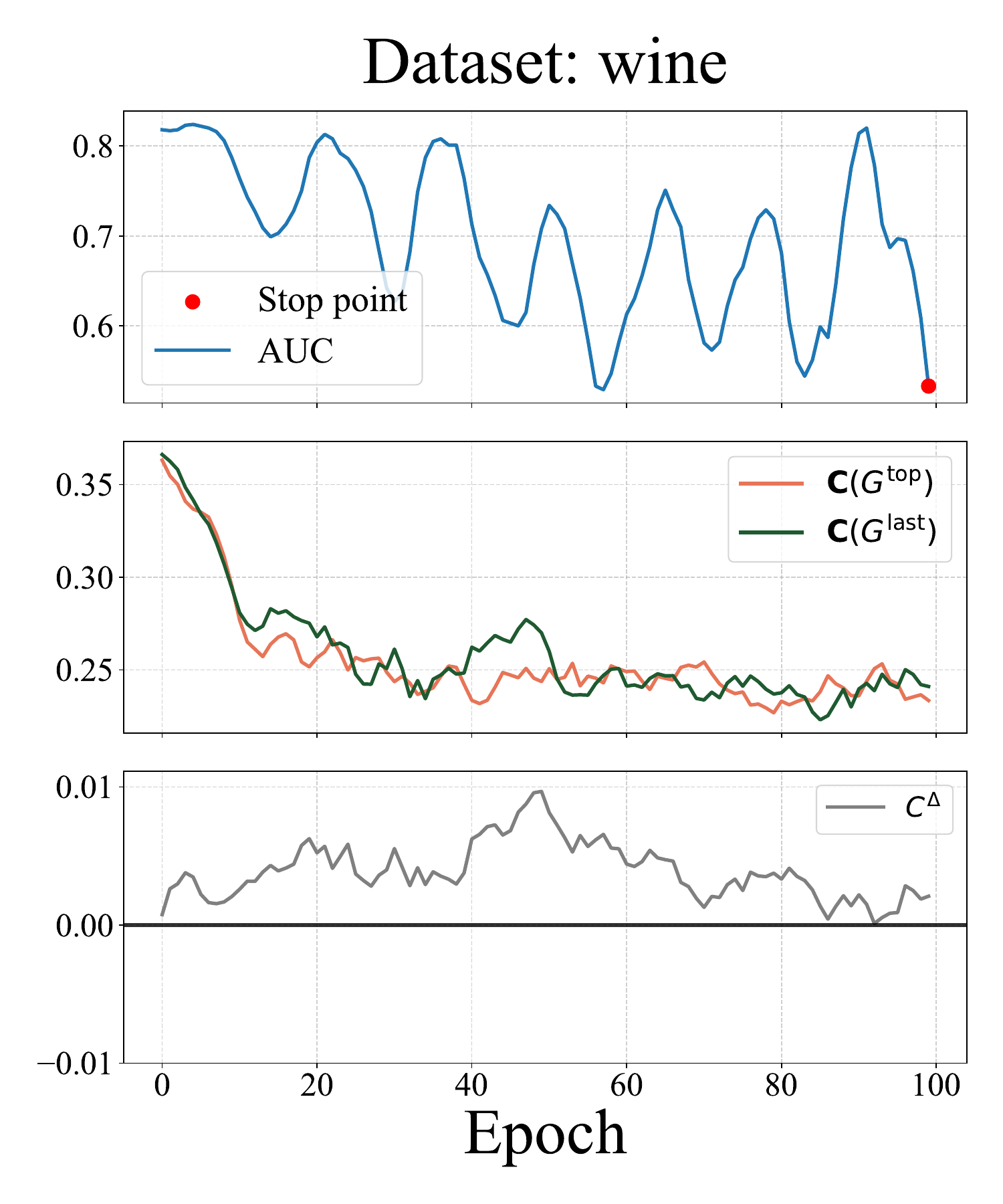}
\includegraphics[width=0.32\textwidth]{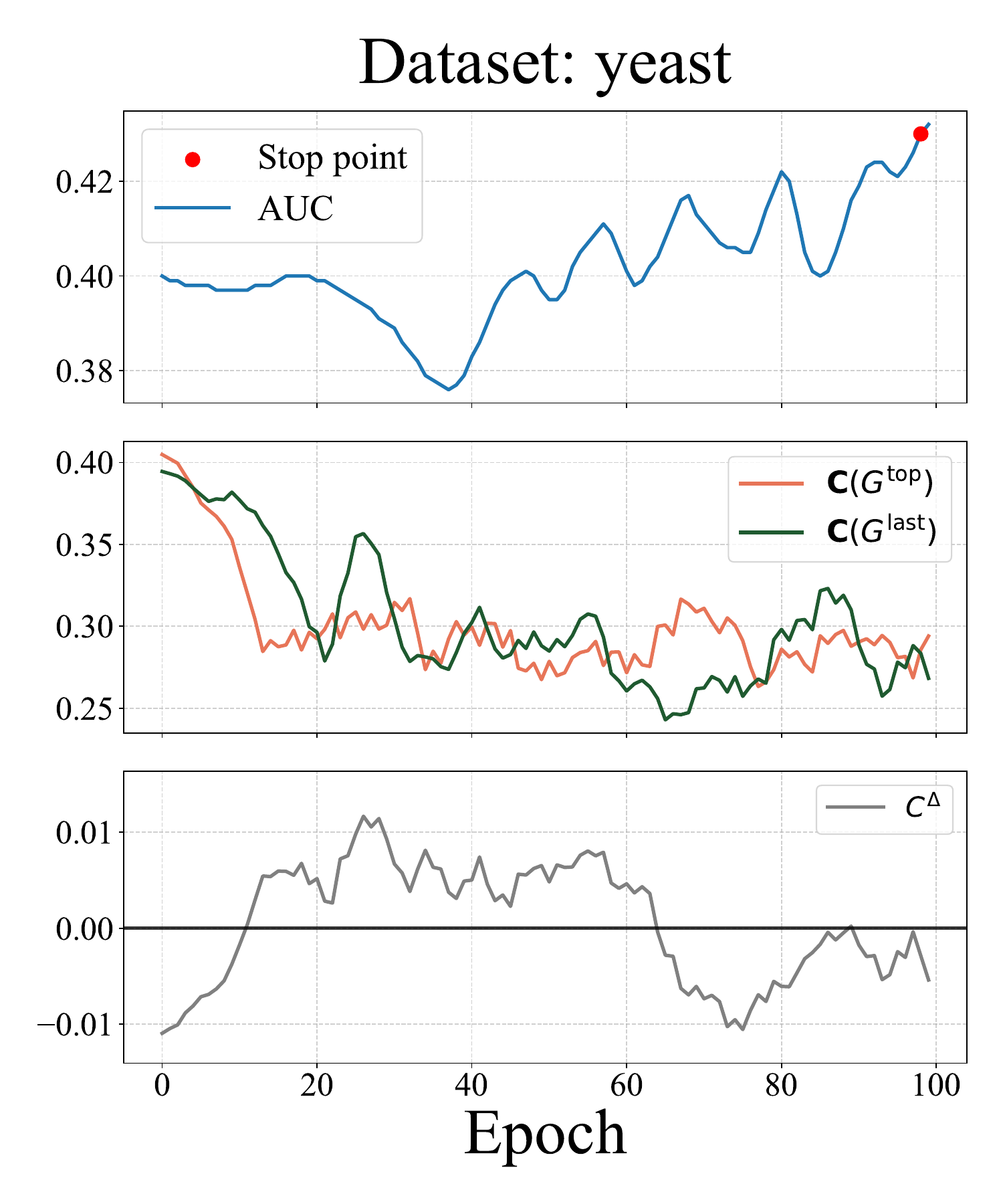}
\includegraphics[width=0.32\textwidth]{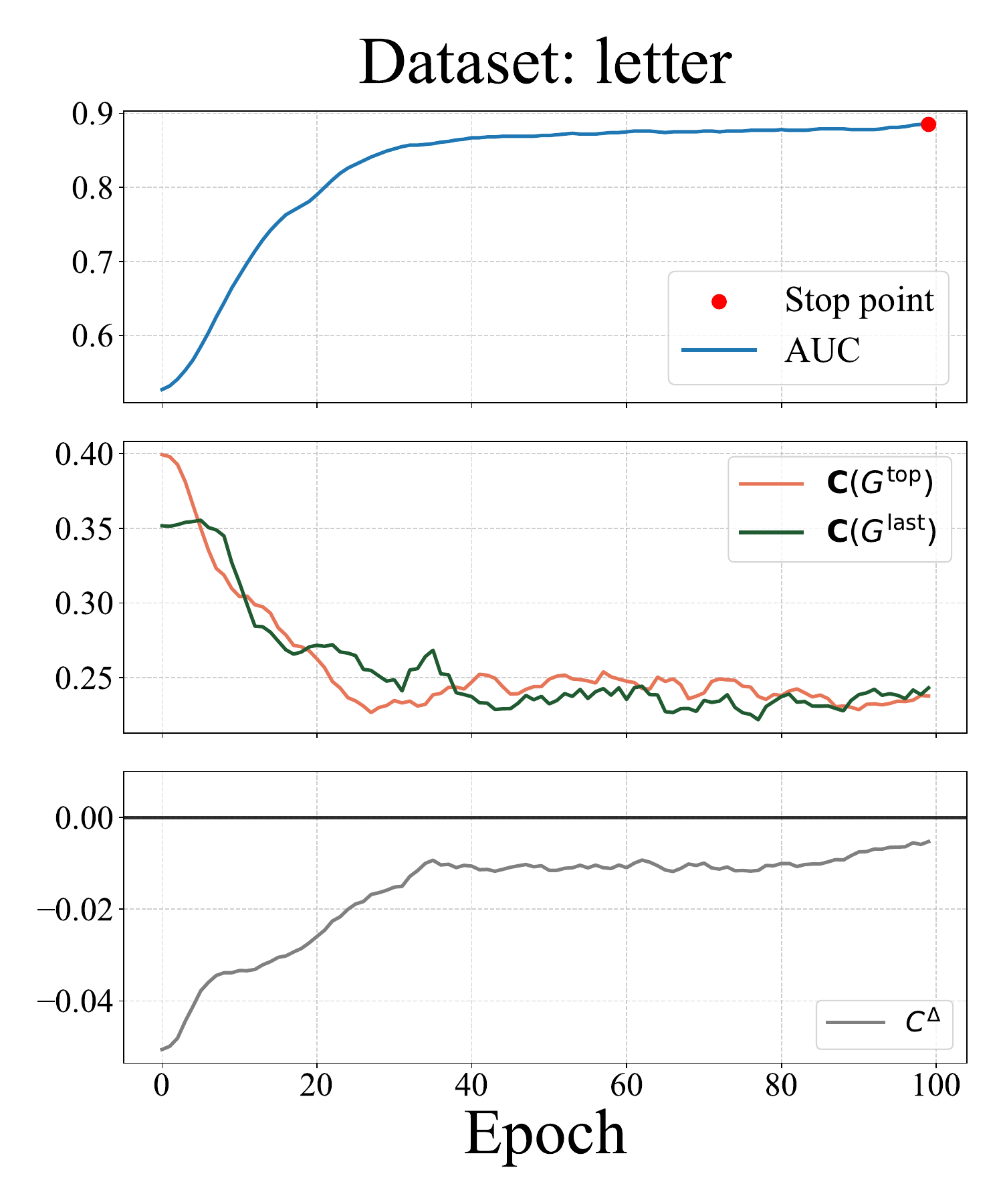}
\includegraphics[width=0.32\textwidth]{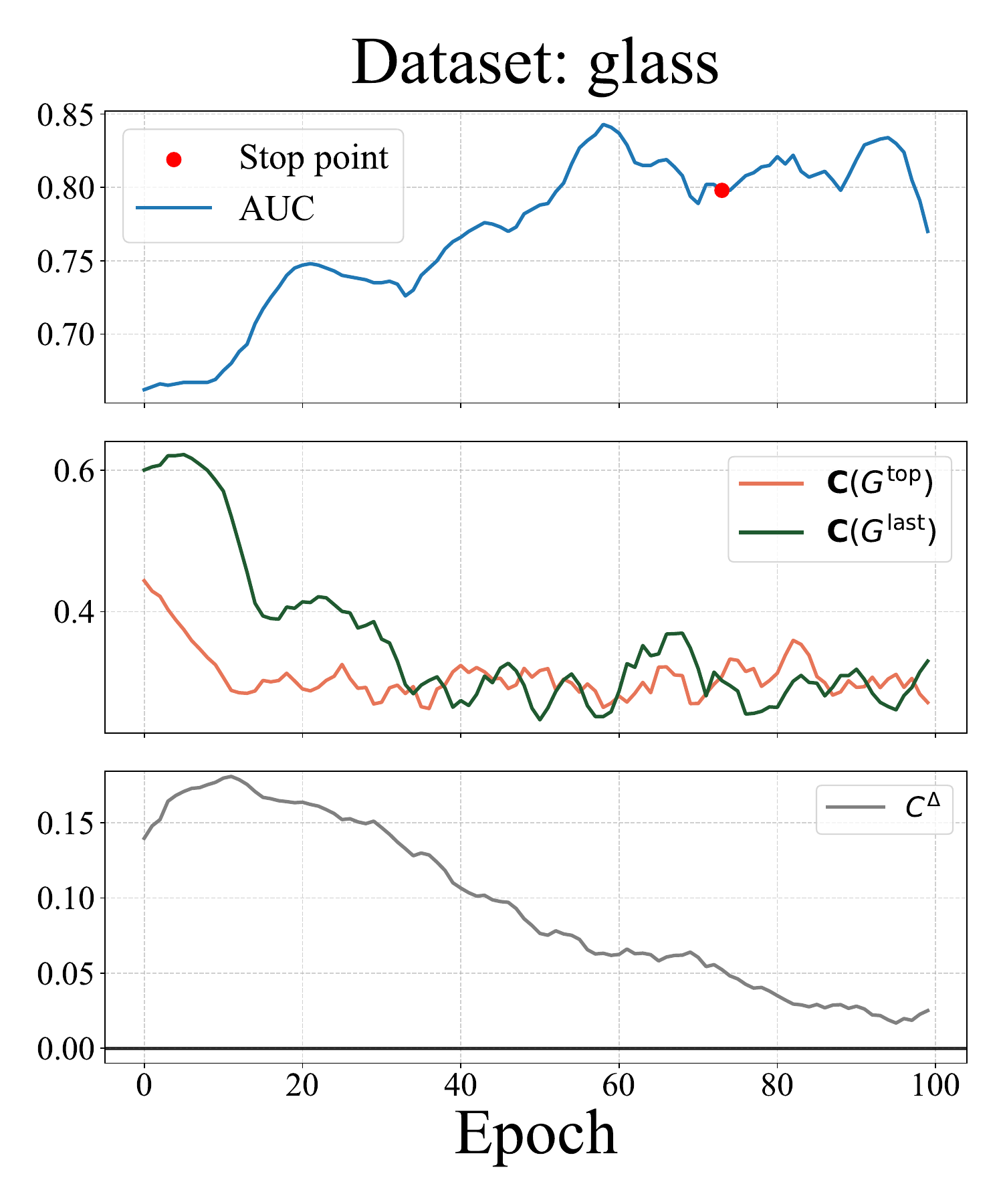}
\includegraphics[width=0.32\textwidth]{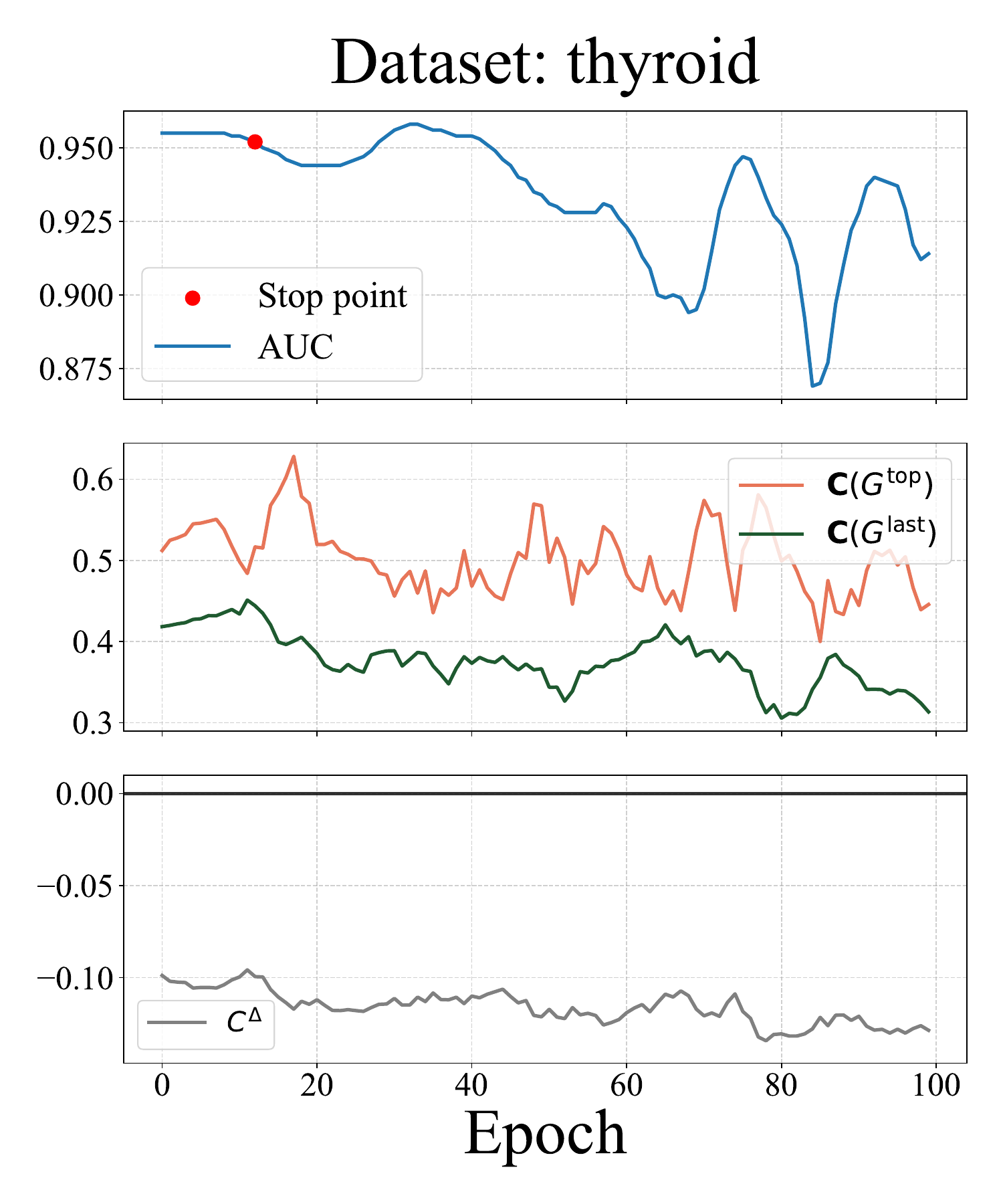}
\includegraphics[width=0.32\textwidth]{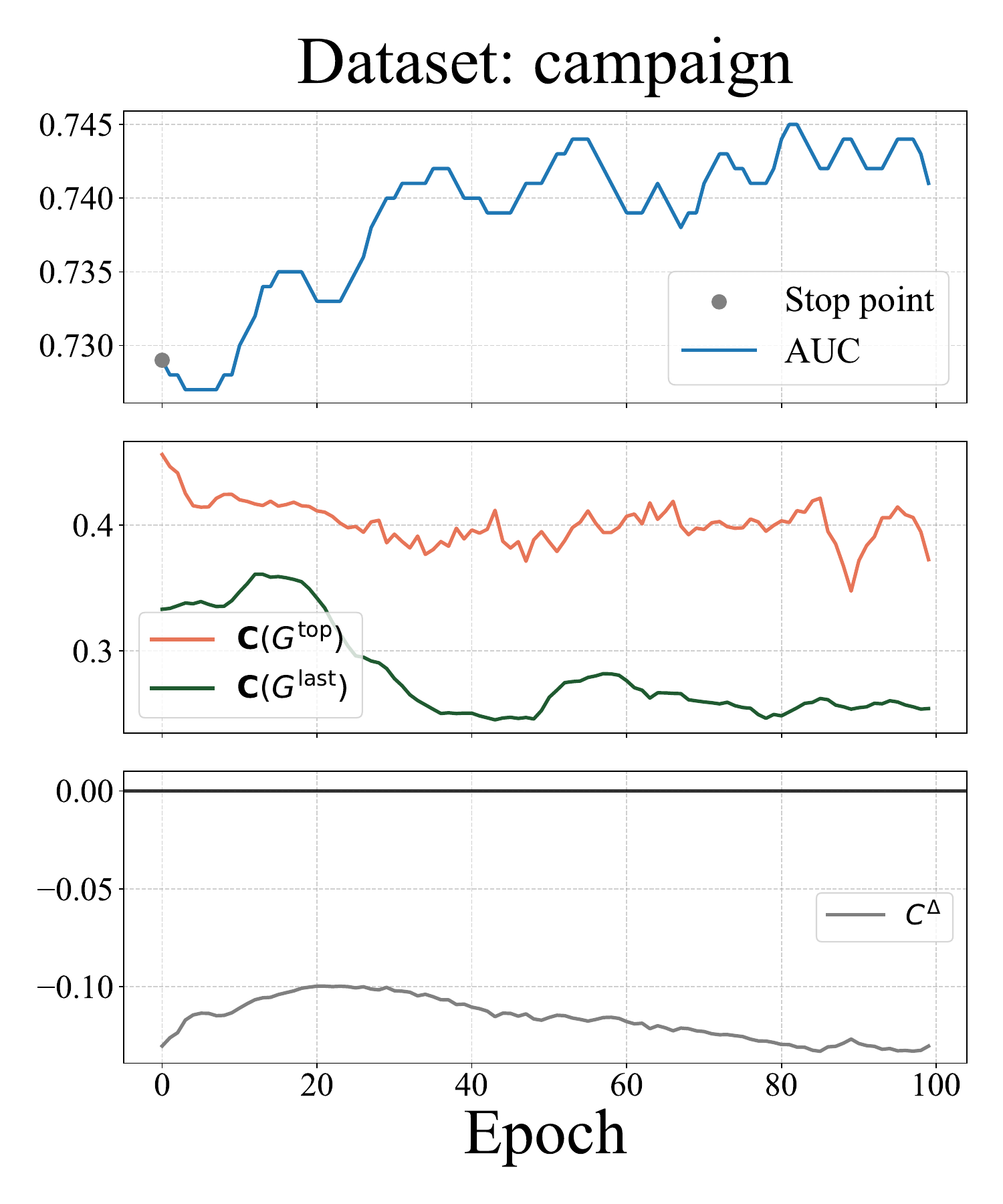}
\includegraphics[width=0.32\textwidth]{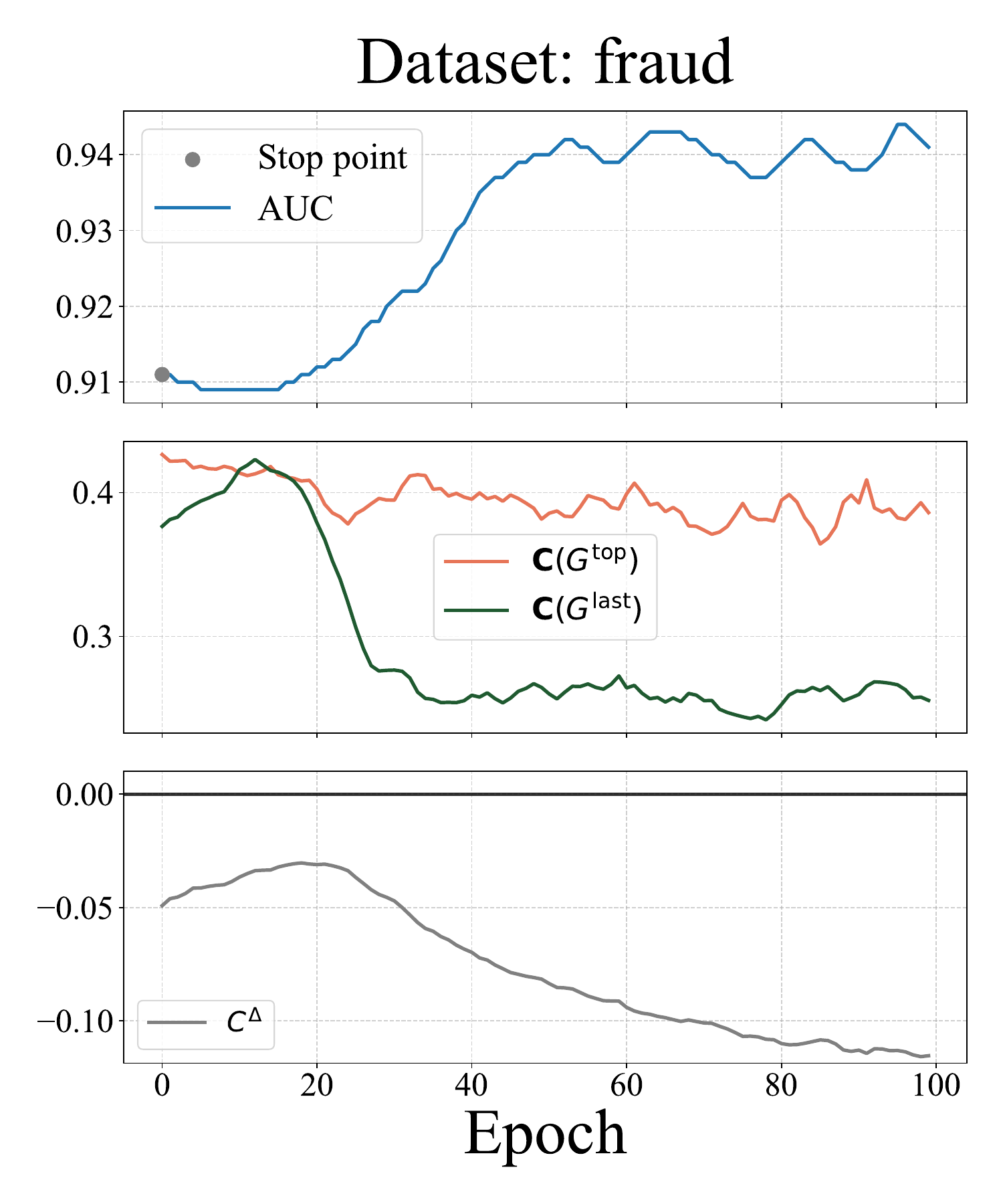}
\includegraphics[width=0.32\textwidth]{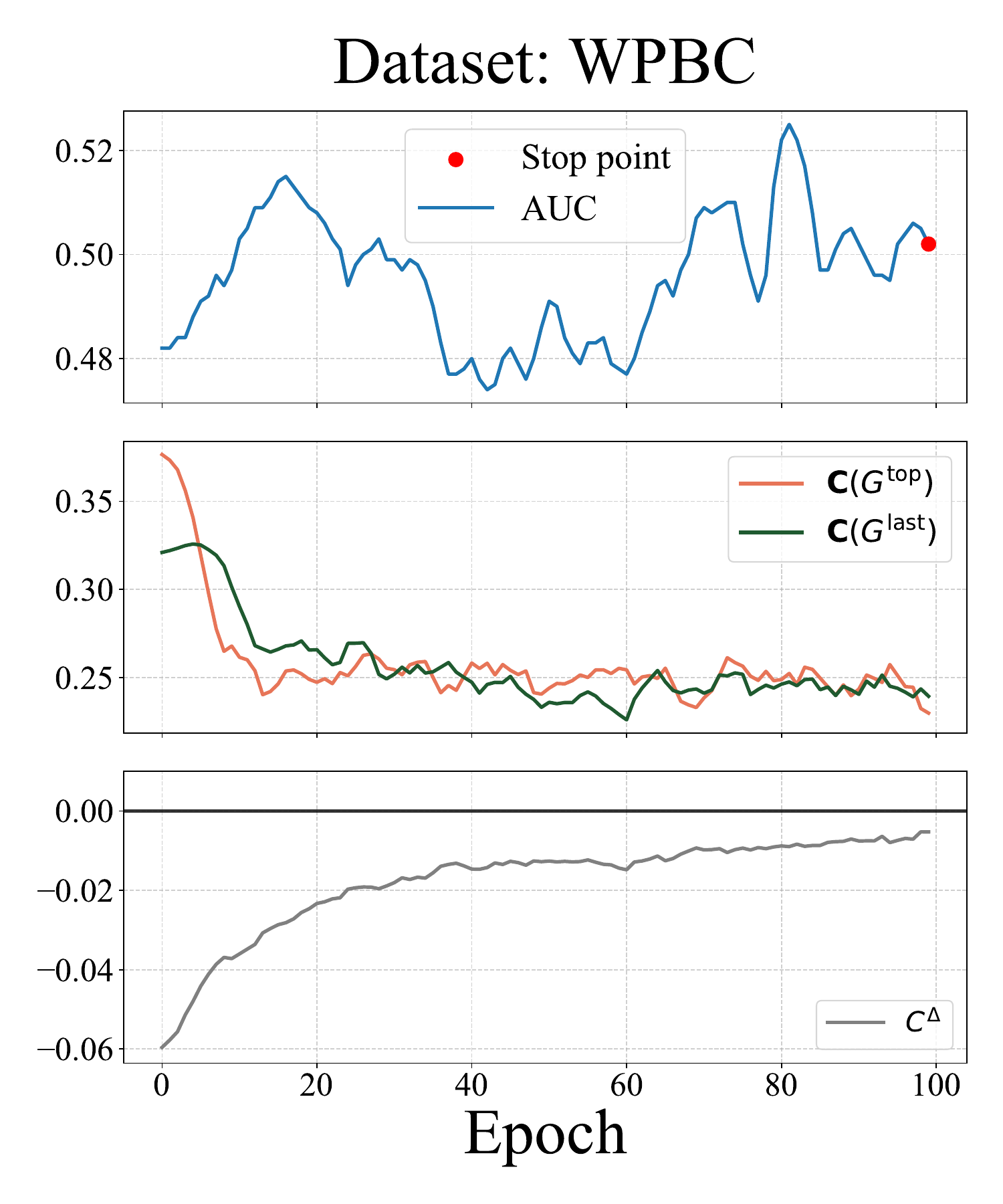}
  \caption{AE: AUC curves vs.  $\mathbf{C^\Delta}$ curves. Top: AUC. Middle: $\mathbf{C}(G^{\text{last}})$ and $\mathbf{C}(G^{\text{top}})$. Bottom: $C^{\Delta}=\mathbf{C}(G^{\text{last}})-\mathbf{C}(G^{\text{top}})$.}
  \label{Fig:all-curve-4}
\end{figure}

\begin{figure}[ht]
  \centering
\includegraphics[width=0.32\textwidth]{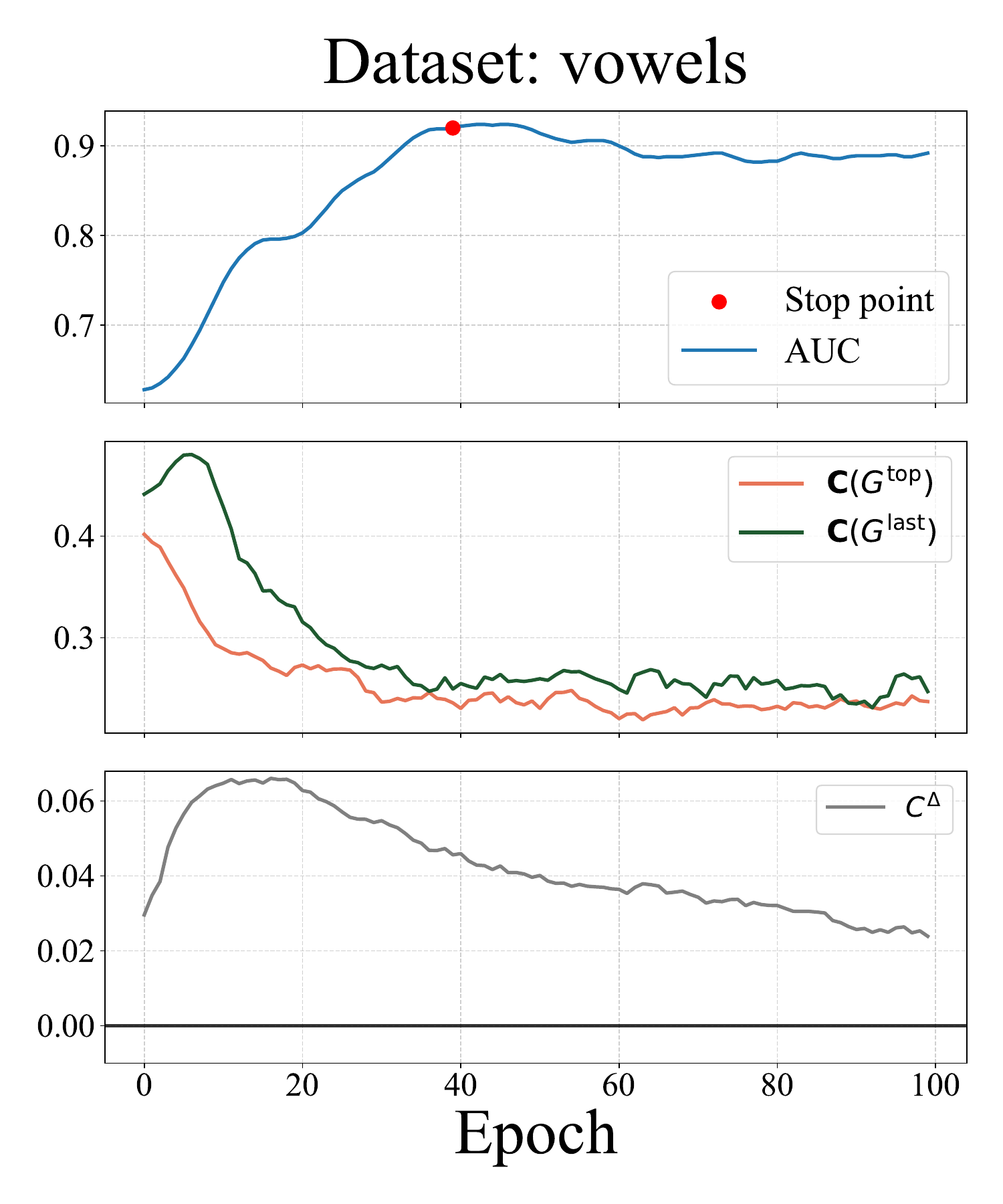}
\includegraphics[width=0.32\textwidth]{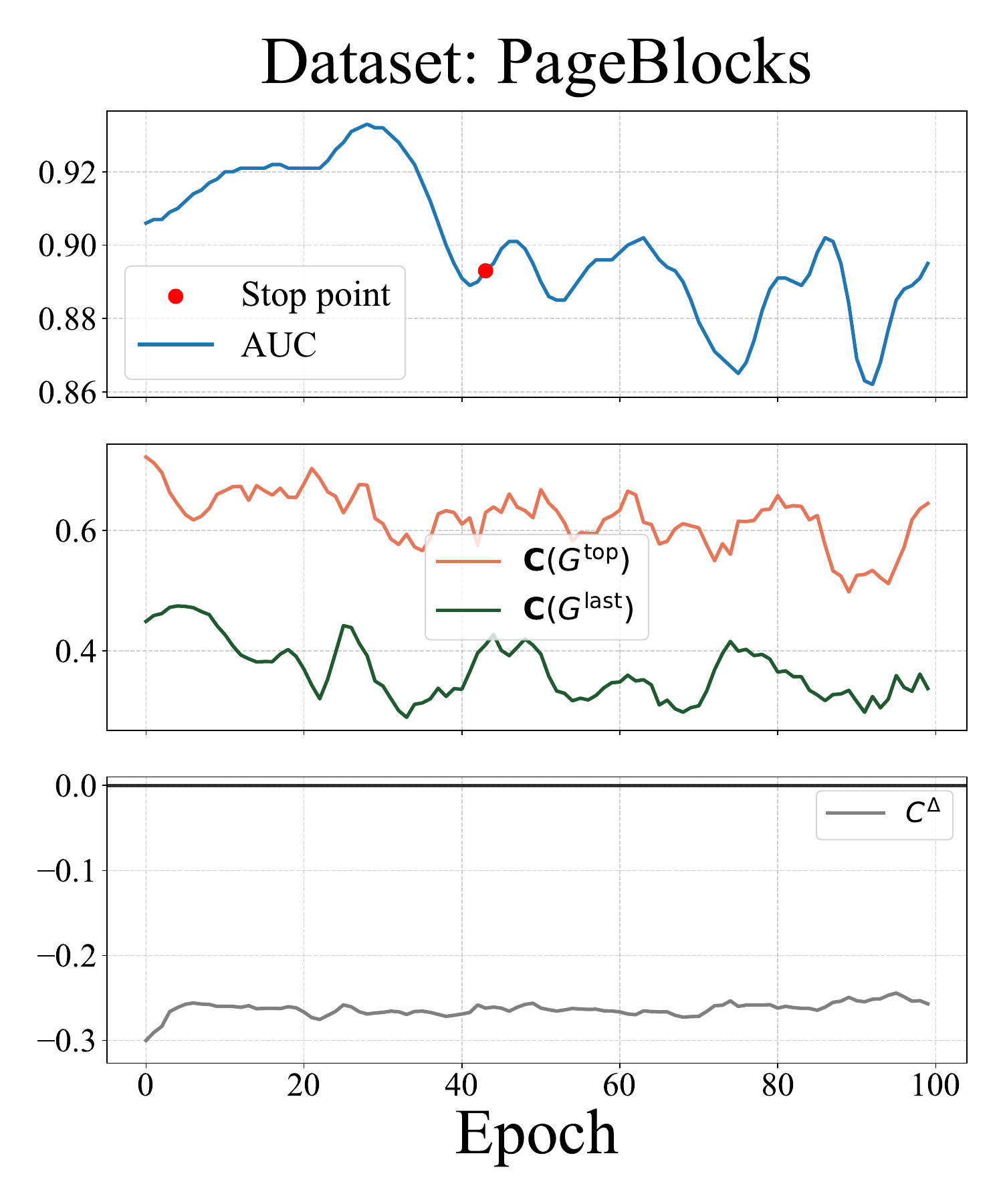}
\includegraphics[width=0.32\textwidth]{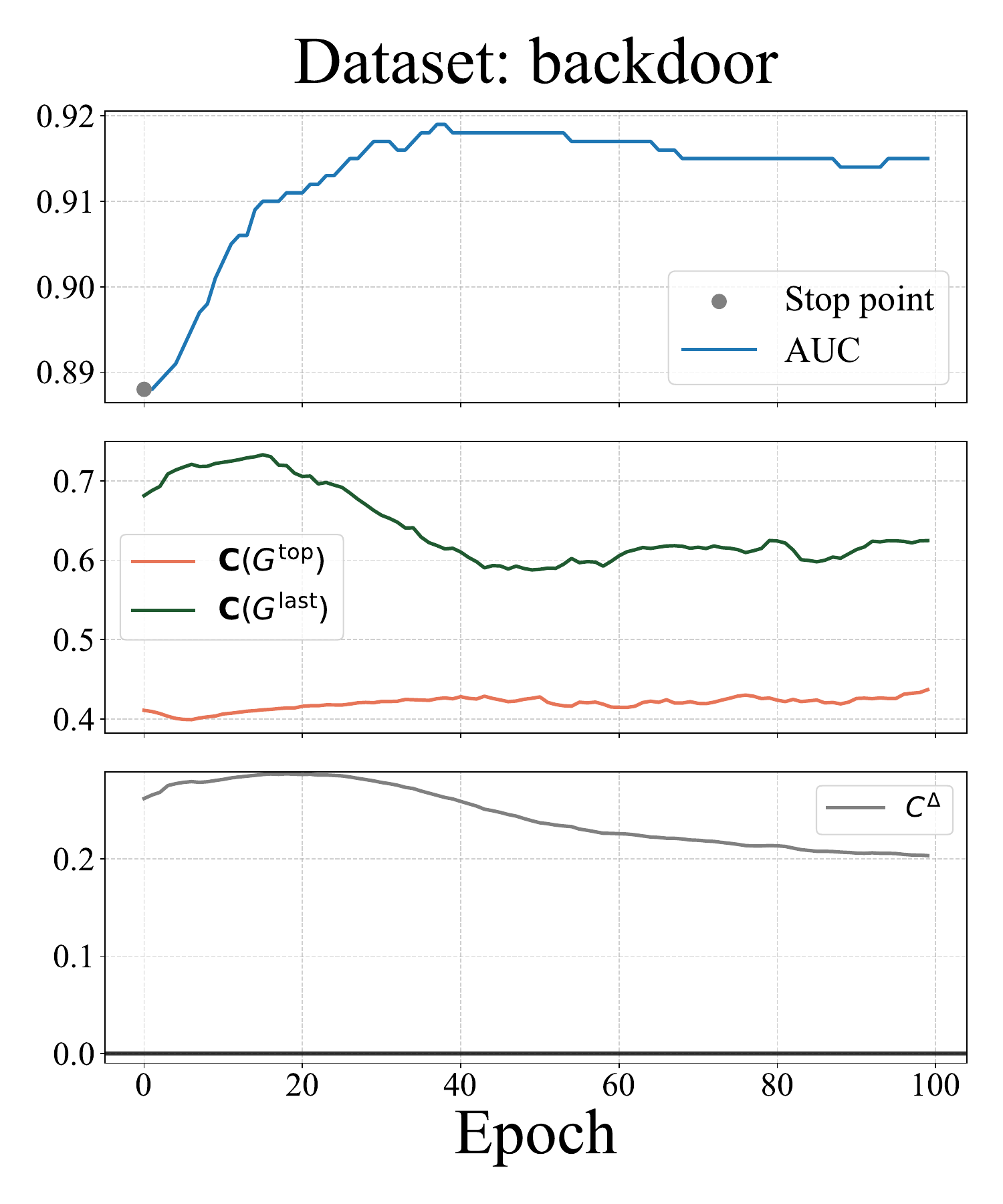}
\includegraphics[width=0.32\textwidth]{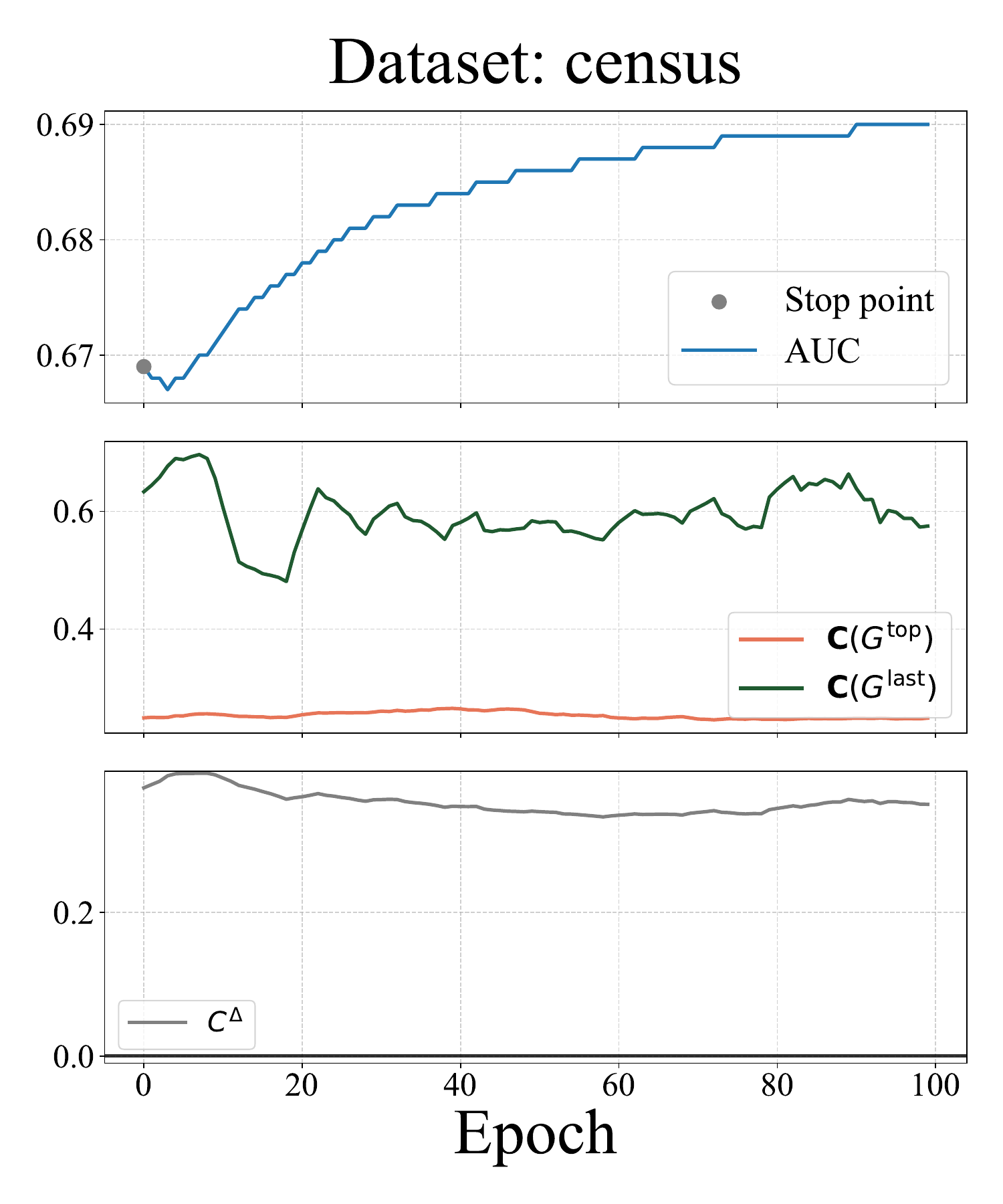}
\includegraphics[width=0.32\textwidth]{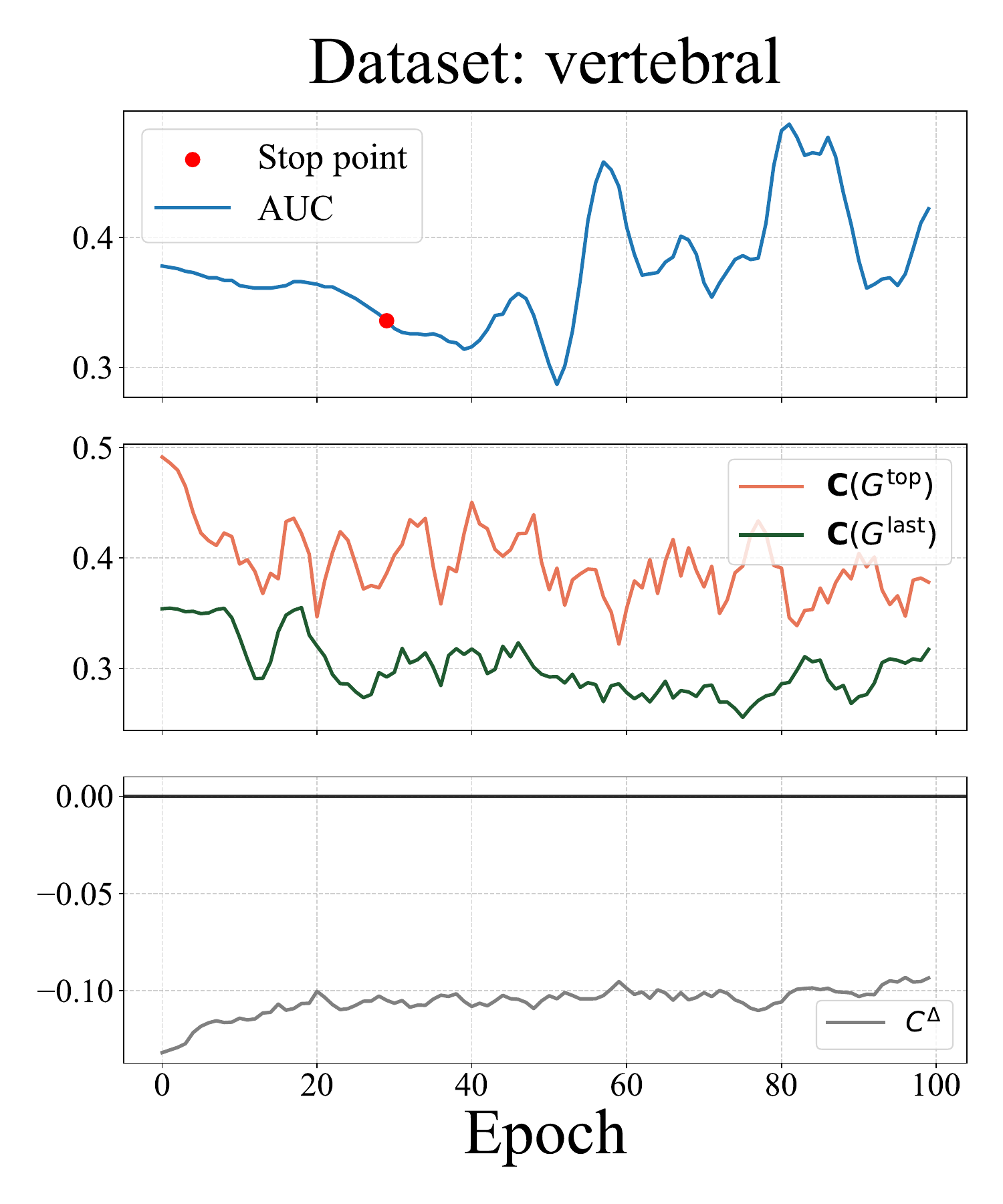}
\includegraphics[width=0.32\textwidth]{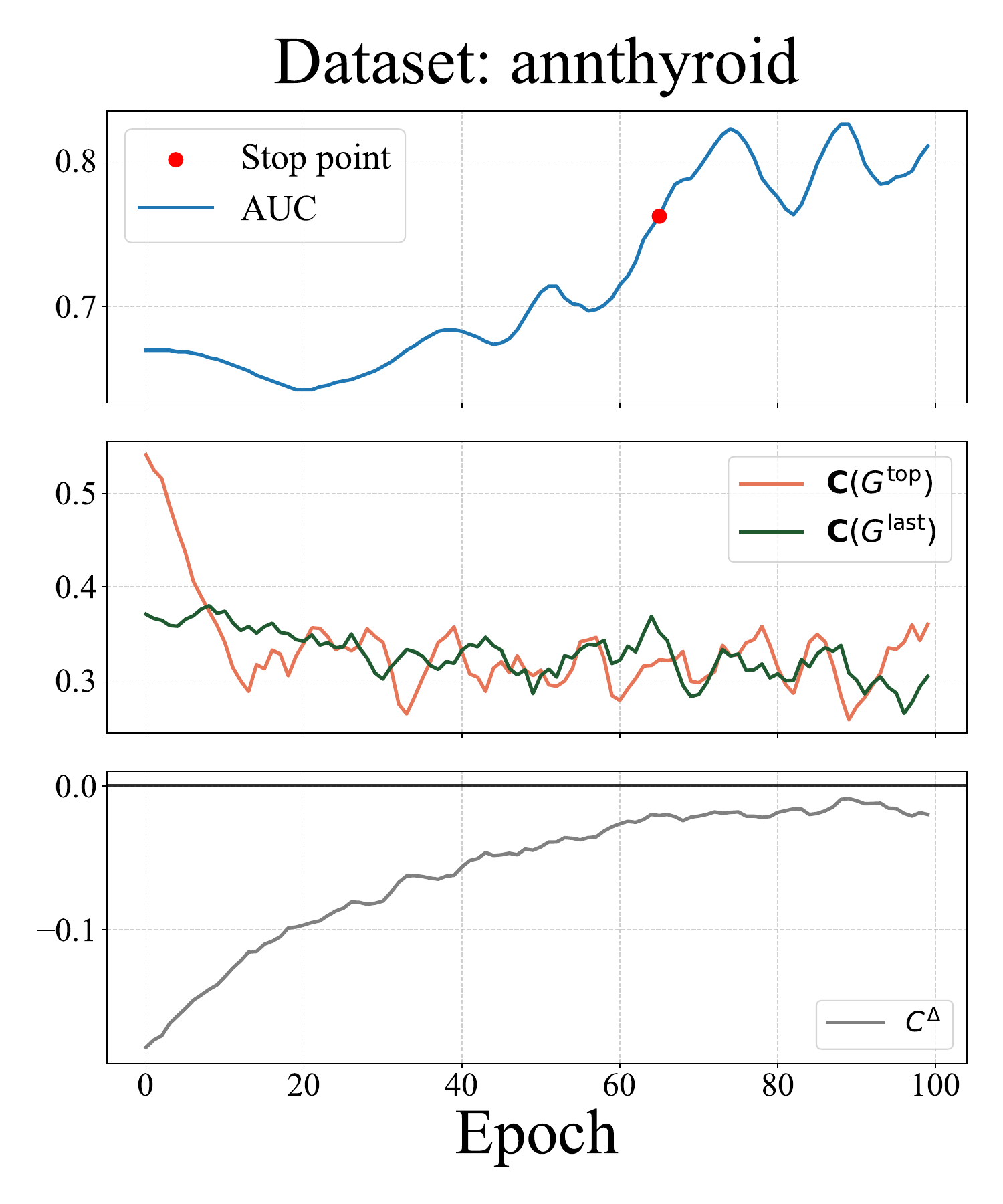}
\includegraphics[width=0.32\textwidth]{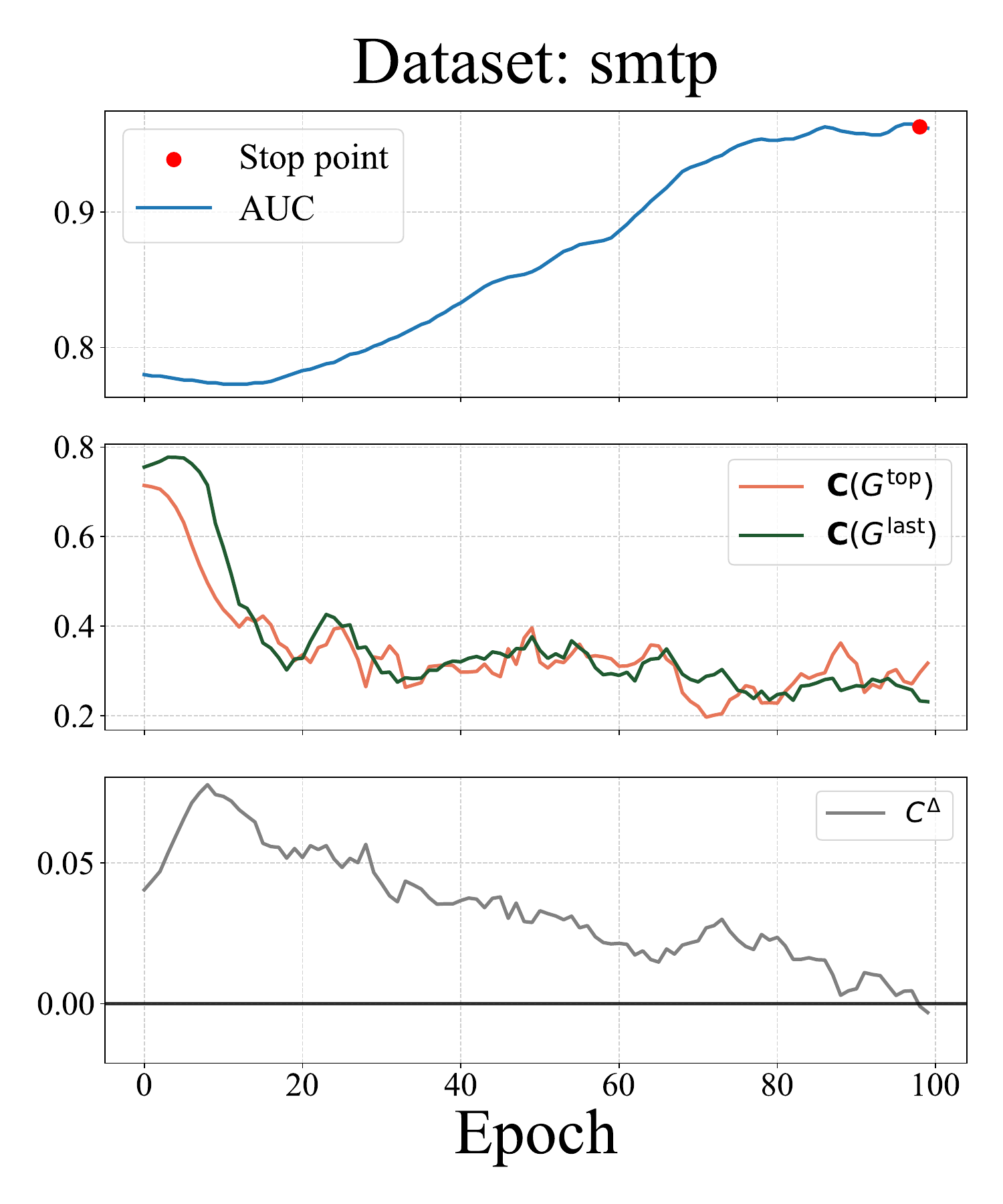}
\includegraphics[width=0.32\textwidth]{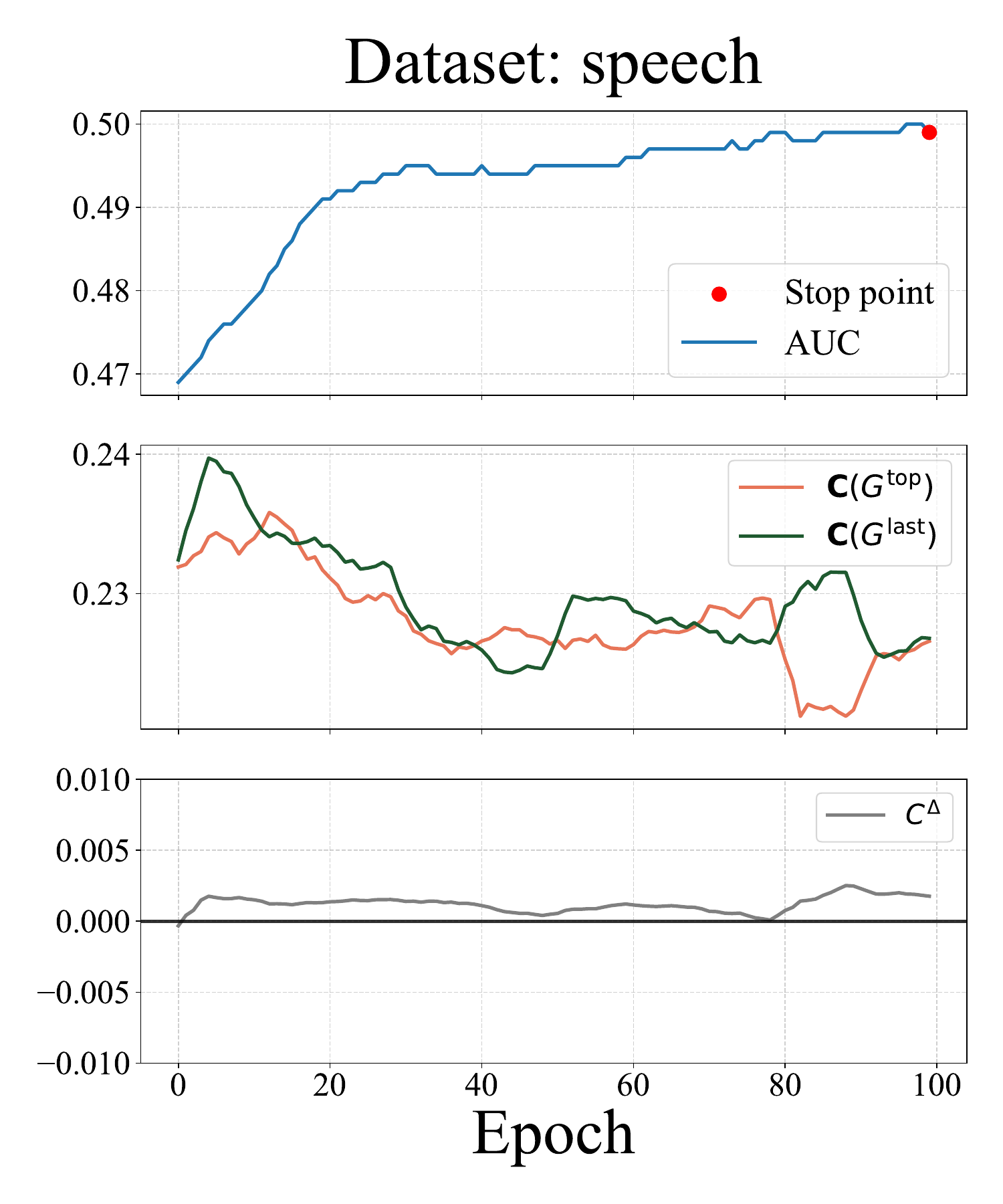}
\includegraphics[width=0.32\textwidth]{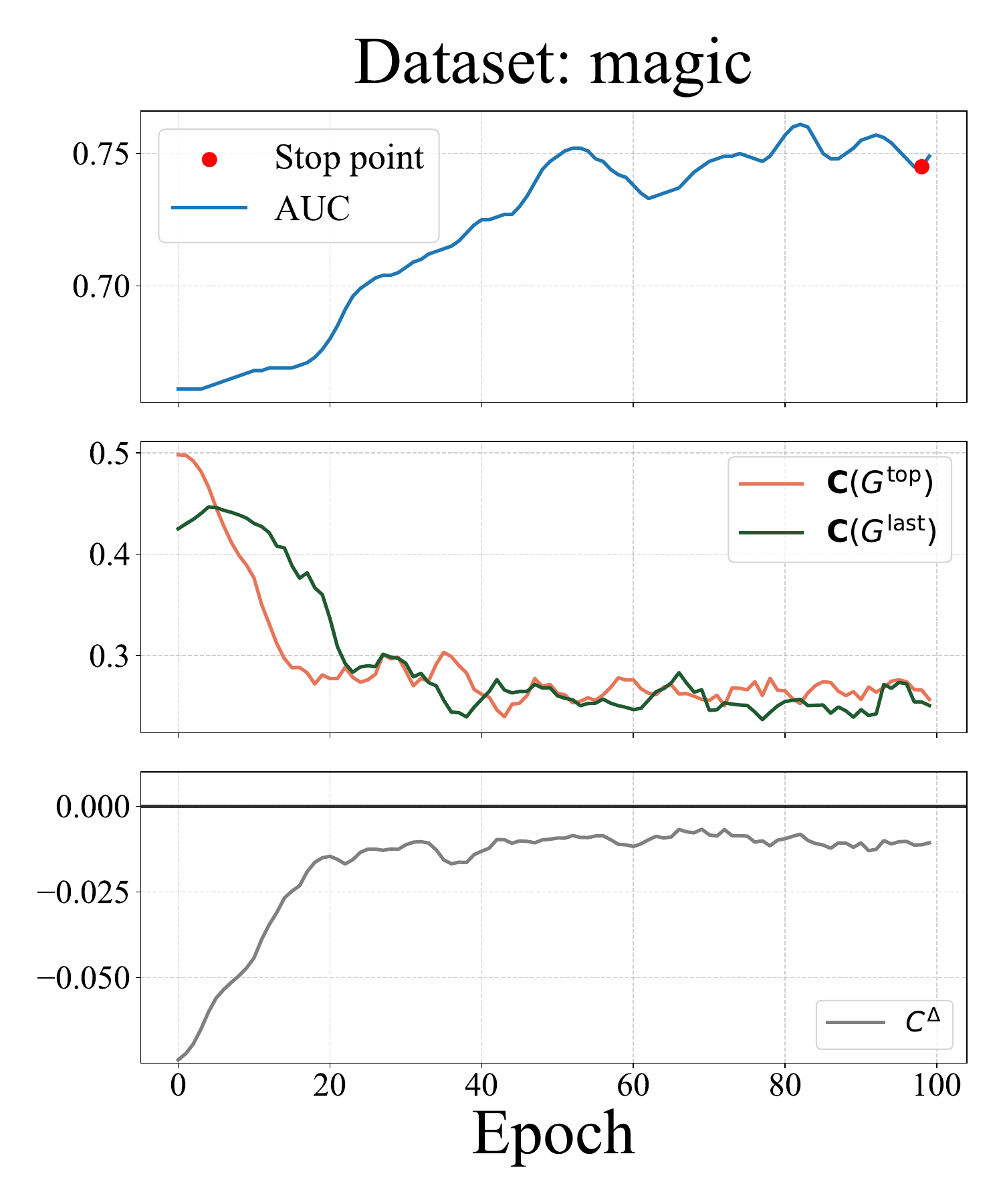}
  \caption{AE: AUC curves vs.  $\mathbf{C^\Delta}$ curves. Top: AUC. Middle: $\mathbf{C}(G^{\text{last}})$ and $\mathbf{C}(G^{\text{top}})$. Bottom: $C^{\Delta}=\mathbf{C}(G^{\text{last}})-\mathbf{C}(G^{\text{top}})$.}
  \label{Fig:all-curve-5}
\end{figure}

\begin{figure}[htbp]
  \centering
\includegraphics[width=0.32\textwidth]{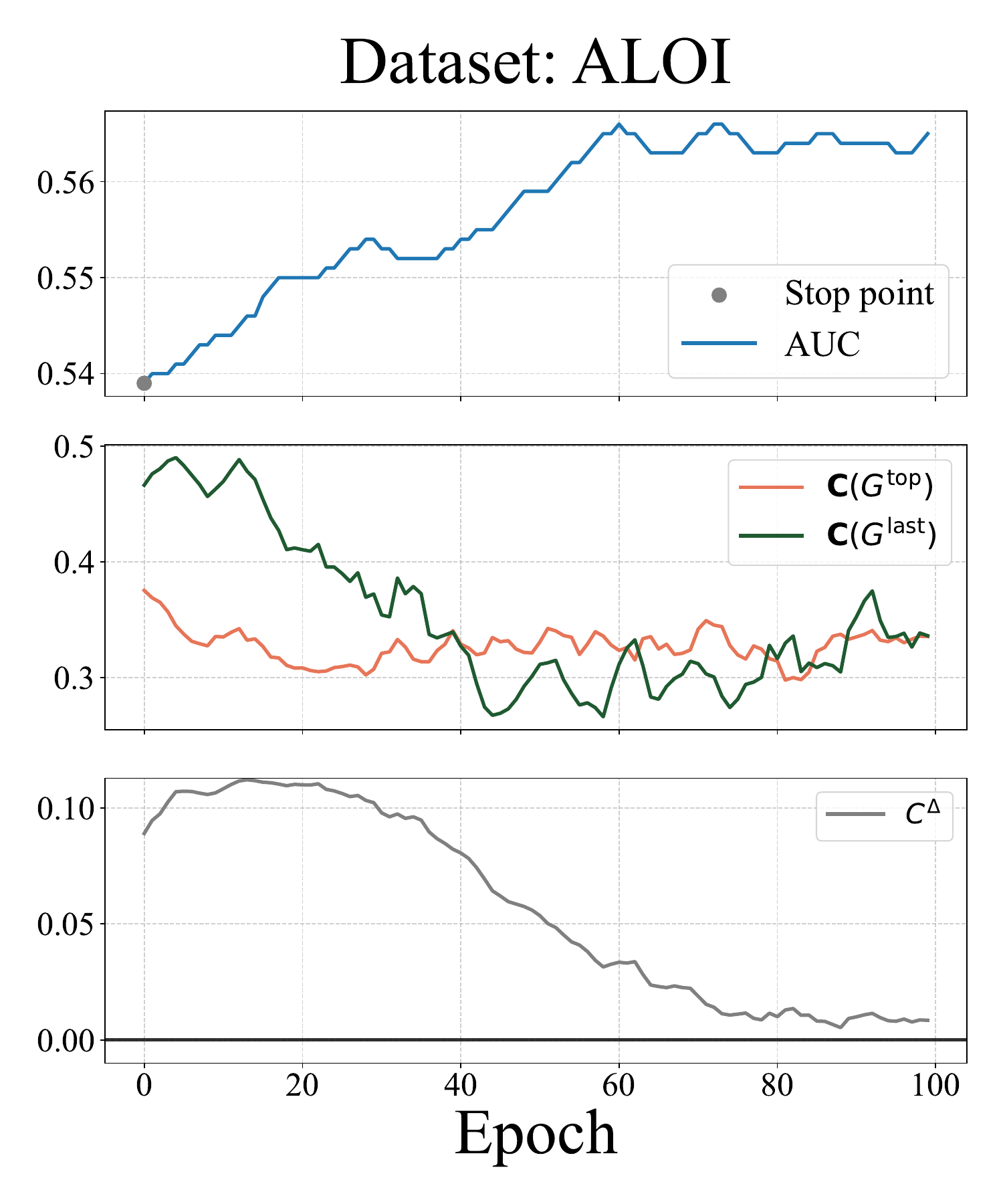}
\includegraphics[width=0.32\textwidth]{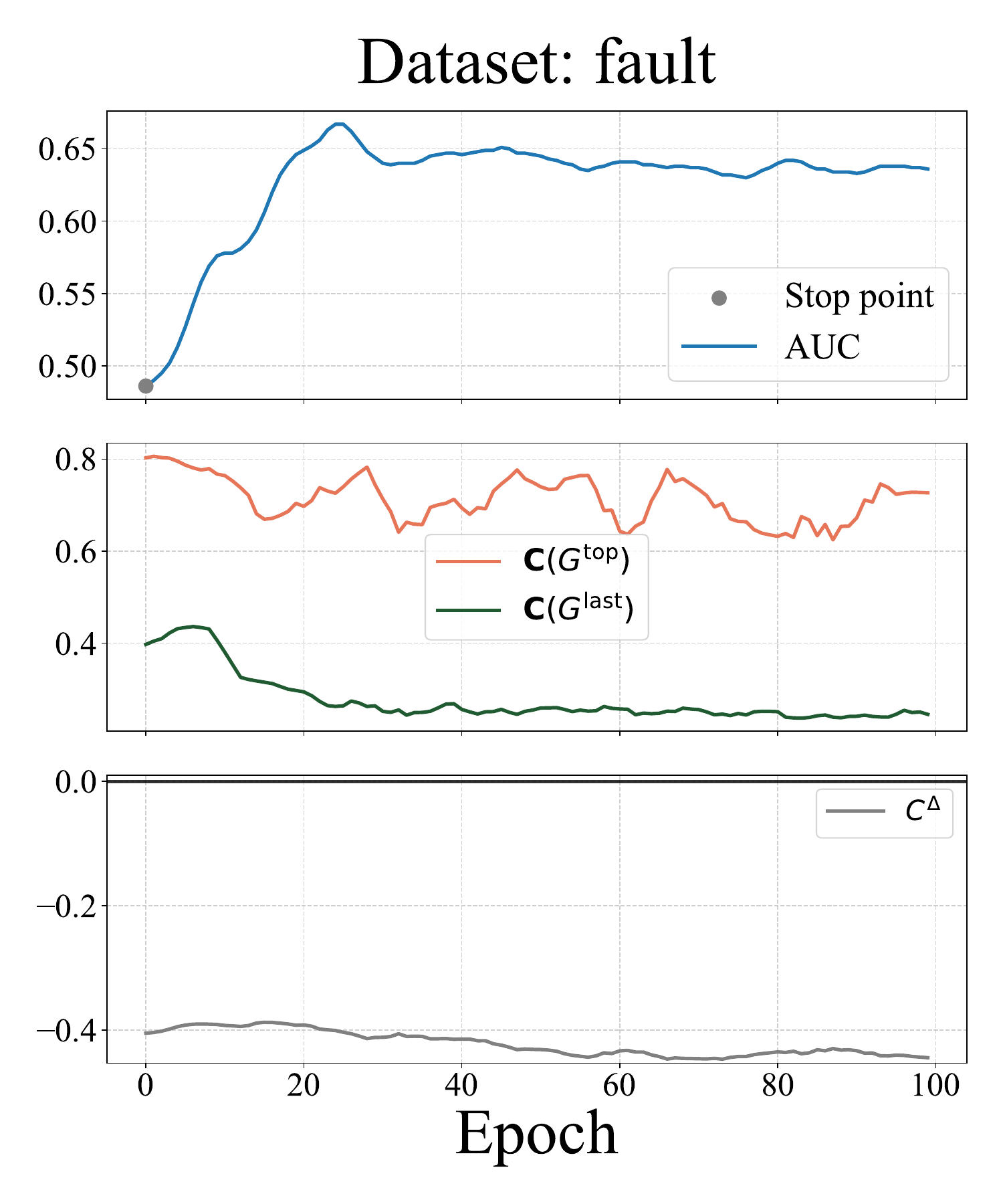}
  \caption{AE: AUC curves vs.  $\mathbf{C^\Delta}$ curves. Top: AUC. Middle: $\mathbf{C}(G^{\text{last}})$ and $\mathbf{C}(G^{\text{top}})$. Bottom: $C^{\Delta}=\mathbf{C}(G^{\text{last}})-\mathbf{C}(G^{\text{top}})$.}
  \label{Fig:all-curve-6}
\end{figure}

We put the training process of AE on all 47 datasets from Fig. \ref{Fig:all-curve-1} to \ref{Fig:all-curve-6}. The meaning of the plots is similar to the ones in Fig. \ref{fig:case-studies} in the case study, except for a red point indicating stop with cohesion metric $C^{\Delta}$ and a grey point indicating stop with divergence metric $\mathbf{D}$.
As can be observed from the figures, GradStop is capable of identifying a better or even the optimal epoch; however, it fails to work effectively on a few datasets and may sometimes lead to a worse performance than the final epoch. Following, we analyze typical dataset scenarios in conjunction with experimental results.

On datasets such as \textit{glass}, \textit{Lymphography}, \textit{landsat}, \textit{letter}, \textit{mnist}, \textit{shuttle}, etc., $C^{\Delta}$ typically functions well. When AUC increases, $C^{\Delta}$ either gradually rises, indicating a trend towards progressively beneficial training, or manifests as a significantly positive value. Conversely, when training becomes toxic, $C^{\Delta}$ often exhibits a declining trend or negative values. Note that in some cases, despite $C^{\Delta}$ being negative, an upward trend still assures training to be beneficial. This is because the GradSample method does not perfectly retrieve in/outlier sets but rather approximates based on the current alignment degree of outlier assumptions. Thus, even if $C^{\Delta}$ is negative, its rising trend suggests beneficial training. In cases like \textit{WBC}, \textit{Stamps}, \textit{wine}, $|C^{\Delta}|$ is extremely small, indicating chaotic training leading AUC to fluctuate.

On datasets such as \textit{SpamBase}, \textit{speech}, \textit{WPBC}, etc., $C^{\Delta}$ proves ineffective. This inefficacy arises from AE's complete inability to effectively detect outliers in these datasets throughout the training process, with an AUC of around 0.5. Neither inlier priority nor the inherent difficulty of outlier construction is satisfied. Consequently, the assumptions of the GradSample method do not hold, making it incapable of approximating in/outlier distributions accurately. In such scenarios, early stopping is meaningless and changing the UOD model is needed.

$\mathbf{D}$ effectively identifies the sufficient effectiveness of random parameters, preventing complete toxic training in datasets such as \textit{InternetAds}, \textit{pendigits}, \textit{cardio}, etc. On these datasets, only the inherent difficulty of outlier construction holds while inlier priority does not. A $\mathbf{D}>\pi/2$ signifies that the model structure with random parameters has adequately distinguished between normal and anomalous data, and further training is unlikely to enhance the separation between their distributions.

However, on certain datasets such as \textit{backdoor}, $\mathbf{D}$ erroneously stops training, despite $C^{\Delta}$ indicating a beneficial training is going on. More sophisticated algorithms that combine $C^{\Delta}$ and $\mathbf{D}$, rather than abruptly stop at $\mathbf{D}>\pi/2$ could potentially improve early stopping outcomes. Since current work mainly focuses on evaluating respective metrics, we aim to explore this approach in our future work.

We also put the ranked performance on individual datasets of AE, DeepSVDD, RDP, VAE, and their GradStop versions in Table \ref{tab:individual-datasets-ae}, \ref{tab:individual-datasets-svdd}, \ref{tab:individual-datasets-rdp}, and \ref{tab:individual-datasets-vae}, respectively. For clarity, only datasets on which the absolute improvement in AUC by GradStop is greater than $5\%$ (for AE and DeepSVDD) and $0.5\%$ (for RDP and VAE) are listed. RDP and VAE only suffer from performance degradation on several datasets due to their strong model constraints. Consequently, as seen in the table, early stopping does not significantly improve or decrease their overall performance and most improvements in the table are close to 0.

\begin{table}[b]
  \centering
  \caption{AEs' performance on individual datasets with $abs(\text{improvements})>5\%$ among 47 datasets.}
  \vskip 0.1in
    \resizebox{0.6\columnwidth}{!}{
    \begin{tabular}{rlccc}
    \toprule
    \toprule
    \multicolumn{2}{c}{\textbf{Dataset}} & VanillaAE    & AE-G  & Improvement \\
    \midrule
    1     & Cardiotocography & 0.571  & 0.751  & 31.39\% \\
    2    & pendigits & 0.769  & 0.930  & 20.99\% \\
    3    & donors & 0.707  & 0.834  & 17.86\% \\
    4     & cardio & 0.809  & 0.950  & 17.43\% \\
    5    & mammography & 0.738  & 0.846  & 14.64\% \\
    6    & optdigits & 0.455  & 0.521  & 14.59\% \\
    7     & celeba & 0.729  & 0.834  & 14.41\% \\
    8    & InternetAds & 0.559  & 0.615  & 10.02\% \\
    9    & satimage-2 & 0.906  & 0.990  & 9.27\% \\
    10    & Stamps & 0.816  & 0.871  & 6.74\% \\
    11    & SpamBase & 0.522  & 0.550  & 5.43\% \\
    12    & Hepatitis & 0.711  & 0.745  & 4.67\% \\
    $\dots$    & $\dots$ & $\dots$  & $\dots$  & $\dots$ \\	
    46     & ALOI & 0.568  & 0.540  & -4.99\% \\
    47    & fault & 0.647  & 0.550  & -15.04\% \\
    \bottomrule
    \bottomrule
    \end{tabular}%
    }
  \label{tab:individual-datasets-ae}%
\end{table}%

\begin{table}[htbb]
  \centering
  \caption{DeepSVDD's performance on individual datasets with $abs(\text{improvements})>5\%$ among 47 datasets.}
  \vskip 0.1in
    \resizebox{0.56 \columnwidth}{!}{
    \begin{tabular}{rlccc}
    \toprule
    \toprule
    \multicolumn{2}{c}{\textbf{Dataset}} & DeepSVDD & \multicolumn{1}{l}{DeepSVDD-G} & Improvement \\
    \midrule
    1   & speech & 0.380  & 0.977  & 157.33\% \\
    2    & Cardiotocography & 0.409  & 0.998  & 143.73\% \\
    3   & Wilt & 0.411  & 0.913  & 121.88\% \\
    4   & satellite & 0.439  & 0.917  & 108.97\% \\
    5   & fraud & 0.377  & 0.744  & 97.52\% \\
    6    & breastw & 0.460  & 0.883  & 92.17\% \\
    7    & musk & 0.381  & 0.727  & 90.65\% \\
    8    & mnist & 0.503  & 0.867  & 72.19\% \\
    9    & WBC & 0.565  & 0.943  & 66.84\% \\
    10    & satimage-2 & 0.482  & 0.800  & 66.04\% \\
    11    & landsat & 0.553  & 0.893  & 61.64\% \\
    12    & ALOI & 0.433  & 0.669  & 54.50\% \\
    13    & skin & 0.490  & 0.753  & 53.57\% \\
    14    & census & 0.524  & 0.791  & 50.79\% \\
    15    & PageBlocks & 0.537  & 0.806  & 49.94\% \\
    16    & letter & 0.517  & 0.723  & 39.76\% \\
    17    & http & 0.490  & 0.676  & 37.99\% \\
    18    & vertebral & 0.413  & 0.530  & 28.35\% \\
    19    & magic.gamma & 0.537  & 0.684  & 27.37\% \\
    20    & Hepatitis & 0.478  & 0.607  & 26.83\% \\
    21    & Lymphography & 0.515  & 0.644  & 24.90\% \\
    22    & wine & 0.434  & 0.528  & 21.83\% \\
    23    & WDBC & 0.498  & 0.604  & 21.43\% \\
    24    & optdigits & 0.540  & 0.634  & 17.48\% \\
    25    & InternetAds & 0.579  & 0.667  & 15.26\% \\
    26    & pendigits & 0.492  & 0.567  & 15.23\% \\
    27    & WPBC & 0.475  & 0.543  & 14.25\% \\
    28    & cardio & 0.460  & 0.515  & 11.88\% \\
    29    & mammography & 0.514  & 0.554  & 7.72\% \\
    30    & glass & 0.542  & 0.581  & 7.07\% \\
    31    & cover & 0.531  & 0.563  & 5.90\% \\
    32    & annthyroid & 0.526  & 0.555  & 5.45\% \\
    33    & celeba & 0.499  & 0.526  & 5.41\% \\
    34    & backdoor & 0.535  & 0.555  & 3.68\% \\
    ...     & ...   & ...     & ...     & ... \\
    41    & fault & 0.622  & 0.596  & -4.23\% \\
    42    & campaign & 0.649  & 0.611  & -5.80\% \\
    43    & Ionosphere & 0.552  & 0.487  & -11.66\% \\
    44    & shuttle & 0.523  & 0.439  & -16.07\% \\
    45    & Waveform & 0.533  & 0.415  & -22.08\% \\
    46    & thyroid & 0.590  & 0.446  & -24.51\% \\
    47    & Stamps & 0.393  & 0.277  & -29.52\% \\
    \bottomrule
    \bottomrule
    \end{tabular}%
    }
  \label{tab:individual-datasets-svdd}%
\end{table}%

\begin{table}[htbp]
  \centering
  \caption{RDP's performance on individual datasets with $abs(\text{improvements})>0.5\%$ among 47 datasets.}
    \vskip 0.1in
    \begin{tabular}{rlccc}
    \toprule
    \toprule
    \multicolumn{2}{c}{\textbf{Dataset}} & RDP  & RDP-G & Improvement \\
    \midrule
    1     & vertebral & 0.556 & 0.708 & 27.35\% \\  
    2     & Ionosphere & 0.410 & 0.481 & 17.32\% \\
    3     & backdoor & 0.496 & 0.563 & 13.50\% \\
    4     & WDBC & 0.706 & 0.723 & 2.41\% \\
    5     & cardio & 0.689 & 0.703 & 2.03\% \\
    6     & fault & 0.648 & 0.661 & 2.01\% \\
    7     & campaign & 0.767 & 0.781 & 1.78\% \\
    8     & optdigits & 0.602 & 0.613 & 1.77\% \\
    9     & Hepatitis & 0.816 & 0.831 & 1.76\% \\
    10    & census & 0.829 & 0.840 & 1.33\% \\
    11    & SpamBase & 0.679 & 0.685 & 0.88\% \\
    12    & glass & 0.699 & 0.705 & 0.86\% \\
    13    & magic.gamma & 0.707 & 0.713 & 0.85\% \\
    14    & Pima & 0.514 & 0.517 & 0.65\% \\
    15    & pendigits & 0.542 & 0.544 & 0.37\% \\
    ...    & ... & ... & ... & ... \\
    35    & skin & 0.893 & 0.889 & -0.45\% \\
    36    & satellite & 0.866 & 0.861 & -0.58\% \\
    37    & Waveform & 0.394 & 0.391 & -0.68\% \\
    38    & annthyroid & 0.740 & 0.735 & -0.72\% \\
    39    & mnist & 0.988 & 0.978 & -1.01\% \\
    40    & wine & 0.704 & 0.692 & -1.61\% \\
    41    & shuttle & 0.331 & 0.325 & -1.61\% \\
    42    & http & 0.847 & 0.831 & -1.89\% \\
    43    & Wilt & 0.926 & 0.908 & -1.91\% \\
    44    & smtp & 0.772 & 0.752 & -2.51\% \\
    45    & celeba & 0.666 & 0.648 & -2.70\% \\
    46    & ALOI & 0.900 & 0.869 & -3.44\% \\
    47    & letter & 0.850 & 0.819 & -3.65\% \\
    \bottomrule
    \bottomrule
    \end{tabular}%
  \label{tab:individual-datasets-rdp}%
\end{table}%

\begin{table}[htbp]
  \centering
  \caption{VAE's performance on individual datasets with $abs(\text{improvements})>0.5\%$ among 47 datasets.}
    \vskip 0.1in
    \begin{tabular}{rlccc}
    \toprule
    \toprule
    \multicolumn{2}{c}{\textbf{Dataset}} & VAE  & VAE-G & Improvement \\
    \midrule
    1     & optdigits & 0.548 & 0.627 & 14.42\% \\
    2     & wine & 0.743 & 0.755 & 1.62\% \\
    3     & Waveform & 0.413 & 0.419 & 1.45\% \\
    4     & donors & 0.536 & 0.541 & 0.93\% \\
    5     & Lymphography & 0.555 & 0.559 & 0.72\% \\
    6     & shuttle & 0.524 & 0.527 & 0.57\% \\
    7     & backdoor & 0.539 & 0.542 & 0.56\% \\
    8     & Hepatitis & 0.729 & 0.733 & 0.55\% \\
    9     & WPBC & 0.822 & 0.826 & 0.49\% \\
    ...   & ... & ... & ... & ... \\
    43    & fault & 0.612 & 0.609 & -0.49\% \\
    44    & campaign & 0.75  & 0.745 & -0.67\% \\
    45    & WDBC & 0.738 & 0.733 & -0.68\% \\
    46    & landsat & 0.916 & 0.908 & -0.87\% \\
    47    & pendigits & 0.531 & 0.479 & -9.79\% \\
    \bottomrule
    \bottomrule
    \end{tabular}%
  \label{tab:individual-datasets-vae}%
\end{table}%

\end{document}